\pdfoutput=1
\documentclass[journal]{IEEEtran}

\IEEEoverridecommandlockouts                              % This command is only needed if 
\usepackage{moreverb,url}
\usepackage[colorlinks,bookmarksopen,bookmarksnumbered,citecolor=red,urlcolor=red,hypertexnames=false]{hyperref}
\usepackage{dsfont}

\usepackage{longtable}
\usepackage{supertabular,booktabs}
\usepackage{comment}
\usepackage{amsmath}
\usepackage{amsthm}
\usepackage{nicefrac}

\usepackage{comment}
\usepackage{siunitx}
\usepackage{relsize}
\usepackage{ifthen}
\usepackage[colorinlistoftodos]{todonotes}

\usepackage[caption=false]{subfig}

\usepackage[vlined,ruled,linesnumbered]{algorithm2e}
\usepackage{graphics} % for pdf, bitmapped graphics files
\usepackage{rotating}
\usepackage{color}
\usepackage{enumerate}
\usepackage[T1]{fontenc}
\usepackage{psfrag}
\usepackage{epsfig} % for postscript graphics files
\usepackage{booktabs}
\usepackage{graphicx,url}
\usepackage{multirow}
\usepackage{array}
\usepackage{latexsym}
\usepackage{amsfonts}
\usepackage{amsmath}
\usepackage{amssymb}
\usepackage{xstring}
\usepackage[noend]{algorithmic}
\usepackage{multirow}
\usepackage{xcolor}
\usepackage{prettyref}
\usepackage{flexisym}
\usepackage{bigdelim}
\usepackage{breqn} % load this last
\usepackage{listings}
\let\labelindent\relax
\usepackage{enumitem}
\usepackage{xspace}
\usepackage{bm}
\graphicspath{{./figures/}}
\usepackage{tikz}
\usetikzlibrary{matrix,calc}

\usepackage{mdwlist}

\let\itemize\undefined
\makecompactlist{itemize}{stditemize}

\newrefformat{prob}{Problem\,\ref{#1}}
\newrefformat{def}{Definition\,\ref{#1}}
\newrefformat{sec}{Section\,\ref{#1}}
\newrefformat{sub}{Section\,\ref{#1}}
\newrefformat{prop}{Proposition\,\ref{#1}}
\newrefformat{app}{Appendix\,\ref{#1}}
\newrefformat{alg}{Algorithm\,\ref{#1}}
\newrefformat{cor}{Corollary\,\ref{#1}}
\newrefformat{thm}{Theorem\,\ref{#1}}
\newrefformat{lem}{Lemma\,\ref{#1}}
\newrefformat{fig}{Fig.\,\ref{#1}}
\newrefformat{tab}{Table\,\ref{#1}}

\newtheorem{theorem}{Theorem}
\newtheorem{problem}{Problem}

\newtheorem{lemma}[theorem]{Lemma}
\newtheorem{assumption}[theorem]{Assumption}
\newtheorem{definition}[theorem]{Definition}
\newtheorem{proposition}[theorem]{Proposition}
\newtheorem{remark}[theorem]{Remark}
\newtheorem{example}[theorem]{Example}

\newcommand{\cf}{\emph{cf.}\xspace}

\newcommand{\bdmath}{\begin{dmath}}
\newcommand{\edmath}{\end{dmath}}
\newcommand{\beq}{\begin{equation}}
\newcommand{\eeq}{\end{equation}}
\newcommand{\bdm}{\begin{displaymath}}
\newcommand{\edm}{\end{displaymath}}
\newcommand{\bea}{\begin{eqnarray}}
\newcommand{\eea}{\end{eqnarray}}
\newcommand{\beal}{\beq \begin{array}{ll}}
\newcommand{\eeal}{\end{array} \eeq}
\newcommand{\beas}{\begin{eqnarray*}}
\newcommand{\eeas}{\end{eqnarray*}}
\newcommand{\ba}{\begin{array}}
\newcommand{\ea}{\end{array}}
\newcommand{\bit}{\begin{itemize}}
\newcommand{\eit}{\end{itemize}}
\newcommand{\ben}{\begin{enumerate}}
\newcommand{\een}{\end{enumerate}}

\newcommand{\calI}{{\cal I}}

\newcommand{\calM}{{\cal M}}

\newcommand{\calO}{{\cal O}}

\newcommand{\calX}{{\cal X}}

\newcommand{\etal}{\emph{et~al.}\xspace}
\newcommand{\setal}{~\emph{et~al.}\xspace}
\newcommand{\eg}{\emph{e.g.,}\xspace}
\newcommand{\ie}{\emph{i.e.,}\xspace}
\newcommand{\myParagraph}[1]{{\bf #1.}\xspace}
\newcommand{\M}[1]{{\bm #1}} % Face for matrices
\renewcommand{\boldsymbol}[1]{{\bm #1}}
\newcommand{\LC}[1]{{\color{red} \textbf{LC}: #1}}

\newcommand{\hide}[1]{}

\newcommand{\hiddenText}{{\color{gray} hidden text.}}
\newcommand{\hideWithText}[1]{\hiddenText}

\newcommand{\dist}{\mathbf{dist}}

\DeclareMathOperator*{\argmin}{arg\,min}

\newcommand{\tran}{^{\mathsf{T}}}

\newcommand{\zero}{{\mathbf 0}}

\newcommand{\Real}[1]{ { {\mathbb R}^{#1} } }

\newcommand{\at}[1]{^{(#1)}}

\newcommand{\SOthree}{\ensuremath{\mathrm{SO}(3)}\xspace}

\newcommand{\MA}{\M{A}}
\newcommand{\MB}{\M{B}}

\newcommand{\MP}{\M{P}}

\newcommand{\MU}{\M{U}}
\newcommand{\MR}{\M{R}}

\newcommand{\MOmega}{\M{\Omega}}

\newcommand{\va}{\boldsymbol{a}}

\newcommand{\vd}{\boldsymbol{d}}

\newcommand{\vt}{\boldsymbol{t}}
\newcommand{\vxx}{\boldsymbol{x}} 
\newcommand{\vy}{\boldsymbol{y}}
\newcommand{\vw}{\boldsymbol{w}}

\newcommand{\scenario}[1]{{\smaller \sf#1}\xspace}

\newcommand{\gtwoo}{{\smaller\sf g2o}\xspace}

\newcommand{\garage}{\scenario{garage}}

\newcommand{\blue}[1]{{\color{blue}#1}}

\newcommand{\red}[1]{{\color{red}#1}}

\newcommand{\linkToPdf}[1]{\href{#1}{\blue{(pdf)}}}
\newcommand{\linkToPpt}[1]{\href{#1}{\blue{(ppt)}}}
\newcommand{\linkToCode}[1]{\href{#1}{\blue{(code)}}}
\newcommand{\linkToWeb}[1]{\href{#1}{\blue{(web)}}}
\newcommand{\linkToVideo}[1]{\href{#1}{\blue{(video)}}}
\newcommand{\linkToMedia}[1]{\href{#1}{\blue{(media)}}}
\newcommand{\award}[1]{\xspace} % {{\red{#1}}} % omit awards

\newcommand{\vz}{\boldsymbol{z}}

\newcommand{\meas}{\boldsymbol{z}} % vector

\lstdefinelanguage{mine} 
{morekeywords={while,True,if,break,=,return,function,for,until,in,input,output,assumptions,assumption,invariant,loop,variant,end,invariant,precondition,variables}, 
sensitive=false, 
morecomment=[l]{\#}, 
morecomment=[il]\%{.}, 
morecomment=[s]{/*}{*/}, 
morestring=[b]", 
} 

\usepackage{calc}
\newlength\listingnumberwidth
\newcommand{\arxivVersion}[2]{{#1}\xspace} % #1 to keep arxivVersion; #2 for reduced.
\newcommand{\arxivThisPaper}{Antonante20arxiv-outlierRobustEstimation} % add label of arxiv version of this paper, e.g., Antonante2020TRO-Robust-Estimation

\newcommand{\AutoTunned}{Minimally Tuned\xspace}
\newcommand{\Autotunned}{Minimally tuned\xspace}
\newcommand{\autotunned}{minimally tuned\xspace}
\newcommand{\algatunnedminimally}{{MinT}}
\newcommand{\gncfree}{\scenario{GNC-\algatunnedminimally}}
\newcommand{\adaptfree}{\scenario{ADAPT-\algatunnedminimally}}

\newcommand{\domainoptshort}{{\substack{ \vxx \;\in \;\calX \\ \outset \; \subseteq \; \measSet}}}
\newcommand{\optmshort}[1]{\displaystyle#1_\domainoptshort}
\newcommand{\KwBreak}{\textbf{break}}

\SetNlSty{algLineStyle}{}{}
\SetCommentSty{algCommentStyle}

\renewcommand{\dist}{\text{dist}}

\newcommand{\omitted}[1]{}
\newcommand{\omittedTwo}[1]{}

\newcommand{\myin}{=}

\DeclareMathOperator{\chiSqinv}{chi2inv}
\DeclareMathOperator{\udchinv}{udchi2inv}
\DeclareMathOperator{\fitChi}{Chi2Fit}

\DeclareMathOperator{\separation}{ClustersSeparation}
\DeclareMathOperator{\sort}{sort}

\DeclareMathOperator{\mean}{mean}
\DeclareMathOperator{\kdist}{diam}

\renewcommand{\meas}{\mathcal{M}} %the measurement set
\newcommand{\vres}{\boldsymbol{\res}}% residuals
\newcommand{\weights}{\bm{w}}
\newcommand{\inlierNoise}{\inthr}
\newcommand{\inlierNoiseAdapt}{\varepsilon}
\newcommand{\thetadmts}{\theta}
\newcommand{\last}{\mathrm{end}}
\newcommand{\movstd}{\mathrm{movstd}}

\newcommand{\supp}{\operatorname{supp}}
\newcommand{\iteration}{t}
\newcommand{\myat}[1]{^{(#1)}}
\newcommand{\samplestoconvergeAdapt}{\textit{SamplesToConverg}\xspace}
\newcommand{\samplestoconvergeGNC}{\textit{SamplesToConverg}\xspace}
\newcommand{\maxiter}{\textit{MaxIterations}\xspace}
\newcommand{\gncdisc}{\textit{MuUpdateFactor}\xspace}
\newcommand{\thrdiscount}{\textit{ThrDiscount}\xspace}

\newcommand{\convergthr}{\textit{ConvergThr}\xspace}%{\textit{ConvergThr}\xspace}
\newcommand{\minnoisebound}{\tau}% {\textit{InlierThr}\xspace}
\newcommand{\noiseupbound}{\textit{NoiseUpBnd}\xspace}
\newcommand{\noiselowbound}{\textit{NoiseLowBnd}\xspace}

\newcommand{\windowsize}{\textit{WindowSize}\xspace}
\newcommand{\samplestoconverg}{\textit{MinSamples}\xspace} %{\textit{SamplesToConverg}\xspace}
\newcommand{\urows}{\phi}

\newcommand{\vresyx}[1]{\vres(\vy_{#1}, \vxx)}%{\res_{#1}}
\newcommand{\resyx}[1]{\res(\vy_{#1}, \vxx)}%{\res_{#1}}
\newcommand{\resSqyx}[1]{\res^2(\vy_{#1}, \vxx)}%{\res^2_{#1}}

\newcommand{\limtinf}{{\lim_{\iteration\rightarrow +\infty}}}

\newcommand{\sumAllPoints}[1]{\sum_{i=1}^{#1}}
\newcommand{\sumAllPointsIn}[1]{\sum_{i \;\in \;#1}}

\newcommand{\measn}{m}
\newcommand{\measSet}{\calM}
\newcommand{\inthr}{\epsilon}
\newcommand{\outfreethrintro}{\tau} %_\mts}
\newcommand{\outfreethr}{\tau} % \inthr
\newcommand{\coefftwo}{\nu_2(\outset)} 
\newcommand{\coeffinfty}{\nu_\infty(\outset)} 
\newcommand{\coeff}{\nu_\ell(\outset)} 
 
\newcommand{\lagtls}{\inthr} 
\newcommand{\lag}{\inthr} %\nu}
\newcommand{\inset}{\calI}
\newcommand{\outset}{\calO}

\newcommand{\MminusO}{\measSet\setminus\outset}
\newcommand{\inthrlong}{inlier threshold\xspace}

\newcommand{\barc}{\inthr}
\newcommand{\barcsq}{\inthr^2}

\newcommand{\poly}{\scenario{poly}}
\newcommand{\bptime}{\scenario{BPTIME}}
\newcommand{\np}{\scenario{NP}}
\newcommand{\mle}{\scenario{MLE}}
\newcommand{\ls}{\scenario{LS}}
\newcommand{\tls}{\scenario{TLS}}
\newcommand{\maxcon}{\scenario{MC}}
\newcommand{\maxconLong}{{Maximum Consensus}\xspace}
\newcommand{\ransac}{\scenario{RANSAC}}

\newcommand{\GNClong}{Graduated Non-Convexity\xspace}
\newcommand{\gnc}{\scenario{GNC}}
\newcommand{\GNC}{\scenario{GNC}}

\newcommand{\adapt}{\scenario{ADAPT}}
\newcommand{\adaptmc}{\scenario{ADAPT(MC)}}
\newcommand{\adaptmts}{\scenario{ADAPT(MTS)}}
\newcommand{\greedy}{\scenario{Greedy}}
\newcommand{\greedymc}{\scenario{Greedy(MC)}}
\newcommand{\greedymts}{\scenario{Greedy(MTS)}}

\newcommand{\adaptLong}{Adaptive Trimming\xspace}
\newcommand{\prNameLong}{Generalized Maximum Consensus\xspace}
\newcommand{\prName}{\scenario{G-}\maxcon}
\newcommand{\proneLong}{Generalized Maximum Consensus\xspace}
\newcommand{\prone}{\scenario{G-}\maxcon}
\newcommand{\prtwoLong}{Generalized Truncated Least Squares\xspace}
\newcommand{\prtwo}{\scenario{G-TLS}}
\newcommand{\eqone}{\eqref{eq:main_1}\xspace}
\newcommand{\eqtwo}{\eqref{eq:main_2}\xspace}

\newcommand{\vxxtrue}{\vxx^\circ}
\newcommand{\xtrue}{x^\circ}
\newcommand{\outsettrue}{\outset^\circ}

\newcommand{\optional}[2]{#2\xspace}
\newcommand{\moveToAppendix}[1]{\LC{hidden text can be moved to appendix}}

\newcommand{\mts}{\scenario{MTS}}
\newcommand{\mtsLong}{Minimally Trimmed Squares\xspace}

\newcommand{\name}{\scenario{ADAPT}}

\newcommand{\grSet}{\calM}

\newcommand{\res}{r}
\newcommand{\selSet}{\calO}

\newcommand{\dcs}{\scenario{DCS}}
\newcommand{\PGO}{\scenario{PGO}}
\newcommand{\BnB}{\scenario{BnB}}
\newcommand{\SLAM}{\scenario{SLAM}}

\newcommand{\pcm}{\scenario{PCM}}

\newcommand{\sesync}{\scenario{SE-Sync}}

\newcommand{\MATLAB}{\textsc{MATLAB}\xspace}

\newcommand{\mpPreSpace}{\hspace{-4mm}}
\newcommand{\mpPostSpace}{\vspace{-3mm}}
\newcommand{\mpColTwo}{9.15cm}
\newcommand{\mpMidSpaceTwo}{\hspace{-4mm}}
\newcommand{\mpColThree}{6.0cm}
\newcommand{\mpMidSpaceThree}{\hspace{-3mm}}
\newcommand{\mpColFour}{4.4cm}
\newcommand{\mpMidSpaceFour}{\hspace{-3mm}}
\newcommand{\mpColSix}{2.9cm}
\newcommand{\mpMidSpaceSix}{\hspace{-3mm}}

\newcommand*{\mySpecialfootnotes}[1]{%
  \patchcmd{\@footnotetext}{\floatingpenalty\@MM}{\floatingpenalty#1\relax}%
           {}{\errmessage{Couldn't patch \string\@footnotetext}}%
}

\usepackage{xr}
\graphicspath{{./figures/}}

\usepackage{cite}

\vbadness=\maxdimen
\begin{document}

%!TEX root = ../main.tex

 \title{\fontsize{22}{22} \selectfont Outlier-Robust Estimation: Hardness, \\ \AutoTunned Algorithms, and Applications
 }

 \author{Pasquale Antonante,$^\star$~\IEEEmembership{Student Member,~IEEE,} Vasileios Tzoumas,$^{\star}$~\IEEEmembership{Member,~IEEE,} \\ 
 Heng Yang,~\IEEEmembership{Student Member,~IEEE,} Luca Carlone,~\IEEEmembership{Senior Member,~IEEE} % <-this % stops a space
 \thanks{$^\star$Contributed equally to this work. 
% 	$^\dagger$Corresponding author.
 }
% \thanks{}
 % \thanks{*This work was not supported by any organization}% <-this % stops a space
\thanks{P.~Antonante, H.~Yang, and L.~Carlone are with the Laboratory for Information \& Decision Systems, Massachusetts Institute of Technology, Cambridge, MA 02139, USA. {\tt\footnotesize \{antonap, hankyang, lcarlone\}@mit.edu}}
\thanks{At the time the paper was accepted for publication, V.~Tzoumas was with the Laboratory for Information \& Decision Systems, Massachusetts Institute of Technology, Cambridge, MA 02139, USA. Currently, he is with the Department of Aerospace Engineering, University of Michigan, Ann Arbor, MI 48109, USA. {\tt\footnotesize vtzoumas@umich.edu}
 }%
 \thanks{This work was partially funded by ARL DCIST CRA W911NF-17-2-0181, ONR RAIDER N00014-18-1-2828, MathWorks, NSF CAREER award ``Certifiable Perception for Autonomous Cyber-Physical Systems'', and Lincoln Laboratory's Resilient Perception in Degraded Environments program.}
 }

\maketitle

\arxivVersion{
	\begin{tikzpicture}[overlay, remember picture]
	\path (current page.north east) ++(-4.4,-0.4) node[below left] {
	This paper has been accepted for publication in the IEEE Transactions on Robotics.
	};
	\end{tikzpicture}
	\begin{tikzpicture}[overlay, remember picture]
	\path (current page.north east) ++(-4.85,-0.8) node[below left] {
	Please cite the paper as: P. Antonante, V. Tzoumas, H. Yang, and L. Carlone,
	};
	\end{tikzpicture}
	\begin{tikzpicture}[overlay, remember picture]
	\path (current page.north east) ++(-1.1,-1.2) node[below left] {
	``Outlier-Robust Estimation: Hardness, Minimally Tuned Algorithms, and Applications'', \emph{IEEE Transactions on Robotics (T-RO)}, 2021.
	};
	\end{tikzpicture}
}{}

%!TEX root = ../main.tex

\begin{abstract}
Nonlinear estimation in robotics and vision is typically plagued with outliers due to wrong
data association, or to incorrect detections from signal processing and machine learning methods. 
This paper introduces two
unifying formulations for outlier-robust estimation, \emph{\prNameLong} (\prName) and \emph{\prtwoLong} (\prtwo),
and investigates 
fundamental limits, practical algorithms, and applications.

Our first contribution is a proof that outlier-robust estimation is \emph{inapproximable}: 
in the worst case, 
it is impossible to (even approximately) find the set of outliers, even with 
slower-than-polynomial-time algorithms (particularly, algorithms running in \emph{quasi-polynomial} time). 
{As a second contribution, we review and extend two general-purpose algorithms.}
The first, \emph{\adaptLong} (\adapt), is combinatorial, and is suitable for \prName; the second, 
\emph{\GNClong} (\GNC), is based on homotopy methods, 
and is suitable for \prtwo.
% \toCheck{Both algorithms are deterministic and do not require an initial guess.}
 % Both algorithms are enabled by non-minimal solvers developed in robotics and vision in recent years, and have linear runtime.
We  extend \adapt and \GNC to the case where 
the user does not have prior knowledge 
of the inlier-noise statistics (or the statistics may vary over time) and is unable to guess
a reasonable threshold 
to separate inliers from outliers 
(as the one commonly used in \ransac). We propose the first \emph{\autotunned} algorithms for outlier rejection, 
that dynamically decide how to separate inliers from outliers.  
Our third contribution is an evaluation of the proposed
algorithms  
 on robot perception problems: mesh 
registration, image-based object detection (\emph{shape alignment}), and pose graph optimization. 
\adapt and \GNC execute in real-time, are deterministic, outperform \ransac, and are robust up to $80-90\%$ outliers.
Their  \autotunned versions also compare favorably with the state of the art, even though they do not rely 
on a noise bound for the inliers.
\end{abstract}

%!TEX root = ../main.tex

\newcommand{\introShiftFig}{-2.5mm}
\newcommand{\introFigTitleWidth}{5mm}
\newcommand{\introFigColWidth}{2.5cm}
\newcommand{\introFigSpacing}{\hspace{-4mm}}
\newcommand{\introFigNameSpacing}{\hspace{-4mm}}

\begin{figure}[t!] 
  \vspace{1.7cm}
  \begin{center}
  \hspace*{\introShiftFig}
  \begin{minipage}{\textwidth}
  \begin{tabular}{p{\introFigTitleWidth}p{\introFigColWidth}p{\introFigColWidth}p{\introFigColWidth}}%
  %%%%%%%%%%%%%%%%%%%%%%%%%%%%%%%%%%%%%%%%%%%%%%%%%%%%%%%%%%%%%%%%%%%%%%%%%%%%%
  \begin{minipage}{\introFigTitleWidth}%
		\rotatebox{90}{Mesh Registration\hspace{-1.5cm}}
  \end{minipage}
  & \introFigSpacing
  \begin{minipage}{\introFigColWidth}%
    \vspace{-1.4cm}
    \centering%
    \includegraphics[width=\columnwidth]{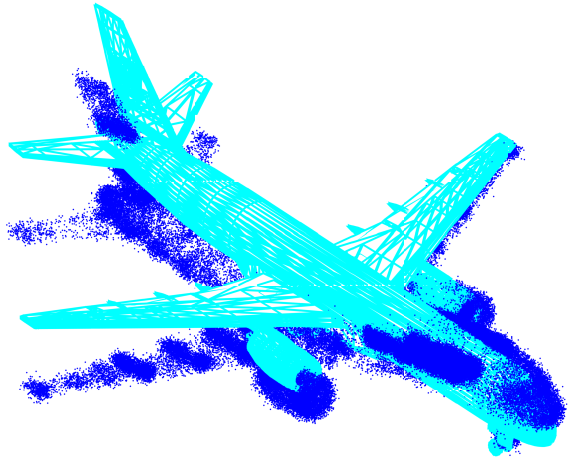}\introFigNameSpacing\\
    (a) \ransac
  \end{minipage}
  & \introFigSpacing
  \begin{minipage}{\introFigColWidth}%
    \vspace{-1.6cm}
    \centering%
    \includegraphics[width=\columnwidth]{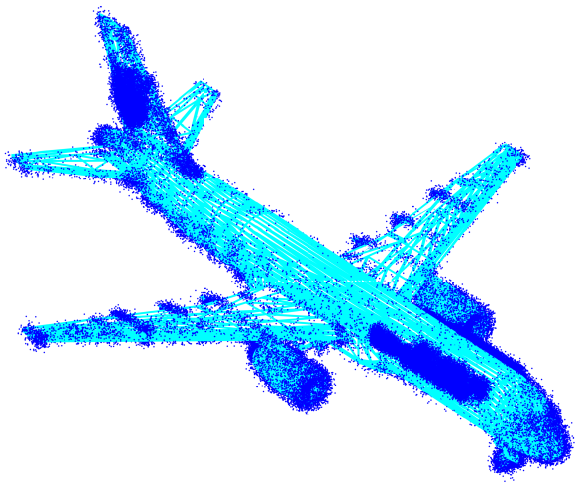}\introFigNameSpacing\\
    (b) \adapt
  \end{minipage}
  & \introFigSpacing
  \begin{minipage}{\introFigColWidth}%
    \vspace{-1.6cm}
    \centering%
    \includegraphics[width=\columnwidth]{overview/mesh_aeroplane_adapt.pdf} \introFigNameSpacing\\
    (c) \adaptfree
  \end{minipage}
  \\[2.3cm] % -----------------------------------------------------------
  \begin{minipage}{\introFigTitleWidth}%
		\rotatebox{90}{Shape Alignment\hspace{-1.8cm}}
  \end{minipage}
  & \introFigSpacing 
  \begin{minipage}{\introFigColWidth}%
    \vspace{-1.4cm}
    \centering%
    \includegraphics[width=\columnwidth]{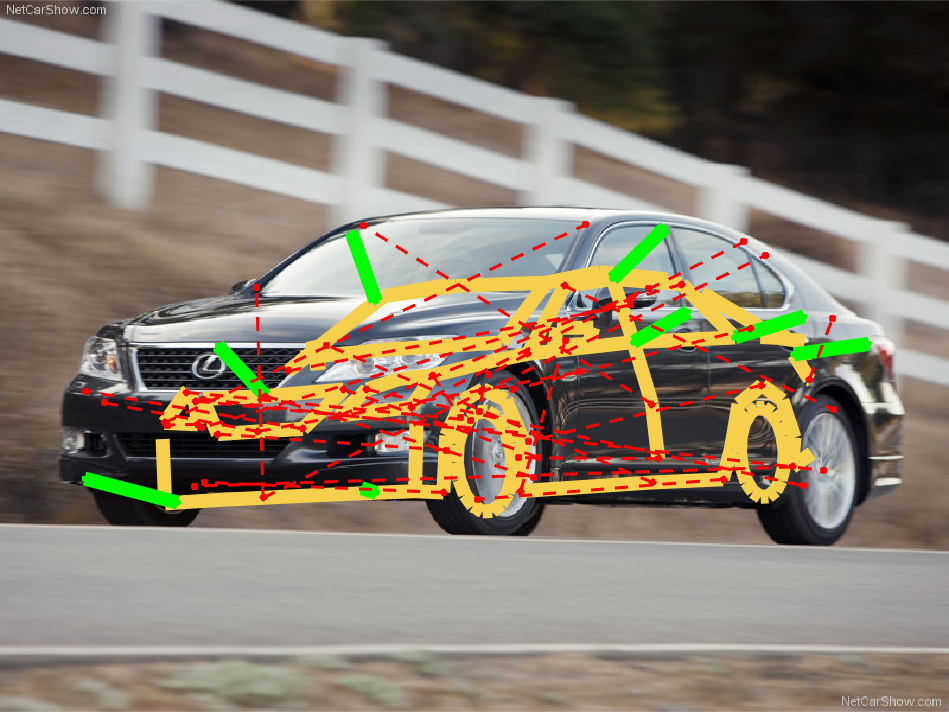}\introFigNameSpacing\\
    (d) \ransac
  \end{minipage}
  & \introFigSpacing 
  \begin{minipage}{\introFigColWidth}%
    \vspace{-1.4cm}
    \centering%
    \includegraphics[width=\columnwidth]{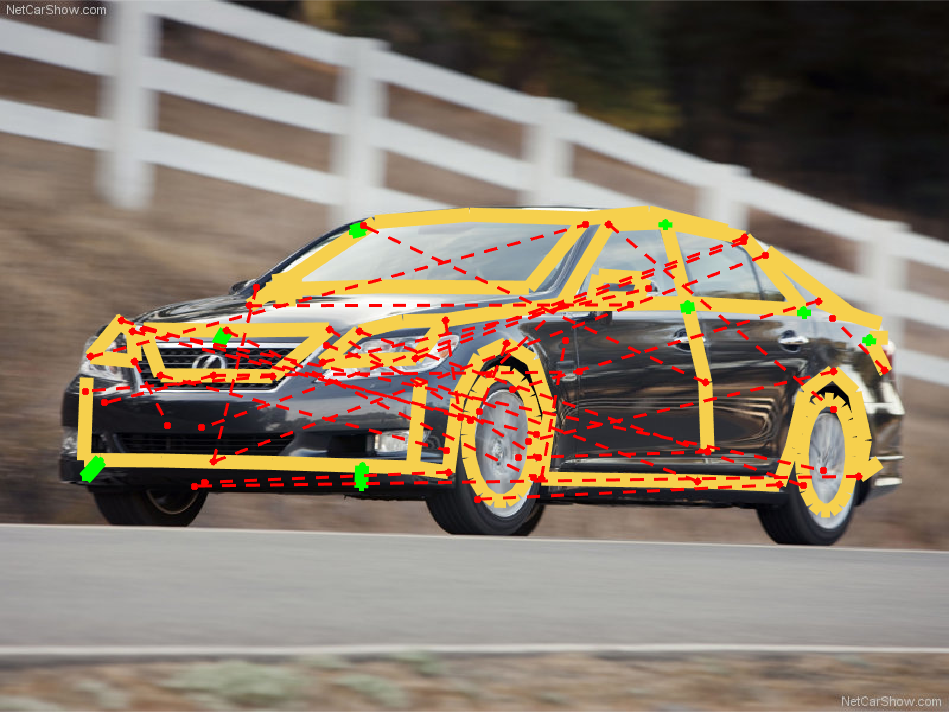}\introFigNameSpacing\\
    (e) \gnc
  \end{minipage}
  & \introFigSpacing
  \begin{minipage}{\introFigColWidth}%
    \vspace{-1.4cm}
    \centering%
    \includegraphics[width=\columnwidth]{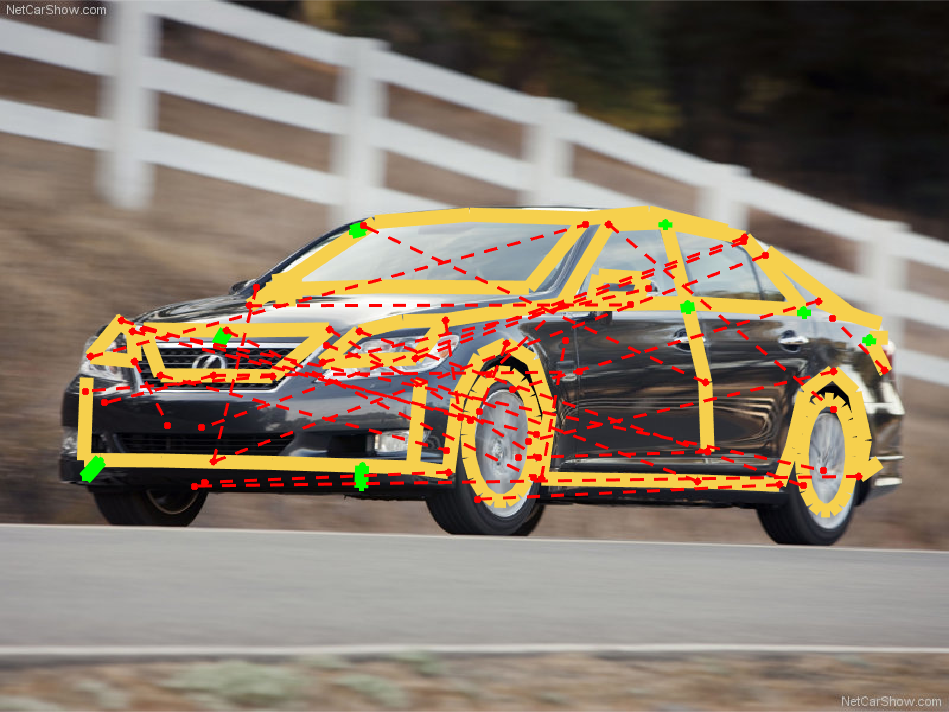} \introFigNameSpacing\\
    (f) \gncfree
  \end{minipage}
  \\[0.6 cm]
  %%%%%%%%%%%%%%%%%%%%%%%%%%%%%%%%%%%%%%%%%%%%%%%%%%%%%%%%%%%%%%%%%%%%%%%%%%%%%%%%%%%%%%%%%%%%%%%%%%%%%%%%%
  \begin{minipage}{\introFigTitleWidth}%
		\rotatebox{90}{Pose Graph\vspace{-5cm}}
  \end{minipage}
  & \introFigSpacing
  \begin{minipage}{\introFigColWidth}%
    \centering%
    \includegraphics[width=0.75\columnwidth]{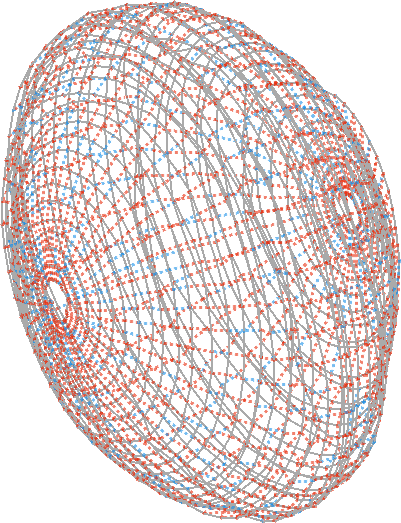} \introFigNameSpacing\\
    (g) \dcs
  \end{minipage}
  & \introFigSpacing
  \begin{minipage}{\introFigColWidth}%
    \centering%
    \includegraphics[width=\columnwidth]{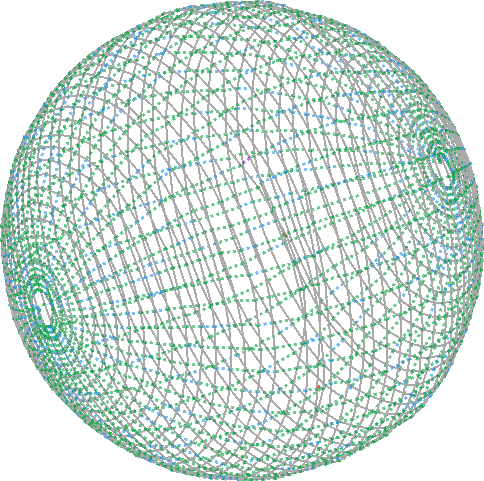} \introFigNameSpacing\\
    (h) \gnc
  \end{minipage}
  & \introFigSpacing
  \begin{minipage}{\introFigColWidth}%
    \centering%
    \includegraphics[width=\columnwidth]{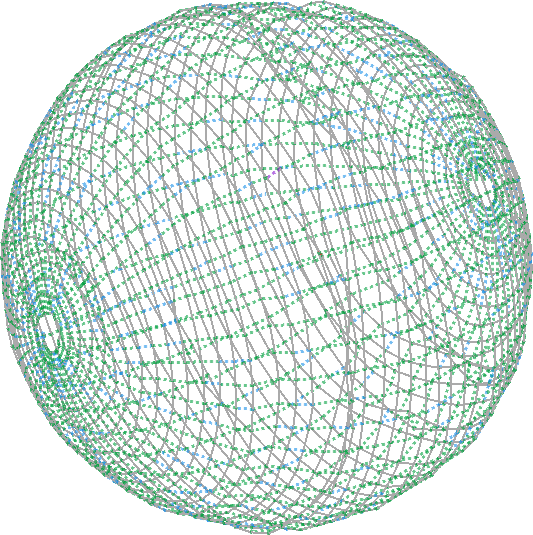} \introFigNameSpacing\\
    (i) \gncfree%
  \end{minipage}
  \end{tabular}
\end{minipage}
% \vspace{-2mm} 
\caption{
We investigate fundamental limits and practical algorithms for outlier-robust estimation.
We discuss two algorithms, 
{\adapt and \GNC,} that outperform the state of the art (\dcs~\cite{Agarwal13icra} and \ransac~\cite{Fischler81} in the figure) in (a-b) mesh registration,
(d-e) shape alignment, and (g-h) pose graph optimization.
Moreover, we propose two variants, \mbox{\adaptfree and \gncfree, that (c,f,i)} perform favorably across robotics applications, 
and do not require parameter \mbox{tuning (\eg kernel size in \dcs, or maximum inlier noise in \ransac).}
\label{fig:introSummary}} 
\end{center}
\end{figure}

%!TEX root = ../main.tex
\vspace{-1em}
\section*{Supplementary Material}

\begin{itemize}
	% \item Example video: \url{https://youtu.be/F0AlzLA0DY8}
	\item Source-code: {\footnotesize \url{https://github.com/MIT-SPARK/GNC-and-ADAPT}}
	\item GTSAM implementation: {\footnotesize \url{https://github.com/borglab/gtsam/blob/develop/gtsam/nonlinear/GncOptimizer.h}}
	\arxivVersion{}{\item ArXiv version: {\footnotesize \url{https://arxiv.org/pdf/2007.15109.pdf}}}
\end{itemize}
%!TEX root = ../main.tex

\section{Introduction}
\label{sec:intro}

\IEEEPARstart{N}{{onlinear}} estimation is a fundamental problem in robotics and computer vision, and
  is the backbone 
of modern perception systems for localization and mapping~\cite{Cadena16tro-SLAMsurvey},  
object  pose estimation~\cite{Yang20tro-teaser,Yang20cvpr-shapeStar}, motion estimation 
and 3D reconstruction~\cite{Choi15cvpr-robustReconstruction,Zhang15icra-vloam}, 
shape analysis~\cite{Maron16tog-PMSDP,Ovsjanikov12TOG-functionalMaps}, virtual and augmented reality~\cite{Klein07ismar-PTAM}, and medical \mbox{imaging~\cite{Audette00mia-surveyMedical}, among others.}

Nonlinear estimation can be formulated as an optimization problem, where one seeks to find the estimate 
that best explains the observed measurements.
A typical perception pipeline includes a \emph{perception front-end} that extracts and matches relevant features 
from raw sensor data (\eg camera images, lidar point clouds). 
These putative feature matches are then passed to a \emph{perception back-end} that uses nonlinear estimation
to compute quantities of interest (\eg the location of the robot, the pose of external objects). 
In the idealized case in which the feature matches are all correct, the back-end can perform estimation 
using a least squares formulation:\footnotemark %allow next-page footnote
%%
%Outlier-free case
\bea\label{eq:leastSquares}
\min_{\vxx \;\in \;\calX} \;\; \sumAllPointsIn{\measSet} \; \resSqyx{i},
\eea
where $\vxx$ is the variable we want to estimate (\eg a 3D pose),
$\calX$ is its domain (\eg the set of poses),
$\measSet$ is the set of given measurements (\eg pixel observations of points belonging to the object),
$\vy_i$ is the $i$-th measurement ($i \in \measSet$), 
and $\resyx{i} \geq 0$ is the \emph{residual} error of the $i$-th measurement, which
quantifies how well a given $\vxx$ fits a measurement $\vy_i$, 
{(\eg $\resyx{i}=|y_i-\mathbf{a}_i\tran \vxx|$ for the linear, scalar measurement case).}
The problem in eq.~\eqref{eq:leastSquares} produces a maximum-likelihood estimate when the measurement noise is Gaussian, see \eg~\cite{Cadena16tro-SLAMsurvey}. 
 However, despite its apparent simplicity, it is already hard to solve globally, since the cost function in~\eqref{eq:leastSquares} 
 and the domain $\calX$ are typically non-convex in robotics applications~\cite{Yang20cvpr-shapeStar,Rosen18ijrr-sesync}.

The development of perception systems that can work in challenging real-world conditions requires the 
design of outlier-robust estimation methods.
 Perception front-ends are typically based on image or signal processing methods~\cite{Lowe99iccv} or 
 learning methods~\cite{gojcic19cvpr-3Dsmoothnet} for feature detection and matching. 
 These methods are prone to produce incorrect matches, which result in completely wrong measurements $\vy_i$ in eq.~\eqref{eq:leastSquares}
 and compromise the accuracy of the solution returned by eq.~\eqref{eq:leastSquares}. 
Computing robust estimates in the face of these \emph{outliers} 
has been a central topic in computer vision and robotics.

For low-dimensional estimation problems, \eg object pose estimation from images or point clouds,
 researchers have often resorted to combinatorial formulations for outlier rejection. 
In particular, a popular formulation is based on \emph{consensus maximization}~\cite{Chin18eccv-robustFitting,Chin17slcv-maximumConsensusAdvances}, which looks for an estimate maximizing the number of measurements explained within a given \emph{inlier threshold} $\inthr$ (or equivalently, minimizes the number of outliers):\!

\footnotetext{We use lowercase characters, \eg $x$, to denote real scalars or functions, bold lowercase characters, \eg $\vxx$, for real vectors, 
	bold uppercase characters, \eg $\MA$ for real matrices,
	and calligraphic uppercase characters, \eg $\measSet$, for sets (either discrete or continuous); as exceptions, 
	we use the standard notation $\mathbb{R}$ to denote the set of real numbers, and $\mathbb{N}$ to denote the set of non-negative integers.
	$|x|$ is the absolute value of $x$, and $|\measSet|$ is the cardinality of $\measSet$.
	If $\vxx = [x_1,x_2,\ldots,x_n]$, then $\|\vxx\|_1\;\triangleq\sumAllPoints{n}|x_i|$ is $\vxx$'s Manhattan norm; $\|\vxx\|_2\;\triangleq\sqrt{\sumAllPoints{n} x_i^2}$ is $\vxx$'s Euclidean norm; and $\|\vxx\|_{\infty}\;\triangleq\max\{|x_1|,|x_2|,\ldots,|x_n|\}$ is $\vxx$'s infinity norm.
}%

\vspace{-4mm}
\beq\label{eq:maxConsensus}
\optmshort{\min}\quad |\outset| \;\;\;\emph{\text{\emph{s.t.}}}\;\;\;  
\resyx{i}   \leq \inthr, \quad \forall i \in \measSet \setminus \outset.
\eeq
{The \emph{\maxconLong} (\maxcon) problem in eq.~\eqref{eq:maxConsensus} is then solved using} \ransac \cite{Fischler81}  or 
branch-and-bound (\BnB)~\cite{Chin17slcv-maximumConsensusAdvances}. However, \ransac is non-deterministic,
requires a minimal solver, and is limited to problems where 
 the estimate can be computed from a small number of measurements~\cite{Bustos18pami-GORE}.
 Similarly, \BnB 
 runs in the worst-case in exponential {time and does not scale to large problems.}

For high-dimensional estimation problems, \eg bundle adjustment and \SLAM,  
researchers have more heavily relied on \emph{M-estimation} to gain robustness against outliers~\cite{Huber81}.
  M-estimation replaces the least-squares cost in~\eqref{eq:leastSquares} with a function $\rho$
  that is less sensitive to measurements with large residuals:
\bea\label{eq:robustEstimation}
\min_{\vxx \;\in \;\calX} \quad \sumAllPointsIn{\measSet} \; \rho(\resyx{i}, \inthr),
\eea
where, for instance, $\rho$ can be a Huber, Cauchy, or  Geman-McClure cost~\cite{Black96ijcv-unification}.   
The M-estimation problem
 in eq.~\eqref{eq:robustEstimation} 
 has the advantage of leading to a continuous (rather than combinatorial as 
in \maxcon) optimization problem, which however is still hard to solve globally due to the typical non-convexity of the 
cost function and constraint set $\calX$.   
Typical robotics applications use iterative local solvers to minimize~\eqref{eq:robustEstimation}, see~\cite{Kuemmerle11icra,Sunderhauf12icra,Agarwal13icra}. However, local solvers require a good initial guess (often unavailable in applications) 
and are easily trapped in local minima corresponding to poor estimates.

All in all, the literature is currently lacking an approach that simultaneously satisfies the following design constraints:
(i) is fast and scales to large problems, 
(ii) is deterministic, 
(iii) can operate without requiring an initial guess. This gap in the literature is the root cause
for the brittleness of modern perception systems and is limiting the use of perception in safety-critical applications, 
from self-driving cars~\cite{Shalev-Shwartz17arxiv-safeDriving}, to autonomous aircrafts~\cite{EASADesignNN}, and spacecrafts~\cite{Chen19ICCVW-satellitePoseEstimation}. 

An additional limitation of state-of-the-art robust estimation algorithms is that they require 
knowledge of the 
expected (inlier) measurement noise. This knowledge is encoded in the parameter $\inthr$ in both 
eq.~\eqref{eq:maxConsensus} and eq.~\eqref{eq:robustEstimation}. 
However, in many problems, characterizing this parameter is time-consuming 
(\eg it requires collecting data in a controlled environment to compute statistics) 
or is based on trial-and-error (\ie requires manual parameter tuning by a human expert). 
Also, this approach is not suitable for long-term operation: imagine a ground robot performing life-long SLAM; after months of operations, the noise statistics may vary (\eg a flat tire leads to increased odometry noise), making the 
factory calibration~unusable.

\myParagraph{Contributions} 
This paper fills these gaps by understanding fundamental computational limits of robust estimation, 
and by designing outlier-robust estimation algorithms that:
(i) are \emph{general-purpose} and usable across many estimation problems, 
(ii) scale to large problems with thousands of variables, 
(iii) are deterministic,
(iv) do not rely on an initial estimate, 
and (v) can potentially work without manual fine-tuning and be resilient to changes in the measurement noise statistics.
We achieve these goals through three key contributions.

\myParagraph{1. General Formulations and Inapproximability} 
Section~\ref{sec:formulation} introduces two 
unifying formulations for outlier-robust estimation, \emph{\prNameLong} (\prName) and \emph{\prtwoLong} (\prtwo).
\prName is a combinatorial formulation and generalizes the popular \maxcon in eq.~\eqref{eq:maxConsensus} and the recently proposed \emph{\mtsLong} (\mts)~\cite{Tzoumas19iros-outliers};
 \prtwo is a continuous-optimization formulation and generalizes the truncated least squares used in M-estimation.
We provide probabilistic interpretations for both formulations:
 \prone 
solves a likelihood-constrained estimation problem, 
while \prtwo is a maximum likelihood estimator.
% looks for an $\vxx$ maximizing the likelihood of a mixture distribution of inliers and outliers\optional{;
%  this implies that \tls is a maximum likelihood estimator when the inliers follow a normal distribution and 
%  the outliers have a constant error.}{.} 

We also provide 
necessary and sufficient 
conditions for \prone and \prtwo to {return the same solution}.  
{We demonstrate that, in general, the conditions {may not be} satisfied, and \prtwo 
 \xspace{can} 
reject more measurements than \prone.  
Notwithstanding, 
we provide counterexamples showing that, while \prtwo 
 \xspace{can reject} 
more measurements, it may lead to more accurate estimates.}

Section~\ref{sec:hardness} proves that \prone and \prtwo are inapproximable even by \emph{quasi-polynomial} algorithms, which are slower than polynomial.\footnote{An algorithm is called \emph{quasi-polynomial}, if its runtime is $k_12^{( \log \measn)^{k_2}}$\!, where $k_1$ and $k_2$ are constants, and $\measn$ is the algorithm's input size.
Evidently, quasi-polynomial time algorithms are slower than polynomial, since $k_1\; 2^{(\log \measn)^{k_2}}\; >\; k_1\;2^{k_2\log \measn} \;=\;k_1 \measn^{k_1}$.} The result holds true subject to a believed conjecture in complexity theory, $\np\; {\notin}\;\bptime({\measn}^{\poly\log \measn})$, which is similar to the widely known $\np\neq \scenario{P}$.\footnote{$\np \notin\bptime(\measn^{\poly\log \measn})$ means there exists no randomized algorithm that outputs solutions to problems in $\np$ with probability $2/3$, after running for $O(\measn^{(\log \measn)^k})$ time, for an input size $m$ and a constant $k$~\cite{Arora09book-complexity}.}  
The result captures the hardness of \prone and \prtwo for the first time: even in simplified cases where one knows the number of outliers to reject or that the residual error for the inliers is zero, it is still impossible to compute an approximate solution for   
\prone and \prtwo within a prescribed approximation bound.
This result strengthens recent inapproximability results for \maxcon that only rule out polynomial time algorithms~\cite{Chin18eccv-robustFitting}.

\myParagraph{2. General-purpose and \AutoTunned Algorithms} 
Our second contribution is to discuss two general-purpose algorithms.
Section~\ref{sec:adaptIntro} presents a combinatorial approach, named \emph{\adaptLong} (\adapt), which is suitable for \prName.
The algorithm works by iteratively removing measurements with large errors, but contrarily to a naive greedy algorithm, 
{it
revisits past decisions and checks for convergence over a sequence of iterations,} leading to more accurate estimates.
Section~\ref{sec:GNC} briefly reviews the \emph{\GNClong} (\GNC) approach of~\cite{Yang20ral-GNC}, which 
is a continuous-optimization approach to solve \prtwo. 
Both algorithms have linear runtime, are deterministic, and do not require an initial guess.
 \omitted{Moreover, they both leverage non-minimal solvers recently developed in robotics and vision,
  which circumvent the need for an initial guess.}

Section~\ref{sec:parameterFree} presents the first outlier-robust estimation algorithms that are able to 
automatically adjust their parameters to perform robust estimation without prior knowledge of the inlier noise statistics.
We present two algorithms, \adaptfree and \gncfree, where \scenario{\algatunnedminimally} stands for ``{\AutoTunned}'', that automatically 
adjust the noise bounds in \adapt and \gnc  without the need for manual fine-tuning.
This is in stark contrast with the techniques in the literature, whose correct operation relies on the knowledge of 
the maximum inlier noise (\cf~parameter $\inthr$ in eq.~\eqref{eq:maxConsensus} and eq.~\eqref{eq:robustEstimation}).
 We call these algorithms ``\AutoTunned'' (rather than Parameter-Free) since they still involve parameters, which 
 however only depend on the type of 
 application, 
 rather than the problem instance 
 (\eg the same parameter values are used to solve any \SLAM~problem). 

\myParagraph{3. Experiments in Robotics and Vision Problems} 
Section~\ref{sec:experiments} evaluates 
the proposed algorithms  
across three robot perception problems: mesh 
registration, image-based object detection (\emph{shape alignment}), and pose graph optimization. 
In mesh registration and shape alignment, 
\adapt and \GNC  execute in real-time, outperform \ransac, and are robust up to {$80\%$ outliers}.
In pose graph optimization, \adapt and \GNC outperform local optimization~\cite{Agarwal13icra}
and ad-hoc techniques~\cite{Mangelson18icra}, and are robust up to {$90\%$ outliers}.
{Their  \autotunned versions compare favorably with the state of the art, exhibiting similar performance as \adapt and \GNC, 
yet
without requiring knowledge of a noise bound for the inliers.}

\myParagraph{Novelty with respect to Previous Work~\cite{Tzoumas19iros-outliers,Yang20ral-GNC}}
 This paper extends \adapt and the hardness result (presented in~\cite{Tzoumas19iros-outliers}) as well as 
 \gnc (presented in~\cite{Yang20ral-GNC}) in several directions. 
 The \prName and \prtwo
 formulations are novel and generalize the formulations in~\cite{Tzoumas19iros-outliers,Yang20ral-GNC}.
 The probabilistic justification of these formulations and their relations have not been published before.
We streamline \adapt to apply to both \maxcon and \mts. 
{We provide a local convergence proof for \gnc.}
 Moreover, the \autotunned algorithms, \adaptfree and \gncfree, are completely novel. 
 We present a more extensive experimental evaluation, including 3D SLAM problems (not considered in~\cite{Tzoumas19iros-outliers,Yang20ral-GNC}).
Finally, this paper includes a more comprehensive discussion ---\eg why using a greedy algorithm is ineffective for outlier rejection (Section~\ref{sec:greedy}),\arxivVersion{ limitations of the proposed algorithms (\ref{sec:limitations}),}{{}} and an extended literature review (Section~\ref{sec:relatedWork})--- and 
 provides complete proofs of the technical results in the appendix.  \arxivVersion{{}}{Limitations of the proposed algorithms are discussed in the extended draft~\cite[Appendix~1]{\arxivThisPaper}.} 

%!TEX root = ../main.tex

\section{Outlier-Robust Estimation:\\ Generalized \maxcon and \tls  Formulations}
\label{sec:formulation}

Sections~\ref{subsec:prone} and~\ref{subsec:prtwo}
 present {\proneLong} (\prone) and {\prtwoLong} (\prtwo). 
Section~\ref{subsec:mle_interpretation} gives \prone's and \prtwo's probabilistic justification
(Propositions~\ref{th:mle_for_mc}-\ref{th:mle_for_tls}).
Section~\ref{subsec:relationship}
provides conditions for \prone and \prtwo to be {equivalent} (Theorem~\ref{th:relationship}). 
\optional{Finally, we demonstrate the conditions can be violated, with \prtwo rejecting more measurements than \prone  as outliers (Example~\ref{ex:tls_vs_mts} in Section~\ref{subsec:relationship}). Therein,  we also show \prtwo may lead to more accurate estimates than \prone, even when \prtwo rejects more measurements as outliers.}{}

We use the following notation:
\begin{itemize}
	\item $\vxxtrue$ is the true value of $\vxx$ we want to estimate;
	\item $\outsettrue$ is the true set of outliers;
	\item $\vresyx{\inset}\triangleq \left[ \resyx{i}\right]_{i\in\inset}$, for any $\inset\subseteq \measSet$; \ie $\vresyx{\inset}$ is the vector of residuals for the measurements $i \in \inset$.
\end{itemize}

\subsection{\proneLong \emph{(\prone)}}\label{subsec:prone}
We present a generalized maximum consensus formulation.

\begin{problem}[\proneLong (\prone)]\label{pr:main_1}  
Find an estimate $\vxx$ by solving the combinatorial problem
\beq 
\optmshort{\min} \quad |\outset| \;\;\;\text{\emph{s.t.}}\;\;\;  \|\; \vresyx{\measSet\setminus \outset} \;\|_\ell \;\leq \outfreethr, \label{eq:main_1}
\tag{{\smaller{G-MC}}}
\eeq
where $\outfreethr \geq 0$ is 
a \emph{given} \inthrlong, 
and $\|\;\cdot\;\|_\ell$ denotes a generic vector norm \emph{(}in this paper, $\ell \in \{2, \infty\}$\emph{)}.
\end{problem}

Since the true number of outliers is unknown, 
\prone rejects the least amount of measurements such that the remaining ones appear as inliers.
In \eqone, $\outset$ is the set of
measurements classified as outliers; 
 correspondingly, $\measSet\setminus \outset$ is the set of
 inliers.
A choice of inliers $\measSet\setminus \outset$ is feasible only if there exists an $\vxx$ such that the cumulative residual error $\|\;\vresyx{\measSet\setminus \outset} \;\|_\ell$ satisfies the \inthrlong $\outfreethr$ (enforced by the constraint).

\prone generalizes existing combinatorial outlier-rejection formulations.
In particular, depending on the choice of $\ell$ and $\outfreethr$ in \eqone, \prone is equivalent to \maxconLong (\maxcon) 
or \mtsLong (\mts):

\begin{itemize}
\item \myParagraph{\maxcon as \prone} If $\ell=+\infty$ and $\outfreethr = \inthr$ in \eqone, then \prone is equivalent to \maxcon (eq.~\eqref{eq:maxConsensus}), since $\|\; \vresyx{\measSet\setminus \outset} \;\|_\infty\;\leq \inthr^2$ implies $\resyx{i} \leq \inthr$, $\forall \; i\in \MminusO$.  

\item \myParagraph{\mts as \prone} If $\ell=2$ in \eqone, then \prone is equivalent to the \mts~\cite{Rousseeuw87book,Tzoumas19iros-outliers} formulation
 \bea\label{eq:mts}
\optmshort{\min}\quad |\outset|  \;\;\;\text{s.t.}\;\;\; \textstyle\sumAllPointsIn{\measSet\setminus \outset} \; \resSqyx{i} \leq \outfreethr^2,
\eea
since $\|\; \vresyx{\measSet\setminus \outset} \;\|_2^2\; = \sumAllPointsIn{\measSet\setminus \outset} \; \resSqyx{i}$.
\end{itemize}

%%%%%%%%%%%%%%%%%%%%%%%%%%%%%%%%%%%%%%%%%%%%%%%%%%%%%%%%%%%%%%%%%%%%%%%%%%%%%%%%
\subsection{\prtwoLong \emph{(\prtwo)}} \label{subsec:prtwo}

We present a second formulation that generalizes truncated least squares in M-estimation~\cite{Lajoie19ral-DCGM,Yang19iccv-QUASAR}.

\begin{problem}[\prtwoLong (\prtwo)]\label{pr:main_2}  
Find an estimate $\vxx$ by solving the program
\beq 
\optmshort{\min} \quad  \coeff \|\; \vresyx{\measSet\setminus \outset} \;\|_\ell^2 \; + \; \lag^2 |\outset|, \label{eq:main_2}
\tag{{\smaller{G-TLS}}}
\eeq
where $\|\;\cdot\;\|_\ell$ denotes a generic vector norm \emph{(}in this paper, $\ell \in \{2, {\infty}\}$\emph{)},
and $\coeff, \lag > 0$ are \emph{given} penalty coefficients; { in particular, $\coefftwo=1$ and $\coeffinfty= |\MminusO|$. }
\end{problem}

\prtwo looks for an outlier-robust estimate $\vxx$ by separating the measurements into inliers and outliers such that the former are penalized with their weighted cumulative residual error $\coeff \|\; \vresyx{\measSet\setminus \outset} \;\|_\ell^2$, and the latter with their weighted cardinality $\lag^2 |\outset|$.  
\optional{Intuitively, if the contribution of  measurement $i$ to the cumulative residual error is more than $\lag$, then the measurement will be classified as an outlier (instead of an inlier).  Evidently, if $\lag \leq 0$, then \prtwo classifies all measurements as outliers, and, trivially, $\|\; \vresyx{\measSet\setminus \outset} \;\|_\ell^2 = 0$; therefore, we generally set $\lag > 0$ in \eqtwo.  Moreover,}{}
For appropriate choices of $\lag$, \prtwo reduces to Truncated Least Squares (\tls) or standard least squares~(\ls):

\begin{itemize}
\item \myParagraph{\tls as \prtwo}  If $\ell=2$, then \prtwo becomes
\beq 
\optmshort{\min} \quad   \sumAllPointsIn{\MminusO}\resSqyx{i} \; + \; \sumAllPointsIn{\outset}\inthr^2 , \label{eq:main_2'}
\eeq
which is equivalent to the \tls formulation~\cite{Lajoie19ral-DCGM,Yang20tro-teaser,Yang19iccv-QUASAR}, commonly written using auxiliary binary variables $w_i$ as\!
\begin{equation}\label{eq:tls}
\hspace{-5mm}\min_{\vxx \;\in \;\calX}\  \sumAllPointsIn{\meas} \min_{w_i \;\in\; \{0,1\}} \;\left[w_i\;\resSqyx{i}  +  (1-w_i)\;\inthr^2\right].
\tag{{\smaller TLS}}
\end{equation}

\item \myParagraph{\ls as \prtwo}  If $\ell=2$ and $\lag=+\infty$, then, \prtwo becomes the least squares formulation in eq.~\eqref{eq:leastSquares}.
\end{itemize}

\subsection{Probabilistic Justification of \prone and \prtwo}\label{subsec:mle_interpretation}

We provide a probabilistic justification for the \prone and \prtwo formulations{, under the standard assumption of independent noise across the measurements.}

\omitted{Our derivations in the rest of this section takes the standard assumption of independent measurement noise.}

\begin{assumption}[Independent Noise]\label{ass:independence}
If $i\neq j$, for any $i,j \!\in\! \measSet$, then $\resyx{i}$ and $\resyx{j}$ are independent random variables.
\end{assumption}

\omitted{Assumption~\ref{ass:independence} implies the joint probability density of the residuals equals the product of each residual's probability density; more formally, for any $\inset\subseteq \measSet$,
\begin{equation*}
d(\vresyx{\inset})=\prod_{i\; \in \; \inset} \;d(\resyx{i}),
\end{equation*}
where $d(\cdot)$ denotes a probability density.}

The results below provide a probabilistic grounding for two \prone's instances, 
\maxconLong (\maxcon) and \mtsLong (\mts), via likelihood estimation.

\begin{proposition}[Uniform Inlier Distribution Leads to \maxcon]\label{th:mle_for_mc}
If $\res(\vy_{i}, \vxxtrue)$ is uniformly distributed with support  $[0,\inthr)$ for any $i \in \measSet \setminus \outsettrue$,  then \maxcon in eq.~\eqref{eq:maxConsensus} is equivalent to
\begin{equation}\label{eq:mle_for_mc}
\optmshort{\min}\quad |\outset| \;\;\;\emph{\text{s.t.}}\;\;\prod_{i\; \in \; \MminusO} u(\resyx{i},\inthr)\; > 0,
\end{equation} 
where the inequality is strict, and $u(\res,\inthr)$ is the probability density function of the uniform distribution with support $[0,\inthr)$.
\end{proposition}

The optimization in eq.~\eqref{eq:mle_for_mc} is a \emph{likelihood-constrained} estimation:
eq.~\eqref{eq:mle_for_mc} finds an $\vxx$ such that the joint likelihood of the inliers is greater than zero.
\optional{\red{either the probability density equals $1/\inthr^{|\MminusO|}$, when all measurements in $\MminusO$ appear as inliers (\ie when $\resyx{i} \leq \inthr$ for all $i \in \MminusO$), or it equals $0$, otherwise.}}{} 
\optional{Overall, eq.~\eqref{eq:mle_for_mc} is an \mle formulation, and \maxcon is equivalent to it.}{}

\begin{proposition}[Normal Inlier Distribution Leads to \mts]\label{th:mle_for_mts}
If $\res(\vy_{i}, \vxxtrue)$ follows a Normal distribution for any $i \in \measSet \setminus \outsettrue$, then \mts in 
eq.~\eqref{eq:mts} is  equivalent to
\begin{equation}\label{eq:mle_for_mts}
\optmshort{\min} \quad |\outset| \;\;\;\emph{\text{s.t.}}\;\;\prod_{i\; \in \; \MminusO} g(\resyx{i}) \geq 
\frac{e^{-\frac{\outfreethrintro^2}{2}}}{(\pi/2)^{\frac{|\MminusO|}{2}}},
\end{equation}
where {$g(\res)\triangleq \sqrt{2/\pi}\; \exp(-\res^2/2)$} is the density of a Normal distribution constrained to the non-negative axis $(\res\geq 0)$.
\end{proposition}

Proposition~\ref{th:mle_for_mts} implies \mts is equivalent to a likelihood-constrained estimation, where the inliers follow a Normal distribution (in contrast to Proposition~\ref{th:mle_for_mc} where the inliers are uniformly distributed).  
\omitted{In other words, \mts looks for an estimate with a large set of inliers that best fit a folded Normal distribution.} 
\optional{The equivalence is evident once we consider the $-\log(\cdot)$ of both the sides of  the inequality in eq.~\eqref{eq:mle_for_mts}.} 

Similarly, we show that an instance of \prtwo, Truncated Least Squares (\tls), 
can be interpreted as a maximum likelihood estimator.  Particularly, {if the number of outliers is known, we show \tls 
selects a set of inliers
and a maximum likelihood estimate assuming the inliers are Normally distributed
(Proposition~\ref{th:mle_for_tls}); and if the number of outliers is unknown, we provide a broader characterization by connecting \tls to a max-mixture of Normal and uniform distributions
 (Proposition~\ref{th:mle_for_tls-normalAndUni}).}

{\begin{proposition}[Normal Distribution and Known Number of Outliers Lead to \tls]\label{th:mle_for_tls}
	Assume $\res(\vy_{i}, \vxxtrue)< \inthr$ for any $i \in \measSet \setminus \outsettrue$ and $|\outsettrue|$ is known.  If $\res(\vy_{i}, \vxxtrue)$ is Normally distributed for each $i \in \MminusO$, then \tls is equivalent to 
	the cardinality-constrained maximum likelihood estimator
	\begin{equation}\label{eq:mle_for_tls}
	\max_{\substack{ \vxx \;\in \;\calX \\ \outset \; \subseteq \; \measSet,\; |\outset|\;=\;|\outsettrue|}}\quad\prod_{i\;\in\;\MminusO} \; g(\resyx{i}).
	\end{equation}
\end{proposition}}

\begin{proposition}[Normal with Uniform Tails Leads to \tls]\label{th:mle_for_tls-normalAndUni}
For any $i \in \measSet$, 
assume 
(i) $\res(\vy_{i}, \vxxtrue) \leq \alpha$ for some number $\alpha$, and 
(ii) $\res(\vy_{i}, \vxxtrue)$
follows a \emph{modified} Normal distribution $\hat{g}(\res)$ where the tail of the Normal distribution for $\res \geq \inthr$ is replaced with a uniform distribution with support $[\inthr,\alpha]$;
particularly,
\beq\label{eq:normalWithUniformTail}
\hat{g}(\res) =\frac{1}{\beta}\left\{\begin{array}{ll}
g(\res), & \res < \inthr;\\
g(\inthr), & \res \in [\inthr,\alpha];\\
0, & \text{otherwise},
\end{array}\right.
\eeq
where $\beta$ is a normalization factor (that depends on $\alpha$) such that $\hat{g}(\cdot)$ is a valid probability density ($\int_{0}^{\alpha}\; \hat{g}(r)\; dr =1$). Then, \tls is equivalent to 
the maximum likelihood estimator
\begin{equation}\label{eq:mle_for_tls-normalAndUni}
\max_{\vxx\; \in\; \calX}\quad\prod_{i\;\in\;\meas} \; \hat{g}(\res(\vy_{i}, \vxx)).
\end{equation}
\end{proposition}

The interested reader can find an alternative probabilistic interpretation of \tls in \arxivVersion{\ref{app:mle_for_tls-weibull}}{\cite[Appendix~3]{\arxivThisPaper}}, 
where \tls is shown to minimize the probability that an estimate becomes inaccurate when measurements are misclassified as inliers instead of outliers, and vice versa.  We describe this probability 
%in \arxivVersion{\ref{app:mle_for_tls-weibull}}{\cite[Appendix~3]{\arxivThisPaper}} 
with a product of Weibull distributions.

%%%%%%%%%%%%%%%%%%%%%%%%%%%%%%%%%%%%%%%%%%%%%%%%%%%%%%%%%%%%%%%%%%%%%%%%%%%%%%%%%%%%%%%%%%%%%%%%%%%%%%%%
\subsection{Relationship Between \prone and \prtwo}\label{subsec:relationship}

{\begin{theorem}[\prone = \prtwo when $\ell=+\infty$]\label{th:relationship_linf}
Choose $\|\;\cdot\;\|_\ell$ to be the infinity norm in \prone and \prtwo,  and $\outfreethr = \inthr$ in \prone.
Also, assume \prone has an optimal solution $(\vxx_\prone,\outset_\prone)$  such that $\|\; \vres(\vy_{\meas\setminus \outset_\prone},\vxx_\prone)\;\|_\infty\; < \epsilon$
\emph{(}\ie \xspace\prone's inequality constraint is strict at an optimal solution\emph{)}.
Then, \prone and \prtwo compute the same set of outliers.
\end{theorem}

The inequality $\|\; \vres(\vy_{\meas\setminus \outset_\prone},\vxx_\prone)\;\|_\infty\; \leq \epsilon$ is strict with probability $1$ when the measurements are random. Hence, \prone = \prtwo  with probability $1$ when $\ell=+\infty$, and, thus, we henceforth focus only on the \tls instance of \prtwo ($\ell=2$).
}

\begin{theorem}[\prone $\neq$ \prtwo when $\ell=2$]\label{th:relationship}
\mbox{Denote by:}
\begin{itemize}
	\item $(\vxx_\mts, \outset_\mts)$ an optimal solution to \mts (\prone's instantiation for $\ell=2$);
	\item $(\vxx_\tls, \outset_\tls)$  an optimal solution to \tls (\prtwo's instantiation for $\ell=2$ and $\coeff=1$);	
	\item  $r^2_\tls(\lagtls)$
	the error of the measurements classified as inliers at $(\vxx_\tls, \outset_\tls)$: $r^2_\tls(\lagtls)\triangleq \|\;\res(\vy_{\measSet\setminus \outset_\tls}, \vxx_\tls)\;\|_\ell^2$;
	\item {$f_\tls(\lagtls)$ the value of \tls: $f_\tls(\lagtls)\triangleq r^2_\tls(\lagtls)+\lagtls^2 |\outset_\tls|$.}
\end{itemize}
\smallskip
Then, for any $\lagtls >0$,
\begin{itemize}
	\item if $\outfreethr = r_\tls(\lagtls)$, then  $|\outset_\tls|\; = |\outset_\mts|$, and, in particular, $(\vxx_\tls, \outset_\tls)$ is also a solution to \mts; 
	\item if $\outfreethr > r_\tls(\lagtls)$, then $|\outset_\tls|\;\geq |\outset_\mts|$;
	\item if $\outfreethr < r_\tls(\lagtls)$, then $|\outset_\tls|\; < |\outset_\mts|$, {and \mts and \tls compute different sets of outliers.}
\end{itemize}
\end{theorem}

Example~\ref{ex:tls_vs_mts} below elucidates the result in Theorem~\ref{th:relationship} by considering  instantiations of \mts and \tls
in a toy example.
The example shows that although {\tls} 
 {may reject} more measurements than {\mts}, 
\tls can lead to more accurate estimates of $\vxxtrue$ since it tends to reject ``biased'' measurements.

\begin{example}[Sometimes, Less is More]\label{ex:tls_vs_mts} Consider an estimation problem where (i) the variable to be estimated is a scalar $x$ with true value 
$\xtrue=0$, (ii) three measurements are available, the inliers $y_1=y_2=0$, and the outlier $y_3=4$, and (iii) 
 the measurement model is $y_i= x+n_i$, for all $i=1,2,3$, where $n_i$ is zero-mean and unit-variance additive Gaussian noise. 
Also, fix $\inthr=2.58$ 
{in \tls} such that $|n_i|\;\leq \inthr$ with probability $\simeq 0.99$, and, correspondingly,
fix $\outfreethr=11.35$ {in \mts} such that $n_1^2+n_2^2+n_3^2\leq \outfreethr$ with probability $\simeq 0.99$.\footnote{If $n_1,n_2,n_3$ are Gaussian random variables, each with mean  $0$ and variance $1$, then (i) $\mathbb{P}(|n_i|\;\leq 2.58)\simeq 0.99506$ for all $i=1,2,3$~\cite{NIST-NormalDistributionTable}, where $\mathbb{P}(\cdot)$ denotes probability; also, (ii) $n_1^2+n_2^2+n_3^2$ follows a $\chi^2$ distribution with $3$ degrees of freedom and $\mathbb{P}(n_1^2+n_2^2+n_3^2\leq 11.35)\simeq 0.99$~\cite{NIST-Chi2DistributionTable}.} 
Evidently, at $\xtrue=0$, $\res(y_1,\xtrue)=\res(y_2,\xtrue)=0$ and $\res(y_3,\xtrue)=4$. 

In this toy problem, \mts returns an incorrect estimate: \mts classifies all measurements as inliers for $x=4/3$, since then $\res^2(y_1,x)+\res^2(y_1,x)+\res^2(y_3,x)$ is minimized and is equal to $32/3\simeq 10.67$, which is less than $\outfreethr$.\footnote{\maxcon (eq.~\eqref{eq:maxConsensus}) also leads to a wrong estimate, selecting all measurements~as inliers (\eg  $x=2$ makes 
	all measurements to have residual smaller than $\inthr$).}

On the other hand, \tls rejects more measurements than \mts but finds the correct estimate: 
\tls returns $x=0$, classifying the third measurement as an outlier.
\end{example}

A comparison of \tls with \maxcon is given in~\cite[Appendix~C]{Yang20tro-teaser}.
%!TEX root = ../main.tex

\section{Inapproximability of \prone and \prtwo} \label{sec:hardness}

This section shows that \prone and \prtwo are computationally hard to solve, and in particular it is hard to even approximate their 
solution in quasi-polynomial time, in the worst case.

We start by recalling the 
$O(\cdot)$ and $\Omega(\cdot)$ notations from computational complexity theory~\cite{Arora09book-complexity}. 

\begin{definition}[O Notation]\label{def:bigO} Consider two functions $h:\mathbb{N}\to \mathbb{R}$ and $g:\mathbb{N}\to \mathbb{R}$. Then, $h(\measn)\!=\!O(g(\measn))$ means \mbox{there exists a} constant $k\!>\!0$ such that $h(\measn)\!\leq\! k g(\measn)$ for large enough $\measn$.
\end{definition}

\begin{definition}[$\Omega$ Notation]\label{def:bigOmega} Consider $h:\mathbb{N}\to \mathbb{R}$ and $g:\mathbb{N}\to \mathbb{R}$. Then, $h(\measn)=\Omega(g(\measn))$ means there exists a constant $k>0$ such that $h(\measn)\geq k g(\measn)$ for  large enough~$\measn$.
\end{definition}

\begin{definition}[$(\lambda, p)$-Approximability]\label{def:approx}
Consider $\lambda\geq 1$, and $p \geq 0$.  {\prone} is \emph{$(\lambda, p)$-approximable} if there exists an algorithm finding a sub-optimal solution $(\vxx,\outset)$ for {\prone}
such that $|\outset|\;\leq \lambda|\outset_\prone|$ and $\|\;\vres(\vy_{\measSet\setminus \outset}, \vxx)\;\|_\ell^2\;\leq \|\;\vres(\vy_{\measSet\setminus \outset_\prone}, \vxx_\prone)\;\|_\ell^2\;+\;p$, where $(\vxx_\prone, \outset_\prone)$ is an optimal solution for \prone. 

{Similarly, {\prtwo} is \emph{$(\lambda, p)$-approximable} if there exists an algorithm finding a sub-optimal solution $(\vxx,\outset)$ for {\prtwo}
such that $|\outset|\;\leq \lambda|\outset_\prtwo|$ and $\|\;\vres(\vy_{\measSet\setminus \outset}, \vxx)\;\|_\ell^2\;\leq \|\;\vres(\vy_{\measSet\setminus \outset_\prtwo}, \vxx_\prtwo)\;\|_\ell^2\;+\;p$, where $(\vxx_\prtwo, \outset_\prtwo)$ denotes an optimal solution to \prtwo.}
\end{definition}

Definition~\ref{def:approx} bounds the sub-optimality of an approximate solution to \prone or \prtwo: if $(\vxx, \outset)$ is an $(\lambda, p)$-approximate solution, then 
$\outset$ rejects up to a multiplicative factor $\lambda$ more outliers than the optimal set of outliers; and
$(\vxx, \outset)$ attains an inlier residual error up to an additive factor $p$ more than the residual error attained at the optimal 
solution.

\begin{theorem}[Inapproximability of \prone and \prtwo]\label{th:hardness}
For any $\delta\in(0,1)$, unless~$\np\; {\in}\;\bptime({|\measSet|}^{\poly\log |\measSet|})$, there exist 
a $\lambda(|\measSet|)=2^{\Omega(\log^{1-\delta} |\measSet|)}$,
a polynomial $p(|\measSet|)$, 
and instances of \prone
such that no quasi-polynomial algorithm makes the instances $(\lambda(|\measSet|), p(|\measSet|))$-approximable.  The result holds true even if the algorithm knows (i) $|\outset_\prone|$, and (ii) that the optimal solution is such that $\|\;\vres(\vy_{\measSet\setminus \outset_\prone}, \vxx_\prone)\;\|_\ell^2\;=0$.   

Similarly, the result holds true for \prtwo, even if the algorithm knows (i) $|\outset_\prtwo|$, and (ii) that the optimal solution is such that  $\|\;\vres(\vy_{\measSet\setminus \outset_\prtwo}, \vxx_\prtwo)\;\|_\ell^2\;=0$.
\end{theorem}

The theorem captures the extreme hardness of \prone and \prtwo: in the worst case, any quasi-polynomial algorithm for \prone and \prtwo 
cannot approximate the solution to \prone and \prone within an $(\lambda, p)$-approximation.
 This holds true  
even if the algorithm is informed with the optimal number of outliers to reject, or knows a priori that the optimal residual error is zero.
The quality of the approximation depends on the parameter $\lambda$ and $p$ in 
Theorem~\ref{th:hardness}, which are both polynomials. 
In particular, it can be seen that $\lambda$ (\cf Definition~\ref{def:approx}) grows with the number of measurements,  
 since $\lambda = \lambda(2^{\Omega(\log^{1-\delta} |\measSet|)})$ is proportional to $|\measSet|$ 
 when $\delta$ is close to $0$.

We remark that, since both $\lambda$ and $p$ in Theorem~\ref{th:hardness} depend on the number of measurements,  $|\measSet|$, the theorem implies there is no quasi-polynomial time algorithm achieving constant sub-optimality bound for \prone and \prtwo.  As such, the theorem strengthens recent inapproximability results for \maxcon that focus, instead, on polynomial-time algorithms only~\cite{Chin18eccv-robustFitting}.

\omittedTwo{We prove Theorem~\ref{th:hardness} by identifying instances of \emph{\maxcon, \mts, and \tls that are inapproximable in quasi-polynomial time}.}  
\omitted{To this end, we reduce to \maxcon, \mts, and \tls the problem \emph{variable selection}, which is inapproximable quasi-polynomial time~\cite{Foster15colt-variableSelectionHard}; variable selection is presented in Appendix~\ref{app:hardness} along with Theorem~\ref{th:hardness}'s proof.}

%!TEX root = ../main.tex

\section{Adaptive Trimming (\name) Algorithm}
\label{sec:adaptIntro}

We present \adapt, {a general-purpose, deterministic, and li- near time algorithm} for \prone, 
that requires no initial guess.  
We first describe a simple greedy algorithm in Section~\ref{sec:greedy}, to build intuition, and then introduce \adapt in Section~\ref{sec:adapt}. 
	%!TEX root = ../main.tex

\subsection{Gentle Start: Greedy Outlier Rejection}
\label{sec:greedy}

We start by describing a simple greedy algorithm for \prone,
to build intuition about \adapt.
The algorithm starts by solving a least squares problem akin to  
eq.~\eqref{eq:leastSquares} over the 
entire set of measurements, and, at each iteration, it rejects the measurement with the largest residual.
The algorithm stops once the condition $\| \vresyx{\measSet\setminus \outset} \|_\ell \;\leq \outfreethr$ in \eqone is satisfied.%

Although the described greedy algorithm is appealing for its simplicity and linear runtime,\footnote{In the literature, there exists an alternative greedy algorithm~\cite{Nemhauser78mp-submodularity} that, at each iteration, 
	tests the impact of rejecting each measurement (by solving multiple estimation problems), and then rejects only the measurement
	that induces the largest decrease in the objective function. 
	We do not consider such a variant since it has
	quadratic complexity in the number of measurements, 
	and does not scale to the problems we consider in Section~\ref{sec:experiments}.
}$^,$\footnote{At each iteration, the described greedy algorithm rejects one measurement, and, as a result, has linear runtime in the number of measurements.}~
% \footnote{\HYedit{In a class of estimation problems using the $\ell_\infty$ norm, it is possible to guarantee the greedy algorithm can remove outliers~\cite{Olsson10cvpr}.}}
\!(i) it cannot correct past mistakes (once a measurement is rejected, it is never reconsidered)
and (ii) the algorithm terminates once the threshold $\outfreethr$ is met, without, however, assessing if all outliers have been rejected, \eg by checking whether $\|\; \vresyx{\measSet\setminus \outset} \;\|_2$ 
has converged. 
\omittedTwo{Indeed,
 the greedy algorithm, being an approximation procedure, may satisfy the threshold $\outfreethr$ by simply over-rejecting measurements, instead of  rejecting all outliers. Therefore, it may be the case $\MminusO$ still contains outliers whose removal would largely reduce $\|\; \vresyx{\measSet\setminus \outset} \;\|_2$. Instead, if $\MminusO$  contains no outliers, $\|\; \vresyx{\measSet\setminus \outset} \;\|_2$ would remain largely unchanged even if more measurements would to be removed from $\MminusO$.}
Because of these, the algorithm can exhibit deteriorated performance; \cf~\SLAM experiments in Section~\ref{sec:exp-slam}.

% \adapt improves upon the greedy algorithm by addressing the greedy's weaknesses described above.

	%!TEX root = ../main.tex

\subsection{Beyond Greedy: \adapt Algorithm}
\label{sec:adapt}

%%%%%%%%%%
%!TEX root = ../main.tex

\begin{algorithm}[t]\label{alg:adapt}
	\caption{\adaptLong (\adapt).}
	\SetAlgoLined
	\KwIn{Measurements $\vy_i, \; \forall i\in\measSet$; thresholds $\minnoisebound$, $\theta$; \textit{MaxIterations}, \samplestoconvergeAdapt $>0$; \thrdiscount$\in (0,1)$.}
	\KwOut{Estimate of $\vxxtrue$ and corresponding inliers.}
	\BlankLine
	$\inset\myat{0} = \meas$;\quad$\vxx\myat{0} \myin \arg\min_{\vxx \in \calX}\l \|\; \vresyx{\inset\myat{0}}\;\|_2^2$\label{line:adapt-initializeStart}\;
	$\inlierNoiseAdapt^{(0)} = \thrdiscount\cdot \max_{i \; \in \; \inset\myat{0}}\; \res(\vy_i,\vxx\myat{0})$;\quad$j=0$\label{line:adapt-initializeEnd}\;
	\For{$ \iteration =1,\ldots, \text{MaxIterations}$\label{line:adapt-forloop} }{
    $\inset\myat{\iteration}= \{ i\;\in\;\meas \;\text{ s.t. }\;  \res(\vy_i,\vxx\myat{\iteration-1}) \leq \inlierNoiseAdapt^{(\iteration-1)} \}$\label{line:adapt-inlierUpdate}\;
		$\vxx\myat{\iteration} \myin  \arg\min_{\vxx \in \calX}\; \|\; \vresyx{\inset\myat{\iteration}}\;\|_2^2$\label{line:adapt-variableUpdate}\;
    \eIf{$\|\; \vresyx{\inset\myat{t}}\;\|_\ell \; < \minnoisebound$
%    	(|\inset\myat{t}|)
    	\label{line:adapt-if-first1}\\ \textbf{\emph{and}} $|\|\; \vresyx{\inset\myat{t}}\;\|_2^2 -\|\; \vresyx{\inset\myat{t-1}}\;\|_2^2|\;< \theta(|\inset\myat{t-1}|, |\inset\myat{t}|)$\label{line:adapt-if-first2}\label{line:adapt-convergenceCheck}}{
        $j++$\;\label{line:adapt-convergenceProgress}
    }{
        $j=0$\;\label{line:adapt-convergenceReset}
    }
    \If{$j=\samplestoconvergeAdapt$\label{line:adapt-if}}{
      \KwBreak\;
    }
    $\inlierNoiseAdapt^{(\iteration)} = \thrdiscount\cdot \max_{i \; \in \; \inset\myat{t}}\; \res(\vy_i,\vxx\myat{t})$\;\label{line:adapt-thresholdUpdate}
	}
	\Return{{$(\vxx\myat{\iteration}, \inset\myat{\iteration})$}.\label{line:adapt-done}}
\end{algorithm}

%%%%%%%%%%

We present the \emph{\adaptLong} (\adapt) algorithm to solve the \prone formulation in Problem~\ref{pr:main_1}.\footnote{The presentation in this section is slightly more general than  our original proposal in~\cite{Tzoumas19iros-outliers}, 
which only focused on \mtsLong. In this paper, we show that changing the stopping condition in \adapt also allows solving 
\maxconLong (\maxcon) problems.}  
The algorithm processes all measurements at each iteration, and \emph{trims} (rejects) measurements violating an inlier threshold (the threshold is set at each iteration and decreases iteration after iteration). The algorithm is \emph{adaptive} in that it dynamically decides the threshold at each iteration.  \adapt is not greedy in that it can correct previous mistakes: 
a measurement that has been deemed to be an outlier at an iteration can be 
re-included in the set of inliers at subsequent iterations, and, similarly, a measurement that has been deemed to be an inlier at an iteration can be (re-)included in the set of outliers at subsequent iterations.
\adapt is not greedy also in that it assesses whether all outliers have been rejected by checking whether $\|\; \vresyx{\measSet\setminus \outset} \;\|_2$ has converged.
Finally, \adapt can reject multiple measurements at each iteration, whereas greedy rejects one.
Its pseudo-code is given in Algorithm~\ref{alg:adapt}. 

\myParagraph{Initialization} 
\adapt's line~\ref{line:adapt-initializeStart} initializes the putative set of inliers to $\inset\myat{0}=\meas$ (all measurements);
at the subsequent iterations $t=1,2,\ldots$, the set $\inset\myat{\iteration}$ will include only the measurements classified as inliers at $\iteration$. Given $\inset\myat{0}$, \adapt sets $\vxx\myat{0} \myin \argmin_{\vxx  \in \calX}\|\; \vresyx{\inset\myat{0}}\;\|_2^2$,
\ie $\vxx\myat{0}$ is the estimate assuming all measurements are inliers; the nonlinear least squares problem can be minimized using non-minimal solvers, see~\cite{Yang20ral-GNC}.
Using $\vxx\myat{0}$, line~\ref{line:adapt-initializeEnd} sets the initial inlier threshold $\inlierNoiseAdapt^{(0)}$ equal to  $\thrdiscount\cdot \max_{i \; \in \; \inset\myat{0}}\; \res(\vy_i,\vxx\myat{0})$, \ie a multiplicative factor \thrdiscount  less than the maximum residual at $\vxx\myat{0}$.  That way, at least one measurement will be classified as an outlier at the next iteration. {In this paper, we always set $\thrdiscount = {0.99}$.}

\myParagraph{Main Loop} After the initialization, \adapt starts the main outlier rejection loop (line~\ref{line:adapt-forloop}).
We describe each step below.

\myParagraph{a) Inlier Set Update} 
At iteration $\iteration$, 
given $\inlierNoiseAdapt^{(t-1)}$, 
 line~\ref{line:adapt-inlierUpdate} updates the set of inliers $\inset\myat{\iteration}$ to contain measurements with residual smaller than $\inlierNoiseAdapt^{(t-1)}$. 
Since \adapt checks \emph{all} measurements in $\meas$, $\inset\myat{\iteration}$ may \emph{contain} measurements that \emph{were not} in $\inset\myat{\iteration-1}$, and may  \emph{not contain} measurements that \emph{were} in $\inset\myat{\iteration-1}$.
This allows \adapt to re-include measurements that were incorrectly rejected as outliers at previous iterations, and to reject measurements that were incorrectly classified as inliers. 
Notably, $\inset\myat{\iteration}$ depends on the history $\inset\myat{1}, \ldots, \inset\myat{\iteration-1}$, since  $\inlierNoiseAdapt^{(t-1)}$ depends on $ \inset\myat{\iteration-1}$ (cf.~line~15 of Algorithm~1), which in turn depends on $\inlierNoiseAdapt^{(t-2)}$, and so forth. 
Therefore, as \adapt iterates, a sequence $(\calI^{(1)},\epsilon^{(1)}), \ldots, (\calI^{(t)}, \epsilon^{(t)}), \ldots$ is generated, and,  \emph{ideally}, even if measurements are misclassified at early iterations, eventually all are classified correctly.

\myParagraph{b) Variable Update} 
Given $\inset\myat{\iteration}$, a new estimate $\vxx\myat{\iteration}$ is computed in line~\ref{line:adapt-variableUpdate}.
Line~\ref{line:adapt-variableUpdate}'s minimization is a nonlinear least squares problem that is solved using non-minimal solvers~\cite{Yang20ral-GNC}.

\myParagraph{c) Inlier Threshold Update}  
If the current estimate is infeasible for \prone and/or convergence of $\|\; \vresyx{\measSet\setminus \outset} \;\|_2$'s value has not been observed for \samplestoconvergeAdapt consecutive iterations (\ie the ``if'' conditions in lines~\ref{line:adapt-if-first1}-\ref{line:adapt-if-first2} and line~\ref{line:adapt-if} are not satisfied), 
\adapt updates $\inlierNoiseAdapt^{(\iteration)}$ (line~\ref{line:adapt-thresholdUpdate}) and moves to the next iteration. 
Similarly to line~\ref{line:adapt-initializeEnd}, line~\ref{line:adapt-thresholdUpdate} updates the threshold by applying a multiplicative factor $\thrdiscount<1$ to the maximum residual at the current iteration; this ensures that at least $1$ measurement is rejected at the next iteration.

\myParagraph{Termination} 
\adapt terminates when: 
	\begin{itemize}
		\item a maximum number of iterations is reached (\cf ``for'' loop in line~\ref{line:adapt-forloop}; in this paper, we set \maxiter $={1000}$);
		\item 
	 a feasible estimate for \prone is found and for \samplestoconvergeAdapt iterations $\|\; \vresyx{\measSet\setminus \outset} \;\|_2$ changes by at most $\theta$ (\cf ``if'' conditions in lines~\ref{line:adapt-if-first1}-\ref{line:adapt-if-first2} and line~\ref{line:adapt-if}).  In this paper, \samplestoconvergeAdapt$=3$.
%	 , while $\theta$ is application dependent; 
	 A probabilistically-grounded method to chose $\theta$ is described in Section~\ref{sec:experiments}.
	\end{itemize}
 Upon termination, \adapt returns the current estimate $\vxx\at{\iteration}$ and inlier set $\inset\myat{\iteration}$ (line~\ref{line:adapt-done}).
 The following remark ensures that \adapt terminates after at most $|\meas|$ iterations.

\begin{remark}[Linear Runtime]
	\adapt's policy to update $\inthr^{(\iteration)}$  (line~\ref{line:adapt-thresholdUpdate}) implies that
	$|\inset\myat{\iteration}|\;\leq |\inset\myat{\iteration-1}|-1$, hence
	\adapt terminates in at most $|\meas|$ (number of measurements) iterations.
\end{remark}

\begin{remark}[vs. \ransac]
\ransac is a \emph{randomized algorithm} for \prone, whereas \adapt is \emph{deterministic}.
\ransac maintains only a \emph{``{local view}'} of the measurement set $\meas$, building an inlier set by sampling a \emph{minimal} set of measurements; instead, \adapt looks at all measurements in $\meas$ to pick an inlier set.
\ransac assumes the availability of \emph{minimal solvers}, while {\adapt} assumes the availability of 
\emph{non-minimal solvers}. 
\ransac is unsuitable for high-dimensional problems, since the number of iterations required to sample an outlier-free set 
increases exponentially in the dimension of the problem~\cite{Bustos18pami-GORE}; in contrast, \adapt runs in \emph{linear time}, terminating in at most as many iterations as the number of~measurements.
\end{remark}
\vspace{-2mm}
%!TEX root = ../main.tex

%%%%%%%%%%%%
%!TEX root = ../main.tex

\begin{figure}[t]
  \vspace{2mm} % To align with text
  \begin{center}
  		\hspace{-3mm}
  	\begin{minipage}{\columnwidth}
  	\begin{tabular}{cc}%
  		\hspace{-4mm}
  		\begin{minipage}{\mpColFour}%
  			\centering%
  			\includegraphics[width=\columnwidth]{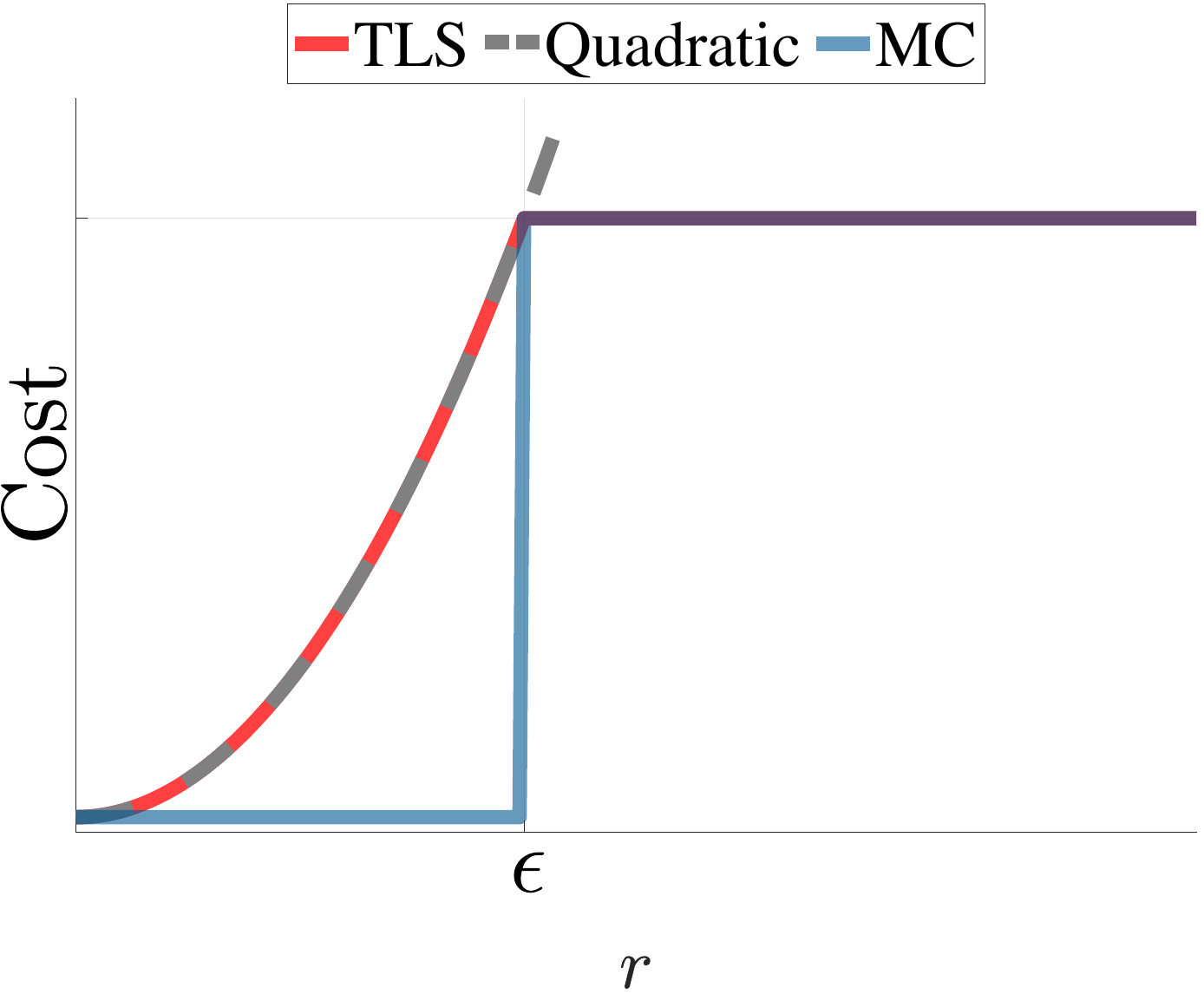}
  		\end{minipage}
  		& \mpMidSpaceFour
  		\begin{minipage}{\mpColFour}%
  			\centering%
  			\includegraphics[width=\columnwidth]{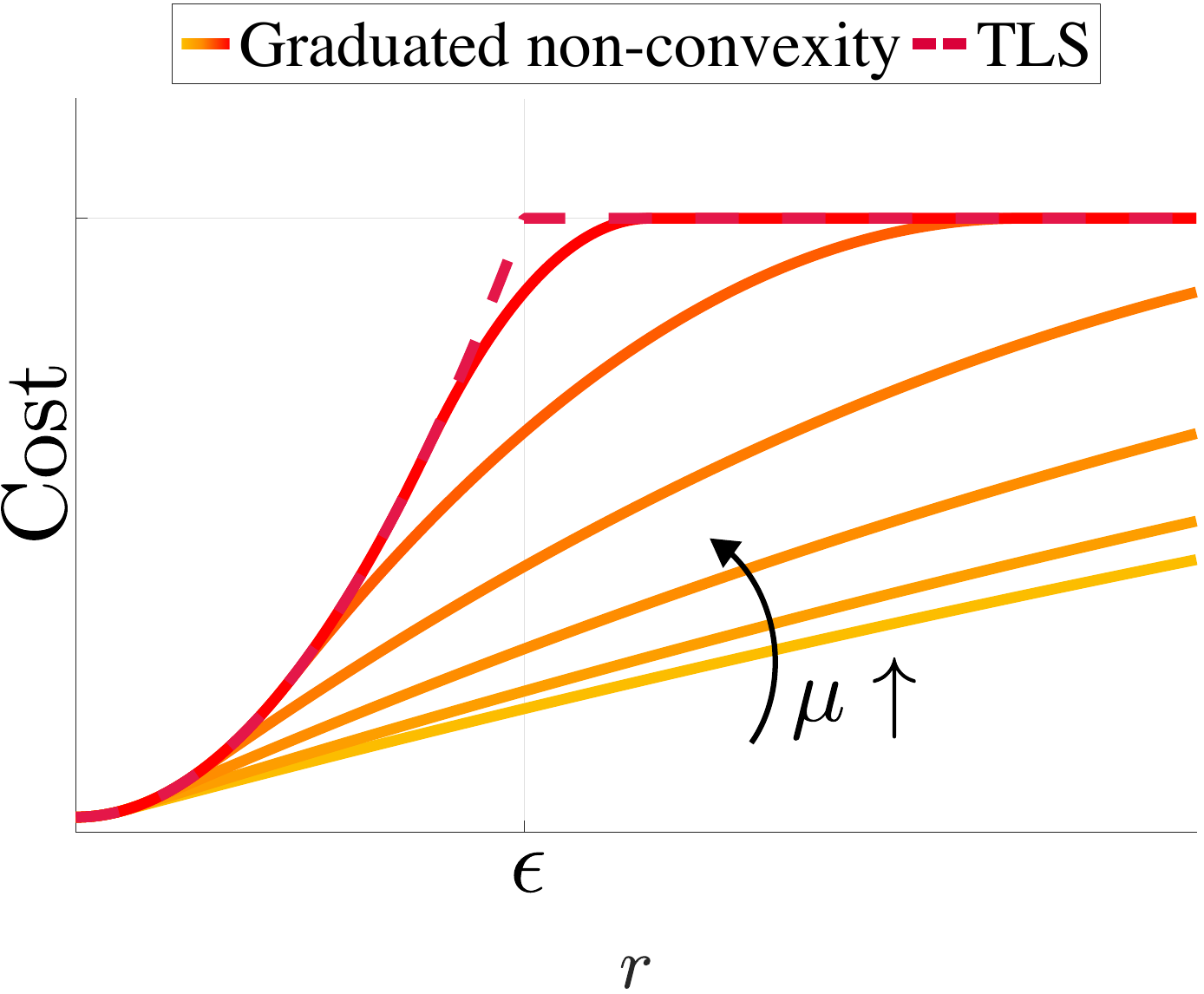}
  		\end{minipage}
  	\end{tabular}
  \end{minipage}
  \begin{minipage}{\textwidth}
  \end{minipage}
  \vspace{-3mm} 
	\caption{(a) \tls, quadratic, and \maxcon cost functions, (b) graduated non-convexity with control parameter $\mu$ for \tls cost function.\label{fig:gnctls}}  
  \label{fig:mestimation}
\end{center}
% \vspace{-5mm} 
\end{figure}

%%%%%%%%%%%%

\section{Graduated Non-convexity (\GNC) Algorithm}
\label{sec:GNC}

We provide a brief review of the \gnc algorithm we presented in previous work~\cite{Yang20ral-GNC}.
We show that ---when considering \tls costs--- the algorithm can be simply explained without invoking 
Black-Rangarajan duality. Moreover, we provide a novel local convergence result (Theorem~\ref{th:weightsTLSlimit} below), which 
enables simpler stopping conditions for the algorithm.

% \gnc is {a general-purpose, deterministic, and scalable algorithm to solve \tls without requiring} an initial guess. 
 We briefly review graduated non-convexity in Section~\ref{subsec:prelim-gnc-tls}, and
  describe \gnc in Section~\ref{subsec:gnc-tlsAlgo}. 

%%%%%%%%%%%%%%%%%%%%%%%%%%%%%%%%%%%%%%%%%%%%%%%%%%%%%%%%%%%%%%%%%%%%%%%%%%%%%%%%%%%%%%%%%%%%%%%%%%%%%
\subsection{\mbox{Preliminaries on Graduated Non-convexity}}\label{subsec:prelim-gnc-tls}

Before introducing the \gnc algorithm we review the notion of graduated non-convexity~\cite{Black96ijcv-unification,Zach14eccv,Mobahi2015WCVPR-continuationConvexEnvelope, Yang20ral-GNC}.

For convenience, we recall that our goal in this section is to solve the \tls problem (already introduced in~\eqref{eq:tls}):
\begin{equation}\label{eq:tls'}
\min_{\vxx \;\in \;\calX}\;\;\;  \sumAllPointsIn{\meas}\; \min_{w_i \;\in\; \{0,1\}} \;\left[w_i\;\resSqyx{i}  +  (1-w_i)\;\inthr^2\right].
\end{equation}

Solving the minimization~\eqref{eq:tls'} is hard because 
the \tls objective function is highly non-convex in the residual errors $r$.
Indeed, the $i$-th summand in~\eqref{eq:tls'}, namely $\min_{w_i \in \{0,1\}}\left[w_i\resSqyx{i}  +  (1\!-\!w_i)\inthr^2\right]$, 
describes a truncated quadratic function, that is nonconvex as shown in Fig.~\ref{fig:gnctls}(a). 

\emph{Graduated non-convexity} circumvents this non-convexity by using a \emph{homotopy} (or \emph{continuation}) method~\cite{Mobahi2015WCVPR-continuationConvexEnvelope}. 
In particular, graduated non-convexity proposes to ``soften'' the non-convexity in \tls by replacing the 
cost with a surrogate function controlled by a parameter $\mu$:
\bea \label{eq:GNC-outlierProcessWithoutIteration}
\min_{\vxx\; \in\; \calX}\;\;\;  \sumAllPointsIn{\measSet}\; \min_{w_i \;\in\; [0,1]} \;\left[ w_i \;r^2(\vy_i,\vxx) + \frac{\mu (1-w_i)}{ \mu + w_i} \;\barcsq \right],
\eea 
where the ``regularization'' term $(1-w_i)\inthr^2$ in eq.~\eqref{eq:tls'} is replaced with $\nicefrac{\mu (1-w_i) \inthr^2 }{(\mu + w_i)}$.
The surrogate function in~\eqref{eq:GNC-outlierProcessWithoutIteration} is such that 
(i) for $\mu\rightarrow 0$, eq.~\eqref{eq:GNC-outlierProcessWithoutIteration} becomes 
a 
convex optimization problem~\cite{Yang20ral-GNC},
and (ii)  
for $\mu\rightarrow +\infty$, the term $\nicefrac{\mu (1-w_i) \inthr^2 }{(\mu + w_i)}  \rightarrow (1-w_i)\inthr^2$, \ie eq.~\eqref{eq:GNC-outlierProcessWithoutIteration}  retrieves the original \tls in eq.~\eqref{eq:tls'}.  
The family of surrogate functions (parametrized by $\mu$) is shown in Fig.~\ref{fig:gnctls}(b).

Given the surrogate optimization problems in eq.~\eqref{eq:GNC-outlierProcessWithoutIteration}, graduated non-convexity starts by solving a 
convex approximation of the \tls problem (\ie for small $\mu$) and then gradually increases the non-convexity (by increasing $\mu$) 
till the original \tls cost is retrieved (\ie for large $\mu$). The estimate at each iteration is used as initial guess for the 
subsequent iteration, to reduce the risk of convergence to local minima.

%%%%%%%%%%%%%%
%!TEX root = ../main.tex

\begin{algorithm}[t]\label{alg:gnc}
	\caption{Graduated Non-Convexity for Truncated Least Squares (\gnc-\tls)~\cite{Yang20ral-GNC}.}
	\SetAlgoLined
	\KwIn{Measurements $\vy_i, \; \forall i\in\measSet$;  threshold $\lag\geq 0$; \textit{MaxIterations} $>0$; $\gncdisc>1$.}
	\KwOut{Estimate of $\vxxtrue$ and corresponding inliers.}
	\BlankLine
	$\mu^{(0)}=\frac{\inthr^2}{2\max_{i\; \in\; \meas}\;\res^2(\vy_i, \vxx\myat{0})-\inthr^2}$\label{line:gnc-initializemu}\;
	\mbox{$\weights\myat{0} = \bm{1}_\meas;\;\;\vxx\myat{0} = \mathrm{VariableUpdate}(\weights^{0})$\tcp*{\hspace{-0.7em}eq.\hspace{0.2em}\eqref{eq:gncVariableInitilize}}}\label{line:gnc-initializeStart}\\%\vspace{-4mm}{\color{white}\;}
	\For{$ \iteration =1,\ldots, \text{MaxIterations}$\label{line:gnc-forloop} }{
	% \While{$\iteration < \text{MaxIterations}$ {\bf and} $\mathrm{IsNotBinary}(\weights\myat{\iteration-1}$)\label{line:whileStart}}{
  \mbox{$\weights\myat{\iteration} = \mathrm{WeightUpdate}(\vxx\myat{\iteration-1},\mu^{(\iteration-1)}, \inlierNoise)$\label{line:gnc-weightsUpdate}\tcp*[l]{\hspace{-0.6em}eq.\hspace{0.2em}\eqref{eq:weightUpdate}}}\\
  $\vxx\myat{\iteration} = \mathrm{VariableUpdate}(\weights\myat{\iteration})$;
  \label{line:gnc-variableUpdate}\hspace{.2em}\tcp{\hspace{-0.6em}eq.~\eqref{eq:variableUpdate}}
%	$v^{(\iteration)} = \sumAllPointsIn{\meas}\; w_i^{(\iteration)}\;\res^2(\vy_i,\vxx\myat{\iteration})$\label{line:gnc-valueUpdate}\;
   $\mu^{(\iteration)} = \gncdisc\cdot \mu^{(\iteration-1)}$\label{line:gnc-muUpdate}\;
   \lIf{$\mathrm{IsBinary}(\weights\myat{\iteration}) = \mathrm{true}$}{\KwBreak\label{line:gnc-convergence}}
	% \lIf{$|v^{(\iteration)}-v^{(\iteration-1)}|\;<\convergthr$}{\KwBreak\label{line:gnc-convergence}}
	}
	\Return $(\vxx\myat{\iteration}, \supp(\bm w\myat{\iteration-1}))$.
\end{algorithm}
%%%%%%%%%%%%%%

%%%%%%%%%%%%%%%%%%%%%%%%%%%%%%%%%%%%%%%%%%%%%%%%%%%%%%%%%%%%%%%%%%%%%%%%%%%%%%%%%%%%%%%%%%%%%%%%%%%%%
\subsection{\gnc-\tls Algorithm}\label{subsec:gnc-tlsAlgo}
The pseudo-code of \gnc-\tls is given in Algorithm~\ref{alg:gnc}. 
Besides leveraging graduated nonconvexity, at each iteration, \gnc-\tls  
minimizes the surrogate function in eq.~\eqref{eq:GNC-outlierProcessWithoutIteration} by alternating a 
minimization with respect to $\vxx$ (with fixed $w_i$) to a minimization of the weights $w_i$ (with fixed $\vxx$).
Both minimizations can be solved efficiently, as described below.  

\myParagraph{Initialization} 
\gnc-\tls's line~\ref{line:gnc-initializemu} initializes the parameter $\mu$ to a small number as suggested in~\cite{Yang20ral-GNC}.
 Line~\ref{line:gnc-initializeStart} also initializes all weights to $1$ (\ie $\vw\myat{0} = \bm{1}_{\meas}$, where $\bm{1}_{\meas}$ is the vector of all ones with length equal to $|\meas|$) 
 and sets the initial $\vxx$ to be the solution of the least squares problem:
\beq\label{eq:gncVariableInitilize}
\vxx\myat{0} \myin  \argmin_{\vxx \in \calX}\; \sumAllPointsIn{\measSet} r^2(\vy_i, \vxx).
\eeq 
which we denote in the algorithm as $\mathrm{VariableUpdate}(\weights^{0})$.

\myParagraph{Main Loop}  After the initialization, \gnc-\tls starts the main outlier rejection loop (line~\ref{line:gnc-forloop}).
At iteration $\iteration$, \gnc-\tls minimizes the surrogate function in eq.~\eqref{eq:GNC-outlierProcessWithoutIteration} 
by alternating a minimization over the weights (line~\ref{line:gnc-weightsUpdate}) and a minimization over the variable 
$\vxx$ (line~\ref{line:gnc-variableUpdate}); then, \gnc-\tls increases the amount of nonconvexity by increasing the parameter $\mu$ (line~\ref{line:gnc-muUpdate}).
The details of these steps are given below.

\myParagraph{a) Weight Update} 
At iteration $t$,
\gnc-\tls updates the weights $\vw\myat{t}$ to minimize the surrogate function in eq.~\eqref{eq:GNC-outlierProcessWithoutIteration} 
while keeping fixed $\vxx\myat{\iteration-1}$ and $\mu^{(\iteration-1)}$ (line~\ref{line:gnc-weightsUpdate}): 
\begin{equation} 
\vw^{(t)}
\in \argmin_{w_i \;\in \; [0,1]} \;\sumAllPointsIn{\measSet} \;\left[ w_i \;\res_i^{(\iteration)}  + \frac{\mu\at{\iteration-1} (1-w_i)}{ \mu\at{\iteration-1} + w_i} \;\barcsq\right], \label{eq:weightUpdate}
\end{equation}
where $\res_i^{(\iteration)} \triangleq r(\vy_i, \vxx\myat{t})$; eq.~\eqref{eq:weightUpdate} splits into $|\meas|$ scalar problems~\cite{Yang20ral-GNC} and admits the following closed-form solution:
\begin{equation}
w_i^{(t)}=	\left\{
\begin{array}{ll}
1,& \res_i^{(\iteration)}  < \inthr\; \sqrt{\frac{\mu^{(t-1)}}{\mu^{(t-1)} + 1}}\\
0, & \res_i^{(\iteration)} > \inthr\; \sqrt{\frac{\mu^{(t-1)}+1}{\mu^{(t-1)}}}\\
\frac{ \inthr\sqrt{ \mu\myat{t-1}(\mu\myat{t-1}+1)}}{ \res_i^{(\iteration)} }- \mu\myat{t-1}, &
\text{otherwise}. \label{eq:weightUpdate-TLS}\\		
\end{array}
\right.
\end{equation}
\omitted{ Expectedly, $\iota_\vw(\res,\mu,\inthr) \in [0,1]$.  In the limit, $\iteration\rightarrow +\infty$, the following holds.}

\myParagraph{b) Variable Update} 
Line~\ref{line:gnc-variableUpdate} updates $\vxx\myat{t}$ by 
minimizing the surrogate function in eq.~\eqref{eq:GNC-outlierProcessWithoutIteration} 
while keeping fixed $\vw\myat{\iteration}$: 
\bea \label{eq:variableUpdate}
\vxx\myat{t} \in \argmin_{\vxx\; \in \;\calX} \; \sumAllPointsIn{\measSet} \; w\at{t}_i \;r^2(\vy_i, \vxx),
\eea
where we dropped the additional summand in eq.~\eqref{eq:GNC-outlierProcessWithoutIteration}, since it is independent of $\vxx$. 
The optimization problem in eq.~\eqref{eq:variableUpdate} is a weighted least squares problem (cf.~eq.~\eqref{eq:leastSquares}), and can be solved globally using certifiably optimal non-minimal solvers~\cite{Yang20ral-GNC}.  

\myParagraph{c) Increasing Non-convexity: $\bm{\mu}$ Update} 
At the end of each iteration, \gnc-\tls increases $\mu$ by a multiplicative factor $\gncdisc>1$ (line~\ref{line:gnc-muUpdate}), 
getting one step closer to the original non-convex \tls cost function (\cf Fig.~\ref{fig:gnctls}(b)).
As in~\cite{Yang20ral-GNC}, we choose \gncdisc $=1.4$ in \gnc.

\myParagraph{Termination} 
\gnc-\tls terminates when (i) the maximum number of iterations is reached (line~\ref{line:gnc-forloop}) 
---in this paper, \maxiter $= {1000}$---, or 
(ii) the weight vector $\vw^{(\iteration)}$ becomes a binary vector (line~\ref{line:gnc-convergence}).
The latter stopping condition is supported by the following theorem.

\begin{theorem}[{$\vw\myat{t}$\! Tends \!to \!a \!{Binary} \!Vector with \!Probability \!1}]\label{th:weightsTLSlimit}
	If $\iteration\rightarrow+\infty$, then $w_i^{(t)}\rightarrow w_i^{(\infty)}$, where, for all $i \in \meas$,
	\begin{equation} 
	w_i^{(\infty)} = \left\{
	\begin{array}{rr}
	1,& \res_i^{(\infty)}  < \inthr;\\
	0, & \res_i^{(\infty)} > \inthr;\\
	1/2, & \res_i^{(\infty)} =\inthr. \label{eq:weightLimit-TLS}\\		
	\end{array}
	\right.
	\end{equation}
Moreover, since the measurements are affected by random noise, the case $\res_i^{(\infty)} =\inthr$ happens with zero probability.
\end{theorem}
Eq.~\eqref{eq:weightLimit-TLS} {agrees with the \tls formulation in eq.~\eqref{eq:tls'},} since $w_i^{(\infty)}=1$ only when $\res_i^{(\infty)} 
<\inthr$, \ie when measurement $i$ is considered an inlier, while $w_i^{(\infty)}=0$ otherwise.

%!TEX root = ../main.tex

\section{\AutoTunned\xspace \adapt and \gnc: \\\adaptfree and \gncfree}
\label{sec:parameterFree}

We now present the \autotunned versions of \adapt and \gnc, namely, \adaptfree and \gncfree.  In contrast to \adapt and \gnc, they do not require knowledge of a threshold to separate inliers from outliers ($\outfreethr\!$ in \adapt, $\inthr\!$ in \gnc).

	%!TEX root = ../main.tex

\subsection{\adaptfree Algorithm}\label{subsec:adapt-auto}

%%%%%%%%%%%%
%!TEX root = ../main.tex

\begin{algorithm}[t]\label{alg:adapt-auto}
	\caption{\Autotunned\xspace \adapt (\adaptfree).}
	\SetAlgoLined
    \KwIn{Measurements $\vy_i, \; \forall i\in\measSet$; \textit{MaxIterations} $>0$; \thrdiscount$\in (0,1)$; 
    		\samplestoconverg, \windowsize, \convergthr $\geq 0$.}
	\KwOut{Estimate of $\vxxtrue$ and corresponding inliers.}
	\BlankLine
	$\inset\myat{0} = \meas$;\quad$\vxx\myat{0} \myin \arg\min_{\vxx \in \calX}\l \|\; \vresyx{\inset\myat{0}}\;\|_2^2$\label{line:adaptauto-initializeStart}\;
	$\inlierNoiseAdapt^{(0)} = \thrdiscount\cdot \max_{i \; \in \; \inset\myat{0}}\; \res(\vy_i,\vxx\myat{0})$\label{line:adaptauto-initializeMiddle}\; %\quad$\iteration = 1$
	$\delta^{(0)} = \separation(\vres(\vy_{\meas},\vxx\myat{0}))$\label{line:adaptAuto-deltaInitialization}\;
	\For{$\iteration = 1, \ldots,\text{MaxIterations}$\label{line:adapAuto-terminationMaxIter}}{
		$\inset\myat{\iteration}= \{ i\;\in\;\meas \;\text{ s.t. }\;  \res(\vy_i,\vxx\myat{\iteration-1}) \leq \inlierNoiseAdapt^{(\iteration-1)} \}$\label{line:adaptAuto-inlierUpdate}\;
		$\vxx\myat{\iteration} \myin  \arg\min_{\vxx \in \calX}\; \|\; \vresyx{\inset\myat{\iteration}}\;\|_2^2$\label{line:adaptAuto-variableUpdate}\;
		$\inlierNoiseAdapt^{(\iteration)} = \thrdiscount \cdot \max_{i \; \in \; \inset\myat{\iteration}}\; \res(\vy_i,\vxx\myat{\iteration})$\label{line:adaptAuto-thresholdUpdate}\;
		$\delta^{(\iteration)} =1/\delta^{(0)} \cdot \separation(\vres(\vy_\meas,\vxx\myat{\iteration}))$\label{line:adaptAuto-deltaUpdate}\;
		$\sigma^{(\iteration)} = \movstd(\delta^{(\iteration)}, \windowsize)$\label{line:adaptAuto-sigmaUpdate}\;
		% \If{$|\sigma^{\last-\text{SamplesToConverg}+1:\last}| < \text{ConvergThr}$}{
		% 	\KwBreak\;
		% }
		% $\iteration++$\;
		\If{$\iteration>\samplestoconverg$ {\bf and} $\sigma^{(\iteration-\samplestoconverg\; :\; \iteration-1)} < \convergthr$\label{line:adaptAuto-ifcondition}}{\KwBreak
    }
	}
	\Return{$(\vxx\myat{\iteration},\inset\myat{\iteration-\samplestoconverg})$.}
\end{algorithm}
%%%%%%%%%%%%

\begin{figure}[h]
	\begin{center}
		\centering%
		\includegraphics[width=1.0\columnwidth]{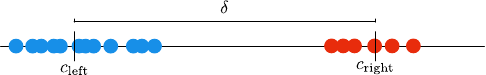}
		\caption{Two clusters of non-negative residuals: the low-magnitude ones (blue) are centered at $c_{\text{left}}$, 
		the high-magnitude ones (red) at $c_{\text{right}}=c_{\text{left}}+\delta$.}
		\label{fig:clusteringExample}
	\end{center}
\end{figure}

\adaptfree is similar to \adapt, but introduces a novel, inlier-threshold-free termination condition.  In contrast to \adapt, which terminates based on a given $\minnoisebound$ (which separates inliers from outliers) 
\adaptfree (i) looks at the residuals of all measurements, given the current estimate $\vxx\myat{\iteration}$, (ii) clusters them into two groups, a group of low-magnitude residuals ---the ``inliers'' (left group in Fig.~\ref{fig:clusteringExample})--- and a group of high-magnitude residuals ---the ``outliers'' (right group in Fig.~\ref{fig:clusteringExample})--- and (iii) terminates once the two groups ``stabilize,'' in particular, when the distance $\delta$ between the centroids of two groups converges to a steady state. To cluster all residuals in $\meas$ into two groups, and to compute their centroids and their in-between distance, \adaptfree calls the subroutine $\separation$ presented in \ref{app:alg-separation} (Algorithm~\ref{alg:separation}).

The pseudo-code of \adaptfree is given in Algorithm~\ref{alg:adapt-auto}.

\myParagraph{Initialization} 
\adaptfree's lines~\ref{line:adapt-initializeStart}-\ref{line:adaptauto-initializeMiddle} are the same as \adapt's, and initialize $\inset\myat{0}$ and  $\vxx\myat{0}$. Line~\ref{line:adaptAuto-deltaInitialization} is new: given $\vxx\myat{0}$, it initializes $\delta^{(0)}$, \ie the distance between the inlier and outlier clusters at $\vxx\myat{0}$.  At the subsequent iterations $\iteration=1,2,\ldots$, the value $\delta^{(0)}$ is used as a normalization factor in the update of $\delta^{(\iteration)}$ (line~\ref{line:adaptAuto-deltaUpdate}, discussed below).

\myParagraph{Inlier Set, Variable, and Inlier Threshold Update} 
Li- nes~\ref{line:adaptAuto-inlierUpdate}, \ref{line:adaptAuto-variableUpdate}, and \ref{line:adaptAuto-thresholdUpdate} in \adaptfree 
describe the same inlier set, variable, and inlier threshold updates used in \adapt.

\myParagraph{{Inlier vs.~Outlier Cluster Separation Update}}   \adaptfree updates $\delta^{(\iteration)}$  with the distance between the inlier and outlier clusters at the current $\vxx\myat{\iteration}$, after normalizing it by $\delta^{(0)}$ (line~\ref{line:adaptAuto-deltaUpdate}). The role of the normalization is discussed in Remark~\ref{rmk:normalization} below.

\myParagraph{Termination} \adaptfree terminates when (i) the maximum number of iterations is reached (\cf ``for'' loop in line~\ref{line:adapAuto-terminationMaxIter}), or (ii)  $\delta^{(\iteration)}$ converges to a steady state value, indicating the inlier and outlier clusters have also converged to a steady state.
Specifically, \adaptfree declares convergence when for \samplestoconverg consecutive iterations $\delta^{(\iteration)}$'s moving standard deviation $\sigma^{(t)}$ is less than \convergthr (line~\ref{line:adaptAuto-ifcondition}).
In more detail, $\sigma^{(\iteration)}$ is the standard deviation of $\delta^{(\iteration)}$ across the 
last \windowsize iterations and
is computed in line~\ref{line:adaptAuto-sigmaUpdate}, where $\movstd$ is the corresponding \MATLAB function. 
In this paper, $\windowsize =\num{3}$, $\samplestoconverg =5$ and $\convergthr={10^{-4}}$ always.

\begin{remark}[Role of Normalization in \adaptfree]\label{rmk:normalization}
The normalization  by $\delta^{(0)}$ in line~\ref{line:adaptAuto-deltaUpdate} is necessary,  since across different applications the residuals can differ by several orders of magnitude, 
and, as a result, the distance between the inlier and outlier clusters can differ by several orders of magnitude. 
The normalization reduces the impact of the magnitude of the residuals on the stopping conditions of \adaptfree.
\end{remark}

\begin{remark}[Tuning \adapt's $\minnoisebound$ vs.~Tuning \adaptfree's \convergthr]
Tuning $\minnoisebound$ requires knowledge of the inlier threshold (or equivalently, the inlier noise), which varies 
not only across applications (\eg mesh registration vs.~\SLAM) but also across problem instances within the same application (\eg different \SLAM datasets).  
In contrast, \convergthr is fixed across instances of an application (for all applications in this paper, in particular, \convergthr is the same), and its value can be set given a single dataset where the ground truth is known. In this sense, \adaptfree is \emph{\autotunned}.
\end{remark}
	%!TEX root = ../main.tex

\subsection{\gncfree Algorithm}\label{subsec:gnc-auto}

%%%%%%%%%%%%
%!TEX root = ../main.tex

\begin{algorithm}[t]\label{alg:gnc-auto}
	\caption{\mbox{\Autotunned\xspace \gnc for \tls (\gncfree).}}
	\SetAlgoLined
  \KwIn{Measurements $\vy_i, \; \forall i\in\measSet$; \textit{MaxIterations}  $>0$;
   $\gncdisc >1$;
 \mbox{\noiseupbound, \noiselowbound $\geq 0$;} \mbox{$\chi^2$ distribution's degrees of freedom $d>0$.}} 
  \KwOut{Estimate of $\vxxtrue$ and corresponding inliers.}
	\BlankLine
  % $\gamma = \ksdistr(1-\alpha);\quad j=1$;\quad $k=0$\label{line:gncAuto-ksdistr}\;
  $\inlierNoise\at{0} = \noiseupbound;\quad j=1$\label{line-gncAuto-inlierthrInitialization}\;
  $\mu^{(0)}=\frac{(\inthr\at{0})^2}{2\max_{i\; \in\; \meas}\;\res^2(\vy_i, \vxx\at{0})-(\inthr\at{0})^2}$\label{line:gncAuto-initializationmu}\;
  $\weights\at{0} = \bm{1}_\meas;\;\;\vxx\at{0} = \mathrm{VariableUpdate}(\weights\at{0})$\label{line-gncAuto-weightVariableInitialization}\;
	\For {$ \iteration =1,\ldots, \text{MaxIterations}$\label{line:gncAuto-termination-maxiter}}{
		$\weights\at{\iteration} = \mathrm{WeightUpdate}(\vxx\at{\iteration-1},\mu^{(\iteration-1)}, \inlierNoise\at{j-1})$\label{line:gncAuto-weight-update}\;
    $\vxx\at{\iteration} = \mathrm{VariableUpdate}(\weights\at{\iteration})$\label{line:gncAuto-variable-update}\;
    $\mu^{(\iteration)}=\gncdisc\cdot\mu^{(\iteration-1)}$\label{line:gncAuto-mu-update}\;
		\If{$\mathrm{IsBinary}(\weights\at{\iteration})$\label{line:gncAuto-convergence}}{
			$\inset\at{j} = \supp(\weights\at{\iteration})$\label{line:gncAuto-inliersatconvergence}\;
      $s\at{j} = \fitChi(\vres(\vy_{\inset\at{j}},\vxx\at{\iteration}), d)$\label{line:gncAuto-fit}\;
      $\widetilde{\weights}\at{j} = \weights\at{\iteration}$; \quad
      $\widetilde{\vxx}\at{j} = \vxx\at{\iteration}$\label{line:gncAuto-storeweights}\;
      $ s_\min= \min_{z \in \{1,2,\ldots,j\}} s\at{z}$\label{line:gncAuto-bestfitsofar}\;
      \uIf{$s\at{j} = s\at{j-1}$\label{line:gncAuto-termination-nochange}}{
        \KwBreak;%\tcp*{Fitness the same}
      }
      \uElseIf{$s\at{j} > s_\min$ \label{line:gncAuto-termination-fitworsening}}{ 
        $k++$\tcp*{Fitness worsens}
        \If{$k=\samplestoconvergeGNC$\label{line:gncAuto-termination-convergence}}{
          \KwBreak; %\tcp*{Converged}
        }
      }\Else{
        $k=0$\label{line:gncAuto-resetconvergence}\;
      }
      \mbox{$\tilde{\inlierNoise} \!=\! \displaystyle\max_{i\; \in \; \inset\at{j}}\;\{\res(\vy_{i},\vxx\at{\iteration}) \;\text{s.t.}\;\res(\vy_{i},\vxx\at{\iteration}) \!<\!\inlierNoise\at{j-1}\}$}\;
      $\inlierNoise\at{j} = (\inlierNoise\at{j-1} + \tilde{\inlierNoise})/2$\label{line:gncAuto-inlierguess-update}\;
      \If{$\inlierNoise\at{j}=\inlierNoise\at{j-1}$ {\bf or} $\inlierNoise\at{j} < \noiselowbound$ \label{line:gncAuto-termination-LowerBound}}{
      	\KwBreak; %\tcp*{Threshold too small}
      }
      {$\mu^{(\iteration)} = \mu^{(0)}; \hspace{.5em}\weights\at{\iteration} = \vw\at{0}; \hspace{.5em}\vxx^t = \vxx\at{0};\hspace{0.5em} j++$}\label{line:gncAuto-reset}\;
    }
  }
	%\tcp{Return inlier set with best fit}
	$j_\min = \argmin_{z \; \in \; \{1,2,\ldots, j\}} \;s^{(z)}$\label{line:gncAuto-bestfit}\;
	\Return{$(\widetilde{\vxx}\at{j_\min}, \supp(\widetilde{\weights}\at{j_\min})$.\label{line:gncAuto-return}}
\end{algorithm}
%%%%%%%%%%%% 7

\gncfree, in contrast to \gnc-\tls, does not require knowledge of a suitable inlier threshold $\inthr$.
Instead, \gncfree requires only an upper and lower bound for $\inthr$, denoted by \noiseupbound and \noiselowbound in the algorithm.
\gncfree uses \noiseupbound as an initial guess $\inthr\at{0}$ to the unknown inlier threshold $\inthr$.
Using $\inthr\at{0}$, \gncfree performs the same \textit{weight, variable, and $\mu$ update} steps as \gnc-\tls until convergence, when $\vw\at{\iteration}$ becomes binary.
At this point, \gncfree (i) scores how well the empirical distribution of the squares of the residuals fits a $\chi^2$ distribution, using the Cram\'{e}r–von Mises test, restricting the test to the measurements classified as inliers at iteration $\iteration$,\footnote{Proposition~\ref{th:mle_for_tls} implies that for \tls the inliers' generative probability distribution is a Normal distribution.   As a result, the square of the inliers' residuals will follow a $\chi^2$ distribution.} 
(ii) stores the score and the current estimate, and
(iii) decreases the value of $\inthr\at{\iteration}$ to prepare for the next iteration.
The algorithm terminates (i) when the $\chi^2$ fitness score either remains unchanged or worsens across consecutive iterations, or (ii) when   
$\inthr\at{\iteration}$ either  remains unchanged across consecutive iterations or becomes less than $\noiselowbound$.
\gncfree is given in Algorithm~\ref{alg:gnc-auto}, and is described in detail below.

\myParagraph{Initialization}
\gncfree first initializes $\inthr\at{0}$ with \noiseupbound. 
Then $\mu\at{0}$, $\weights\at{0}$, and $\vxx\at{0}$ are initialized similarly to \gnc but using $\inthr\at{0}$ instead of $\inthr$
(lines~\ref{line:gncAuto-initializationmu}-\ref{line-gncAuto-weightVariableInitialization}).
\gncfree also introduces the counter $j$ (initialized to $1$ in line~\ref{line-gncAuto-inlierthrInitialization}), which counts how many times $\inthr\at{\cdot}$ has been updated. 

\myParagraph{Weight, Variable, and $\bm{\mu}$ Update} 
Lines~\ref{line:gncAuto-weight-update}, \ref{line:gncAuto-variable-update}, and \ref{line:gncAuto-mu-update}
 in \gncfree are the same as the corresponding updated in \gnc, with the exception that the current guess $\inthr\at{j-1}$ is used in line~\ref{line:gncAuto-weight-update} instead of the unknown $\inthr$. 
Since these updates are the same as \gnc, Theorem~\ref{th:weightsTLSlimit} guarantees that 
the weights $\vw\at{\iteration}$ eventually become binary (for some $\iteration$), \ie\xspace \gncfree's iterations of weight, variable, and $\mu$ update converge. Line~\ref{line:gncAuto-convergence} checks whether this is indeed the case.
% (the check is effectively the same as the one used as stopping condition in~\gnc).

%!TEX root = ../main.tex

\begin{table*}[t]
\centering\footnotesize
\setlength{\tabcolsep}{4pt}
\setlength{\extrarowheight}{3pt}
% \begin{tabular*}{lccccccc}\toprule
\begin{tabular*}{\textwidth}{l @{\extracolsep{\fill}} ccccccc}
\emph{Application} & \greedymc & \greedymts & \adaptmc & \adaptmts & \adaptfree & \gnc & \gncfree \\
\midrule
\scenario{Mesh Registration} & 80\% [\SI{13.71}{\second}] & 80\% [\SI{12.98}{\second}] & 80\% [\SI{14.42}{\second}]& 80\% [\SI{14.39}{\second}] & 80\% [\SI{12.36}{\second}] & 80\% [\SI{5.12}{\second}] & 80\% [\SI{10.68}{\second}]\\
\scenario{Shape Alignment} & 80\% [\SI{0.15}{\second}] & 80\% [\SI{0.15}{\second}] & 80\% [\SI{0.22}{\second}]& 80\% [\SI{0.23}{\second}] & 80\% [\SI{0.25}{\second}] & 80\% [\SI{0.03}{\second}] & 80\% [\SI{0.06}{\second}]\\
\scenario{PGO (2D)} & 60\% [\SI{5.04}{\second}] & 10\% [\SI{0.76}{\second}] & 80\% [\SI{5.04}{\second}]& 80\% [\SI{4.92}{\second}] & 60\% [\SI{5.61}{\second}] & 90\% [\SI{1.41}{\second}] & 80\% [\SI{2.17}{\second}]\\
\scenario{PGO (3D)} & 60\%[\SI{9.23}{\hour}] & 40\%[\SI{9.55}{\hour}] & 60\%[\SI{60.4}{\minute}] & 40\%[\SI{42.04}{\minute}] & 90\%[\SI{61.3}{\minute}] & 90\%[\SI{85.8}{\second}] & 90\%[\SI{101.62}{\second}]\\
\bottomrule\hline
\end{tabular*}
\caption{\b{\textbf{Robustness of proposed algorithms.} Robustness to outliers and average of median running time of the proposed algorithms.}}\label{table:robustness}
\vspace{-7mm}
\end{table*}

\myParagraph{$\bm{\chi}^2$ Fitness Test} 
 Once $\vw\at{\iteration}$ has converged,
 \gncfree checks how well the residuals classified as inliers fit a $\chi^2$ distribution. 
 Line~\ref{line:gncAuto-inliersatconvergence} collects the inliers, and 
 line~\ref{line:gncAuto-fit} computes the fitness score $s\at{j}$ by calling $\fitChi$ (Algorithm~\ref{alg:fitcdf} in \ref{app:alg-fitcdf}). {The score $s\at{j}$ is such that $s\at{j}>0$; smaller value indicates better fit.} 
 Line~\ref{line:gncAuto-storeweights} stores the current estimate and weights.

\myParagraph{Inlier Threshold Update} 
Once the fitness score at $\vxx\at{\iteration}$ has been computed, \gncfree 
updates the inlier threshold guess to the mean between the current inlier threshold guess and the largest residual among the measurements currently classified as inliers 
(line~\ref{line:gncAuto-inlierguess-update}).
Evidently, $\inthr^{(j)} \leq  \inlierNoise\at{j-1}$.

\myParagraph{Re-initialization of Weights, Variable, and $\bm{\mu}$} Once $\inthr\at{j}$ has been updated, \gncfree re-initializes $\mu\at{\iteration}$, $\vw\at{\iteration}$, and $\vxx\at{\iteration}$ (line~\ref{line:gncAuto-reset}),
in preparation for another round of \gnc with the new threshold $\inthr\at{j}$.
The counter $j$ is also increased by $1$ (line~\ref{line:gncAuto-reset}).

\myParagraph{Termination} 
%Line~\ref{line:gncAuto-bestfitsofar} extract the best fitness score the algorithm attained so far.
\gncfree terminates when either
\begin{itemize}
	\item the maximum number of iterations is reached (line~\ref{line:gncAuto-termination-maxiter}), or
	\item the fitness score remains unchanged across $2$ consecutive iterations (line~\ref{line:gncAuto-termination-nochange})
	or the fitness score worsens for $\samplestoconvergeGNC$ consecutive iterations (lines~\ref{line:gncAuto-termination-fitworsening}-\ref{line:gncAuto-resetconvergence}; in this paper, $\samplestoconvergeGNC = 2$),\footnote{The intuition is that if outliers exist among the measurements, then decreasing $\inthr\at{j-1}$ to $\inthr\at{j}$ leads to rejecting more outliers, leading to a better $\chi^2$ fit. 
		But if all outliers have been rejected, then decreasing $\inthr\at{j-1}$ results into rejecting inliers, worsening the $\chi^2$ fit or keeping it the same.} \!or
	\item it is no longer possible to decrease $\inthr\at{j}$ (line~\ref{line:gncAuto-termination-LowerBound}) (when $\inthr\at{j}=  \inthr\at{j-1}$, then \gncfree would converge again to the same solution if it were to continue running).
\end{itemize}

Upon termination, \gncfree returns the inlier set with the best $\chi^2$ fitness score (lines~\ref{line:gncAuto-bestfit}-\ref{line:gncAuto-return}).

\begin{remark}[Tuning \gnc-\tls's $\inthr$ vs. Tuning~\gncfree's \noiseupbound and \noiselowbound]
Knowing $\inthr$, or estimating it accurately, can be hard and time consuming: $\inthr$ typically varies across both applications and problem instances within the same application.
In contrast, guessing upper and lower bounds for $\inthr$ is easier, making \gncfree \emph{\autotunned}.
\end{remark}

\begin{remark}[Termination in \gncfree]
In Proposition~\ref{th:mle_for_tls}, we observed 
\tls implicitly searches for inliers with Normally distributed residuals.
At the same time, the sum of the squares of Normally distributed variables 
%(with some unknown variance $\sigma^2$) 
follows a $\chi^2$ distribution~{\cite{Abramowitz48book-handbookMath}}.
For this reason, the stopping condition for \gncfree is based on a $\chi^2$ fitness test, performed by the $\fitChi$ routine used in line~\ref{line:gncAuto-fit}.  $\fitChi$ estimates the variance of the $\chi^2$ distribution, hence it implicitly guesses the magnitude of the inlier noise.
\end{remark}
%!TEX root = ../main.tex

\section{Experiments and Applications:\\ Mesh Registration, Shape Alignment, and \PGO}
\label{sec:experiments}

We showcase the proposed algorithms in three robot perception problems: mesh registration
(Section~\ref{sec:exp-registration}), 
shape alignment (Section~\ref{sec:exp-shape}), 
and Pose Graph Optimization (\PGO) (Section~\ref{sec:exp-slam}).
We performed all the experiments in \MATLAB running on a Linux machine with the Intel i-97920X (\SI{4.3}{\giga\hertz}).
No GPU support was used.

The results show that \adapt and \GNC outperform the state of the art and are robust up to $80-90\%$ outliers. 
Their \autotunned versions achieve similar performance, without relying on the knowledge of the inlier noise.
We summarize the observed performance of the algorithms (robustness to outliers and average median running time) in Table~\ref{table:robustness}, where we also include  \greedy's performance.  In Table~\ref{table:robustness}, we observe:
	\begin{itemize}
		\item \greedy is on average slower than the proposed algorithms ($2$ times slower than \gnc in mesh registration and shape alignment, and up to $100$ times slower than \gnc in \PGO); in addition to being slower,  \greedy is also less robust than both \adapt and \gnc in \PGO, and even against \adapt's and \gnc's \autotunned versions.
		% From the experimental evaluation (summarized in Table~\ref{table:robustness}) we observe that \gnc and \gncfree are superior in terms of balancing running time and estimation accuracy across all the applications. While \adapt and \adaptfree can exhibit superior accuracy, it usually come at the expense of running time.
		% The \greedy algorithm can perform similarly to \adapt and \adaptfree but its running time can be prohibitively slow(\eg~\SLAM 3D).
		\item \gnc and \gncfree achieve the lowest running time, retaining, at the same time, the robustness to outliers achieved by all proposed algorithms.  Specifically, in mesh registration and shape alignment, \adapt and \gnc, as well as their \autotunned versions, are practically on par with each other in terms of their robustness to outliers, yet \gnc and \gncfree are $2$ to $10$ times faster; and in the \PGO experiments, \adapt and \adaptfree can exhibit similar, or even higher accuracy than \gnc and \gncfree (cf.~Fig.~\ref{fig:slam_csail}), yet \gnc and \gncfree are on average $10$ times faster than \adapt and \adaptfree.
	\end{itemize} 

\myParagraph{Choice of Parameters} 
We refer to \adapt as \adaptmc if it solves the \maxcon problem ($\ell=+\infty$ in line~\ref{line:adapt-if-first1} of Algorithm~\ref{alg:adapt}), and as \adaptmts if it solves the \mts problem ($\ell=2$). 
We also compare against the greedy algorithm of Section~\ref{sec:greedy}, which we stop when the constraint in~\eqref{eq:main_1} 
is satisfied. We denote the corresponding technique with the label \greedymc and \greedymts, 
when we use $\ell=+\infty$ and  $\ell=2$ in~\eqref{eq:main_1}, respectively. 
In all applications, we set in
\begin{itemize}
	\item  \adapt: $\minnoisebound=\sqrt{\chiSqinv(0.99, nd)}$, where $d$ is the number of degrees of freedom of the measurement noise and depends on the application,
%	\footnote{$d$ is equal to $3$ for mesh registration, $2$ for shape alignment, $3$ for 2D \SLAM, and $6$ for 3D \SLAM.} 
	and $n$ is the cardinality of the chosen inlier set at the current iteration (\ie at \adapt's iteration $\iteration$, $n=|\inset\at{\iteration}|$; \cf~\adapt's line~\ref{line:adapt-inlierUpdate});  $\thetadmts=\sqrt{\udchinv(0.05, n_1d, n_2d, \sigma^2))}$, where  $\sigma$ is the standard deviation of the noise,  $n_1=|\inset\at{\iteration}|$ and $n_2=|\inset\at{\iteration-1}|$, 
	while $\udchinv$ is the inverse of the cumulative probability distribution of a random variable $z=|z_1-z_2|$, where $z_1$ and $z_2$ are $\chi^2$ random variables (\cf line~\ref{line:adapt-convergenceCheck} of \adapt);\footnote{We set $\thetadmts=\sqrt{\udchinv(0.05, n_1d, n_2d, \sigma^2))}$ assuming the measurement noise is normally distributed, since, then, $z_1=\|\; \vresyx{\inset\myat{t}}\;\|_2^2$ and $z_2=\|\; \vresyx{\inset\myat{t-1}}\;\|_2^2$ are indeed $\chi^2$ random variables.
  } \!\maxiter $= 1000$;  \samplestoconvergeAdapt $= 3$; and \thrdiscount $= 0.99$.
	\item \adaptfree:  \maxiter $= 1000$; \thrdiscount $= 0.99$; \samplestoconverg $=2$; \windowsize $=3$; \convergthr $=10^{-4}$.
	\item \gnc:  $\inthr=\sigma\sqrt{\chiSqinv(0.99, d)}$;  \maxiter $= 1000$; and \gncdisc $= 1.4$.
	\item \gncfree: \maxiter $= 1000$;  $\gncdisc= 1.4^2$;\footnote{We set $\gncdisc= 1.4^2$ in \gncfree such that the algorithm has similar runtime as \gnc. On average, by choosing $\gncdisc= 1.4^2$, instead of $1.4$, 
%		where the latter is $\gncdisc$'s value in \gnc, 
		we speed-up the convergence of the weights $\vw\at{\iteration}$ to a binary vector (\gncfree's line~\ref{line:gncAuto-convergence}) by a multiplicative factor of $2$.}	\!$\noiseupbound$ and $\noiselowbound$ depend on the application, and are described in the subsections below.
\end{itemize} 

		%!TEX root = ../main.tex

% ===================================
%!TEX root = ../../main.tex

\begin{figure*}[t!]
	\begin{center}
	\begin{minipage}{\textwidth}
	\begin{tabular}{ccc}%
		%%%%%%%%%%%%%%%%%%%%%%%%%%%%%%%%%%%%%%%%%%%%%%%%%%%%%%%%%%%%%%%%%%%%%%%%%%%%%%%%%%%%%%%%%%%%%%%%%%%%%%%%%
		\mpPreSpace
			\begin{minipage}{\mpColThree}%
        \centering%
        \includegraphics[width=\columnwidth]{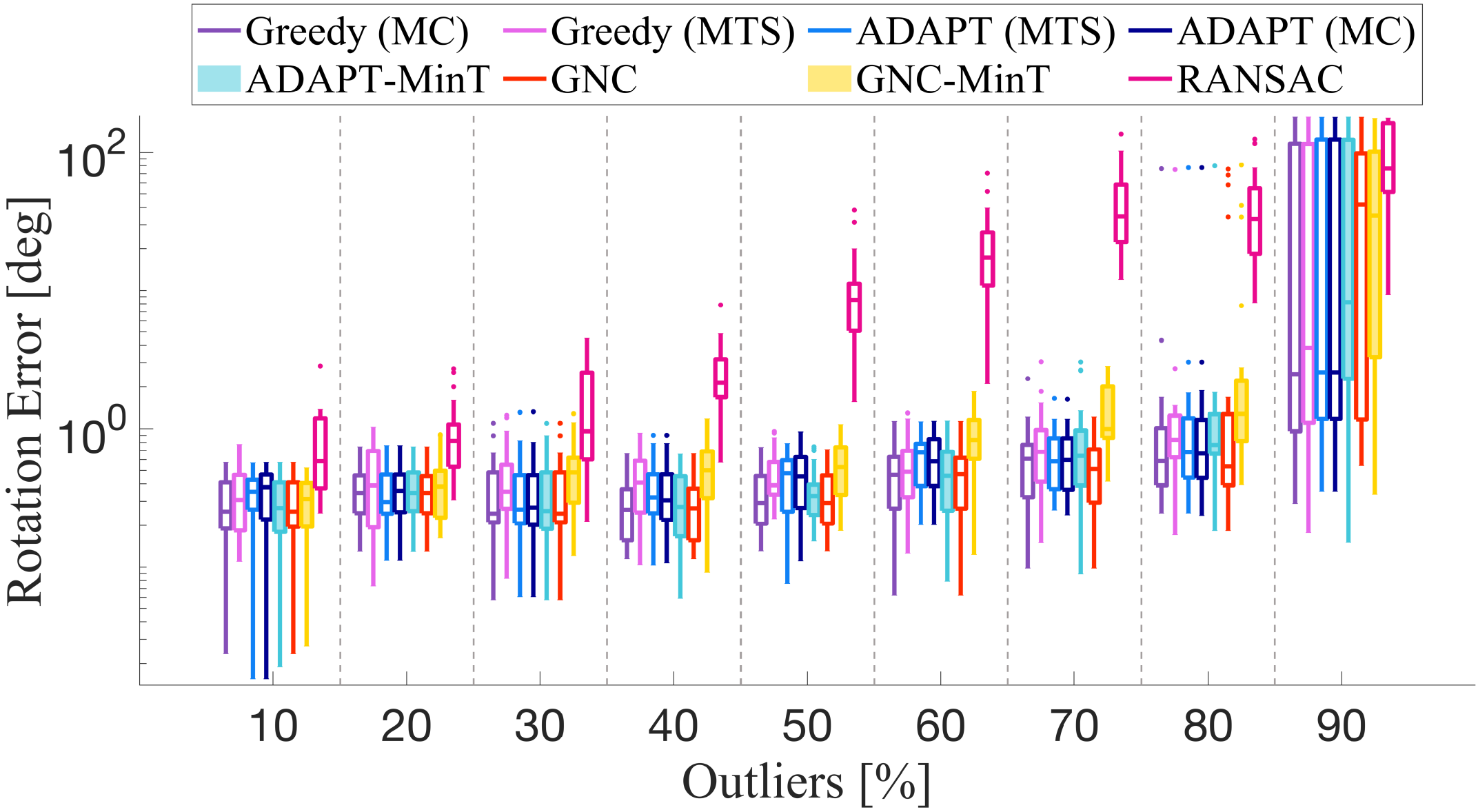} \\
			\end{minipage}
		& \mpMidSpaceThree
			\begin{minipage}{\mpColThree}%
        \centering%
        \includegraphics[width=\columnwidth]{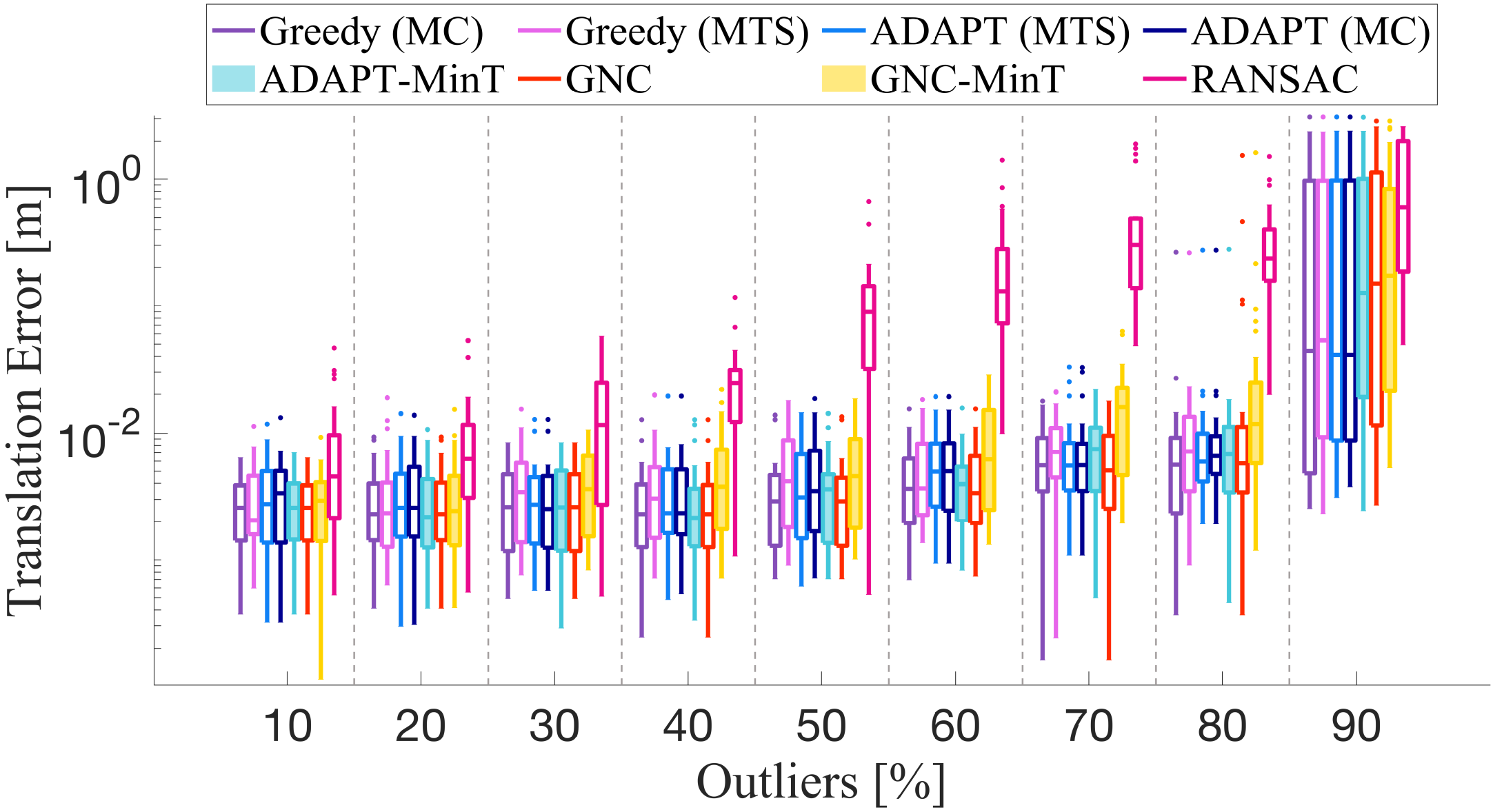} \\
      \end{minipage}
    & \mpMidSpaceThree
			\begin{minipage}{\mpColThree}%
        \centering%
        \includegraphics[width=\columnwidth]{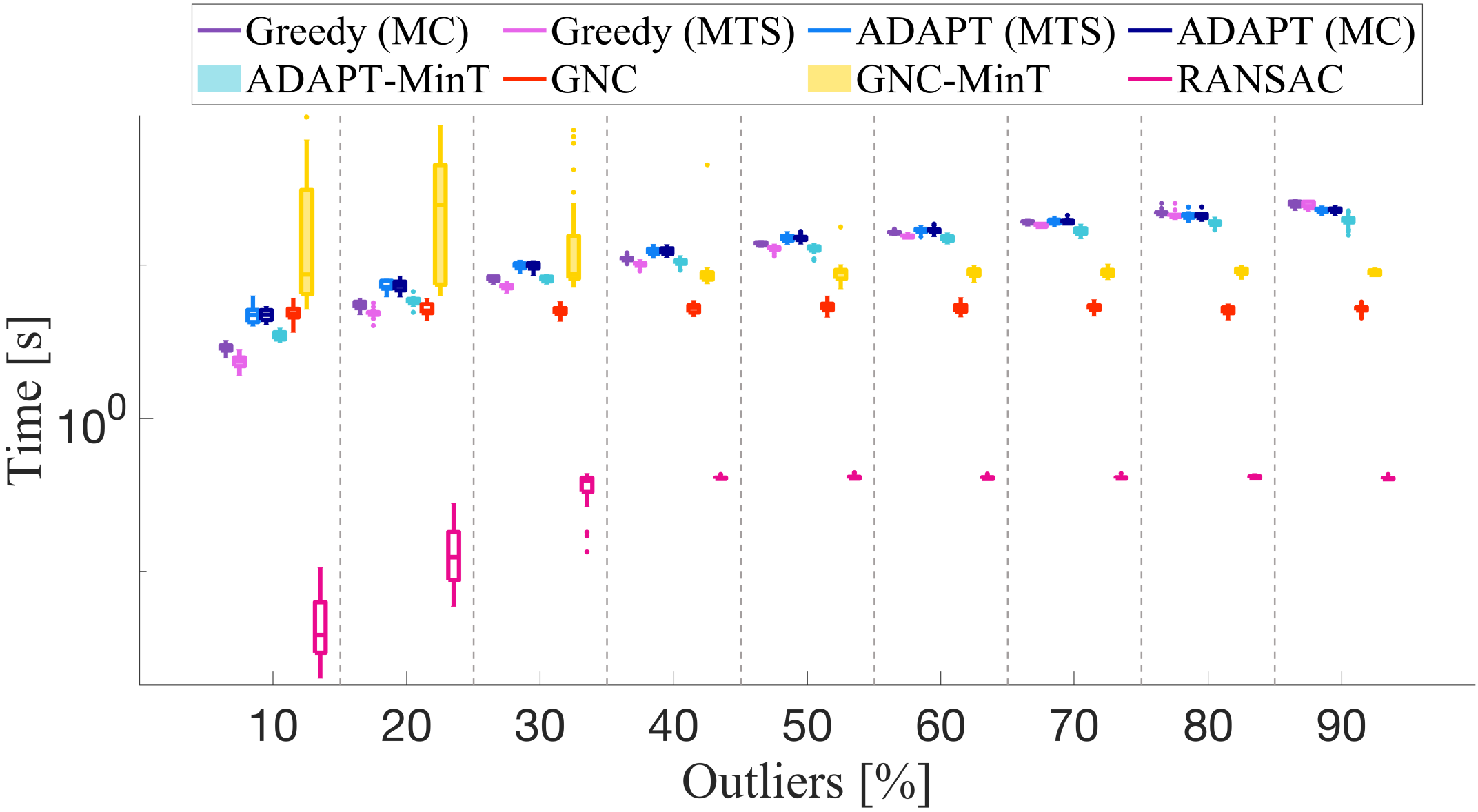} \\
			\end{minipage}
		\end{tabular}
	\end{minipage}
	\mpPostSpace
	\caption{ {\bf Mesh Registration.} Rotation error (left), translation error (center), and running time (right) of the proposed algorithms, compared to \ransac,  on the \scenario{PASCAL+}``aeroplane-2'' dataset~\cite{Xiang2014WACV-PASCAL+}.
	Statistics are computed over 25 Monte Carlo runs and for increasing percentage of outliers.}
	\label{fig:mesh}
	\vspace{-5mm} 
	\end{center}
\end{figure*}
% ===================================

% ===================================

\begin{figure*}[ht!]
	\begin{center}
	\begin{minipage}{\textwidth}
	\begin{tabular}{ccc}%
		%%%%%%%%%%%%%%%%%%%%%%%%%%%%%%%%%%%%%%%%%%%%%%%%%%%%%%%%%%%%%%%%%%%%%%%%%%%%%%%%%%%%%%%%%%%%%%%%%%%%%%%%%
		\mpPreSpace
			\begin{minipage}{\mpColThree}%
        \centering%
        \includegraphics[width=\columnwidth]{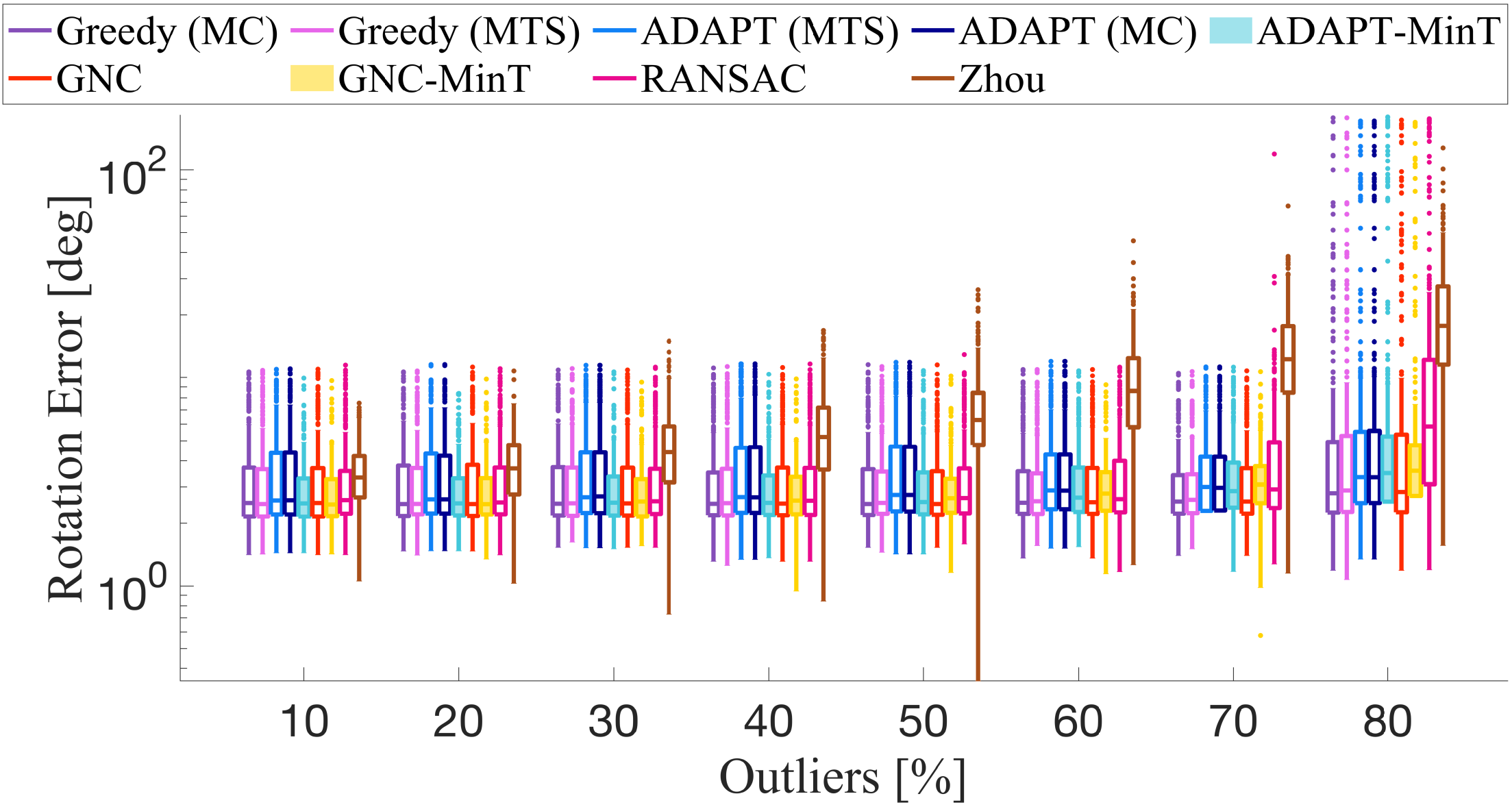} \\
			\end{minipage}
		& \mpMidSpaceThree
			\begin{minipage}{\mpColThree}%
        \centering%
        \includegraphics[width=\columnwidth]{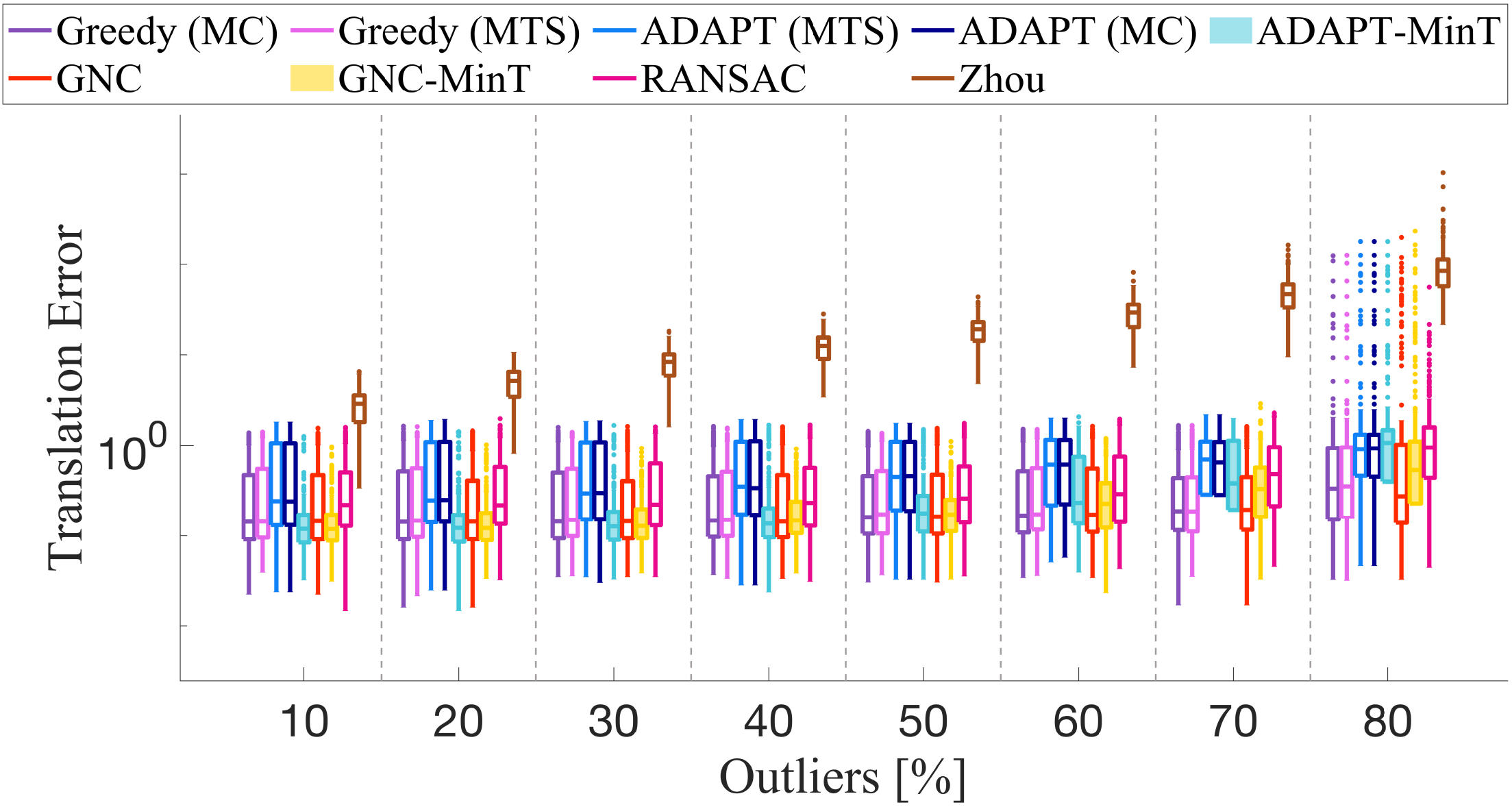} \\
      \end{minipage}
    & \mpMidSpaceThree
			\begin{minipage}{\mpColThree}%
        \centering%
        \includegraphics[width=\columnwidth]{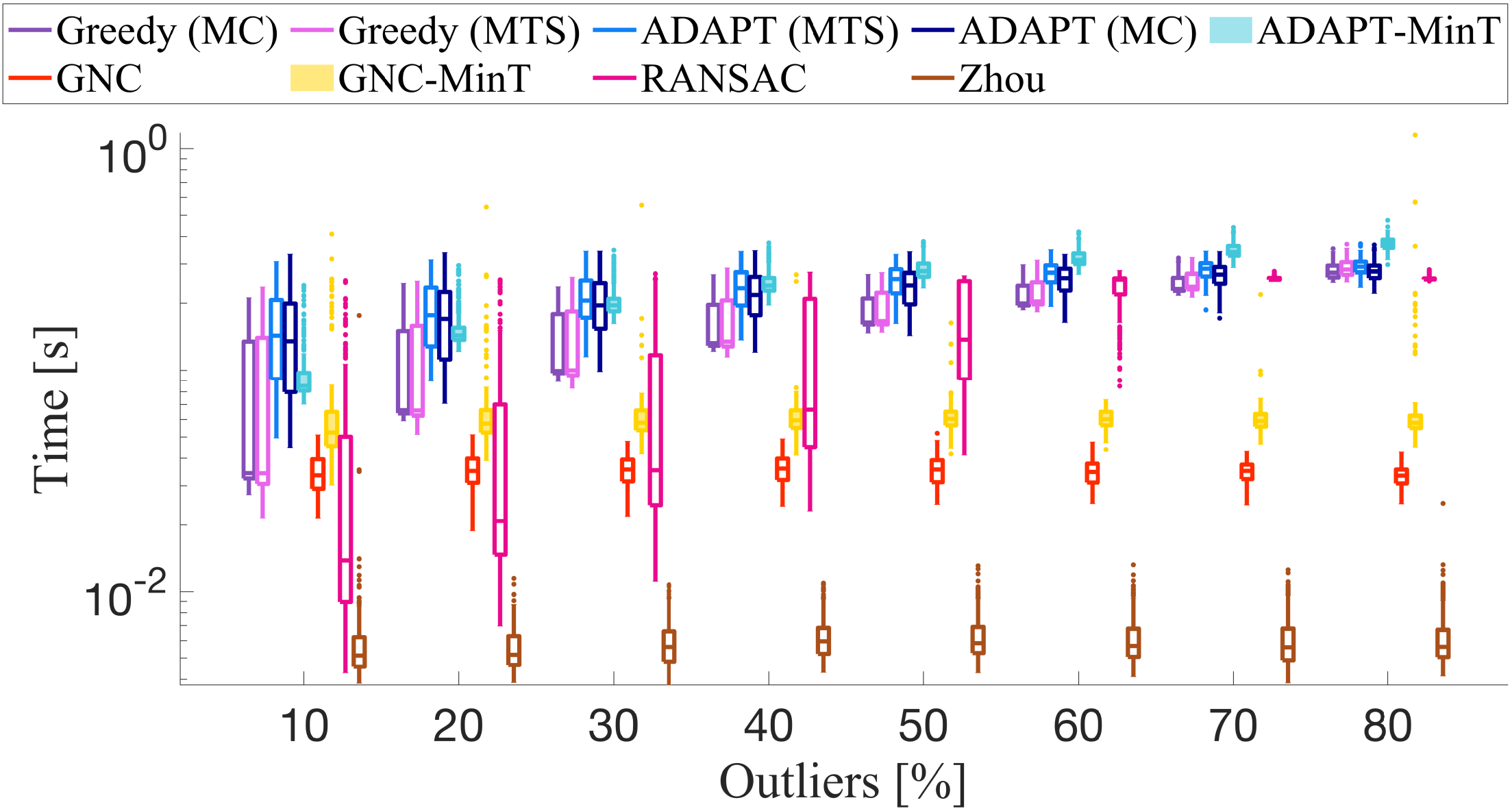} \\
			\end{minipage}
		\end{tabular}
	\end{minipage}
	\mpPostSpace
	\caption{ {\bf Shape Alignment.} 
Rotation error (left), translation error (center), and running time (right) of the proposed algorithms, compared to state-of-the-art techniques,  on the \scenario{FG3DCar} dataset~\cite{Lin14eccv-modelFitting}.
	Statistics are computed over 25 Monte Carlo runs and for increasing percentage of outliers.}
	\label{fig:shape}
	\vspace{-5mm} 
	\end{center}
\end{figure*}
% ===================================

\subsection{Mesh Registration}
\label{sec:exp-registration}

In mesh registration, given a set of 3D points 
$\va_i \in \Real{3}$, $i \in \meas$, 
and a set of primitives 
$\MP_i$, $i \in \meas$ 
(being points, lines and/or planes) with putative correspondences 
$\va_i \leftrightarrow \MP_i$, the goal is to find the best rotation $\MR \in \SOthree$ and translation $\vt \in \Real{3}$ that align the point cloud to the 3D primitives. 
In practice, the primitives $\MP_i$ often correspond to vertices, edges, or faces of the CAD model of an object, while 
the points $\va_i$ are measured points (\eg from a lidar observing a scene containing that object), 
and mesh registration allows retrieving the pose of the (known) object in the point cloud.

The residual error in mesh registration is 
$r(\MR,\vt) = \dist(\MP_i,\MR \va_i + \vt)$, 
where $\dist(\cdot)$ denotes the distance between a primitive $\MP_i$ and a point $\va_i$ after the transformation $(\vt,\MR)$ is applied. 
The formulation can also accommodate weighted distances to account for heterogeneous and anisotropic measurement noise. 
In the outlier-free case, Briales \etal~\cite{Briales17cvpr-registration} developed a certifiably optimal non-minimal solver when the 3D primitives include points, lines, and planes and the noise is anisotropic. 
We use \gnc, \adapt, and their \autotunned versions to efficiently robustify Briales' non-minimal solver. 

\myParagraph{Experimental Setup}
We use the ``aeroplane-2'' mesh model from the \scenario{PASCAL+} dataset~\cite{Xiang2014WACV-PASCAL+}. 
We compute statistics over 20 Monte Carlo runs, with increasing amounts of outliers.
At each Monte Carlo run, we generate a point cloud from the mesh by randomly sampling points lying on the vertices, edges, and faces of the mesh model, and then apply a random transformation, 
adding Gaussian noise with $\sigma=0.05d_{\text{mesh}}$, where $d_{\text{mesh}}$ is the diameter of the mesh. 
We establish \num{40} point-to-point, \num{80} point-to-line, and \num{80} point-to-plane correspondences, and create outliers by adding incorrect point to point/line/plane correspondences. 
Since the number of degrees of freedom of the measurement noise is $d=3$, $\inthr \!=\! \sigma\sqrt{\chiSqinv(0.99, d)} \!=\! \num{0.0128}$.  Moreover,
we choose $\noiseupbound \!=\! 3\inthr \!=\! 0.0384$, and $\noiselowbound  \!=\! \inthr/3 \!=\! 0.0043$.

We benchmark our algorithms against a \ransac implementation with \num{400} maximum iterations, using the \num{12}-point minimal solver presented in~\cite{Khoshelham16jprs-ClosedformSolutionPlaneCorrespondences}.

\myParagraph{Mesh Registration Results} 
Fig.~\ref{fig:mesh} shows the rotation error, translation error, and running time for each technique (all plots are in log-scale). 
The \greedymc, \greedymts, \gnc, \adaptmc, and \adaptmts, as well as the \autotunned \adaptfree have comparable performance, and are robust against up to 80\% outliers.
\gncfree has similar performance, exhibiting slightly higher errors.
All proposed methods outperform \ransac, which starts breaking at 30\% of outliers. 

In terms of runtime, \ransac's runtime grows with the number of outliers. 
Instead, \greedy's, \adapt's, and \adaptfree's runtimes grow linearly with the number of outliers, while \gnc's and \gncfree's remain roughly constant. 

Qualitative results for mesh registration are given in Fig.~\ref{fig:introSummary}.
		%!TEX root = ../main.tex

\subsection{Shape Alignment}
\label{sec:exp-shape}

In shape alignment, given 2D features $\vz_i \in \Real{2}, i \in \meas$ in a single image and 3D points $\MB_i \in \Real{3}, i \in \meas$
of an object with putative correspondences $\vz_i \leftrightarrow \MB_i$ (potentially including outliers), the goal is to find the best scale $s >0$, rotation $\MR$, and translation $\vt$ of the object that projects the 3D shape to the 2D image under weak perspective projection. 
In practice, the 3D points $\MB_i$ often correspond to distinguishable points on the CAD model of an object, while 
the 2D features $\vz_i$ are measured pixels (\eg from a camera observing a scene containing that object), 
and shape alignment allows retrieving the pose of the (known) object in the image. 

The residual error in shape alignment is $r(s,\MR,\vt) = \| \vz_i - s\Pi\MR \MB_i - \vt \|$, where $\Pi \in \Real{2 \times 3}$ is the weak perspective projection matrix (equal to the first two rows of a $3\!\times\!3$ identity matrix).
Note that $\vt$ is a 2D translation, but under weak perspective projection one can extrapolate a 3D translation (\ie recover the distance of the camera to the object) using the scale $s$. 
We use the closed-form solution introduced in~\cite{Ramakrishna12eccv-humanPose} as non-minimal solver. While potentially suboptimal, the solver in~\cite{Ramakrishna12eccv-humanPose} works
well in practice, and is faster than the certifiably optimal solver proposed in~\cite{Yang20ral-GNC}.

\myParagraph{Experimental Setup} 
We test the performance of \gnc, \gncfree, \adapt, and \adaptfree on the \scenario{FG3DCar} dataset~\cite{Lin14eccv-modelFitting} against (i) Zhou's method~\cite{Zhou17pami-shapeEstimationConvex}, and (ii) \ransac with 400 maximum iterations using a 4-point minimal solver.
We use the ground-truth 3D shape model as $\MB$ and the ground-truth 2D landmarks as $\vz$. 
To generate outliers for each image, we set random incorrect correspondences between 3D points and 2D features. 
We assume $\sigma = \sqrt{\num{1e-5}} = 0.0032$, and, since $d=2$, $\inthr = \sigma \sqrt{\chiSqinv(0.99, d))} = 0.0096$.  
Also, similarly to mesh registration, $\noiseupbound = 3\inthr = 0.0288$, and $\noiselowbound= \inthr/3 = 0.0032$.

\myParagraph{Shape Alignment Results} 
Fig.~\ref{fig:shape} shows in log-scale the rotation and translation error, and running time for all techniques. 
Statistics are computed over all the images in the \scenario{FG3DCar} dataset.
Zhou's method degrades quickly with increasing number of outliers.
Instead, all other algorithms are robust against 80\% of outliers.

\ransac's runtime grows exponentially with the number of outliers.
\gnc, \gncfree, and Zhou's method runtime is constant, being smaller than \ransac's for outlier rates more than 40\%. \adapt's and \adaptfree's runtimes grow linearly.

Qualitative results for shape alignment are given in Fig.~\ref{fig:introSummary}.
		%!TEX root = ../main.tex

% ===================================

\begin{figure*}[ht!]
	\begin{center}
	\begin{minipage}{\textwidth}
	\begin{tabular}{ccc}%
		%%%%%%%%%%%%%%%%%%%%%%%%%%%%%%%%%%%%%%%%%%%%%%%%%%%%%%%%%%%%%%%%%%%%%%%%%%%%%%%%%%%%%%%%%%%%%%%%%%%%%%%%%
		\mpPreSpace
			\begin{minipage}{\mpColTwo}%
        \centering%
        \includegraphics[width=\columnwidth]{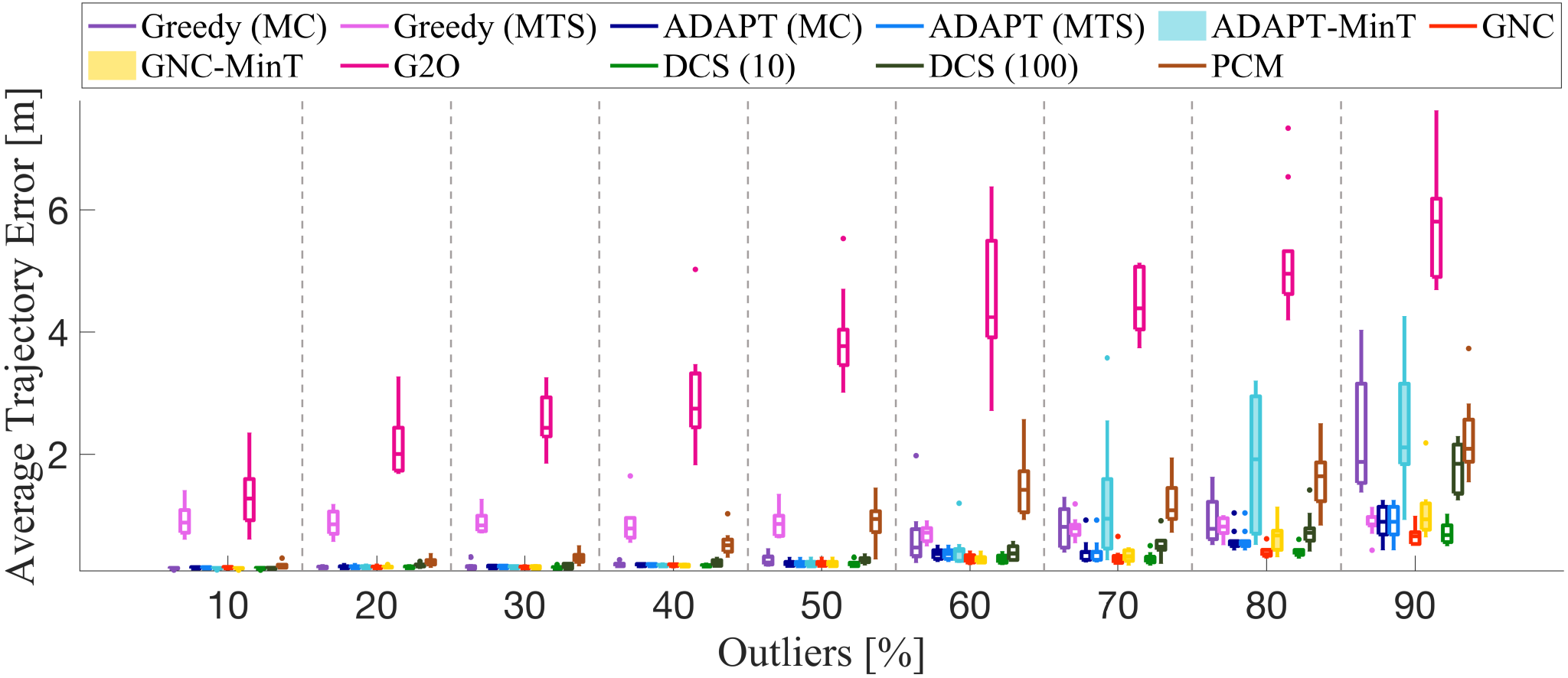} \\
			\end{minipage}
		& \mpMidSpaceTwo
			\begin{minipage}{\mpColTwo}%
        \centering%
        \includegraphics[width=\columnwidth]{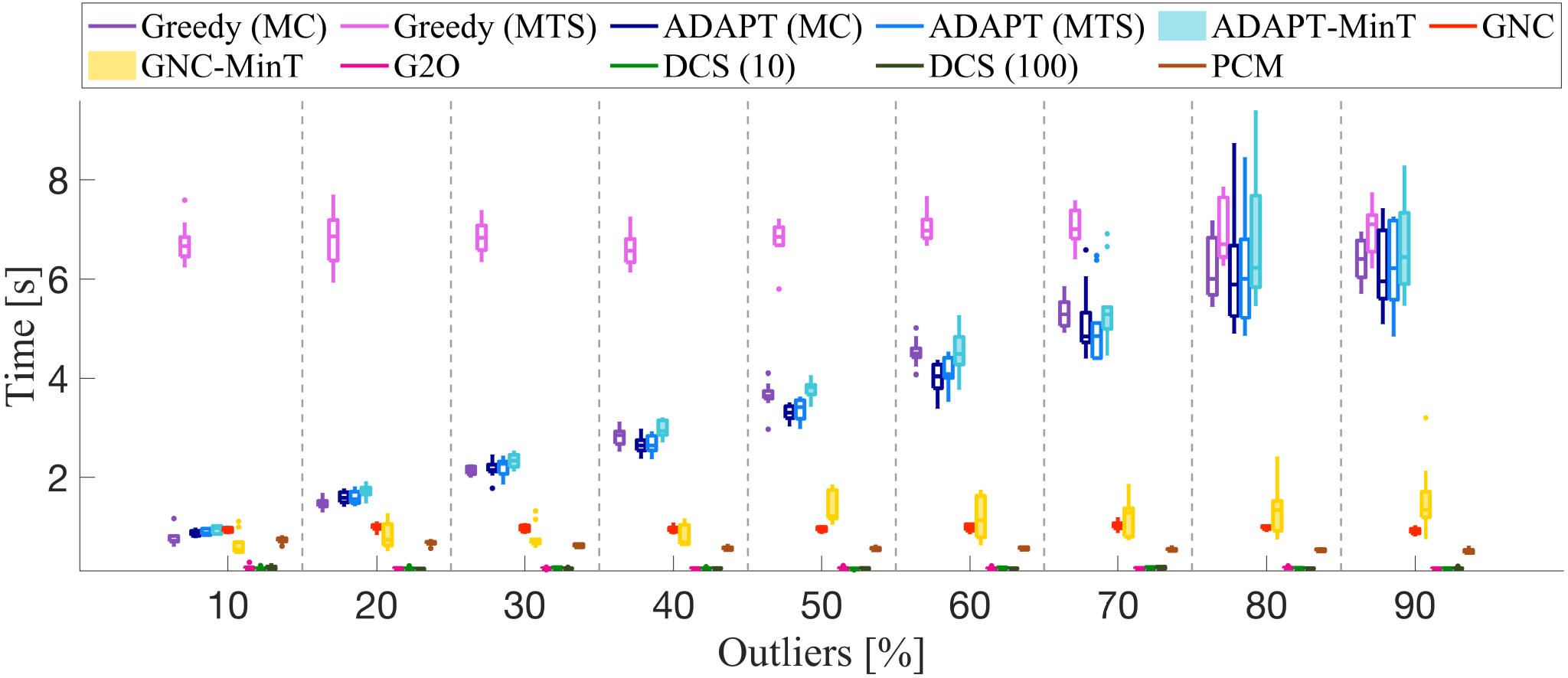} \\
			\end{minipage}
		\end{tabular}
	\end{minipage}
	\mpPostSpace
	\caption{ {\bf 2D SLAM (Grid).} Average Trajectory Error (ATE) and running time of the proposed algorithms compared to state-of-the-art techniques on a synthetic grid dataset for increasing outliers.}
	\label{fig:slam_grid}
	\vspace{-5mm} 
	\end{center}
\end{figure*}

\begin{figure*}[ht!]
	\begin{center}
	\begin{minipage}{\textwidth}
	\begin{tabular}{ccc}%
		%%%%%%%%%%%%%%%%%%%%%%%%%%%%%%%%%%%%%%%%%%%%%%%%%%%%%%%%%%%%%%%%%%%%%%%%%%%%%%%%%%%%%%%%%%%%%%%%%%%%%%%%%
		\mpPreSpace
			\begin{minipage}{\mpColTwo}%
        \centering%
        \includegraphics[width=\columnwidth]{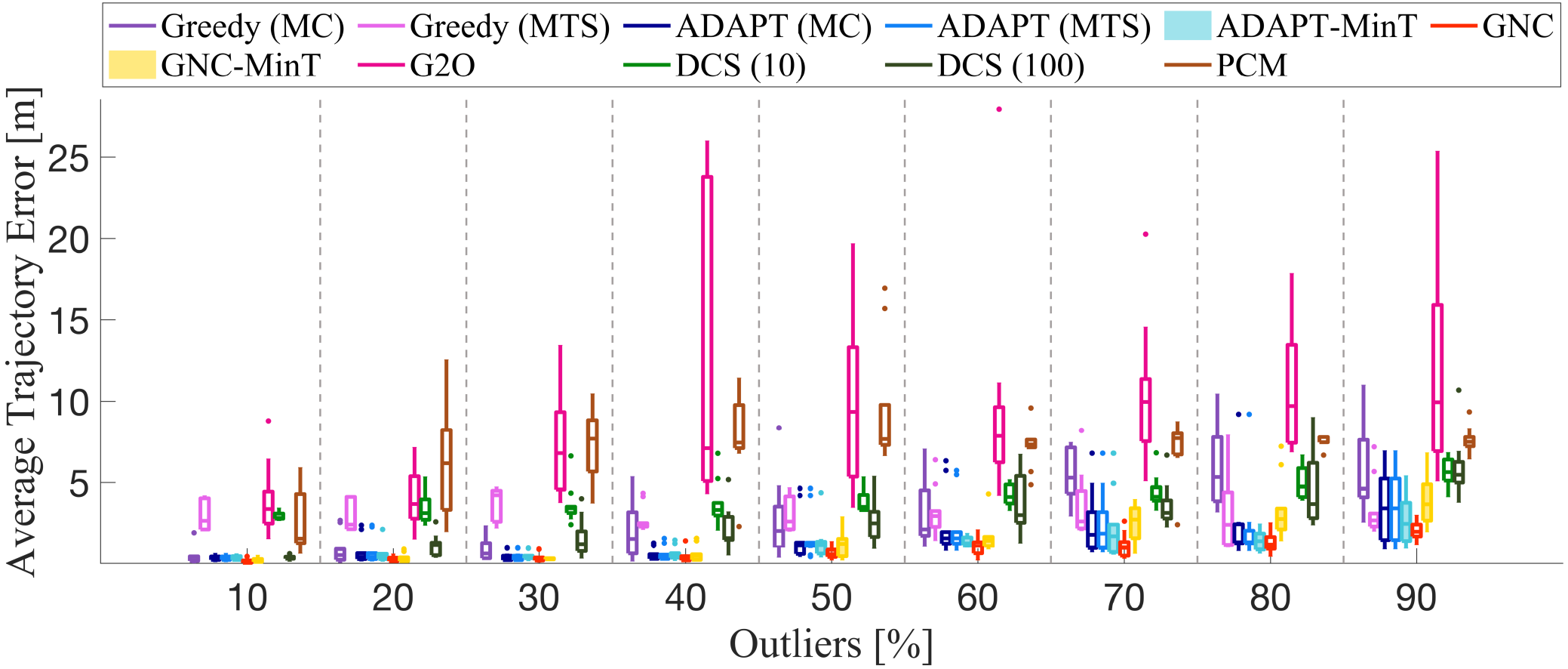} \\
			\end{minipage}
		& \mpMidSpaceTwo
			\begin{minipage}{\mpColTwo}%
        \centering%
        \includegraphics[width=\columnwidth]{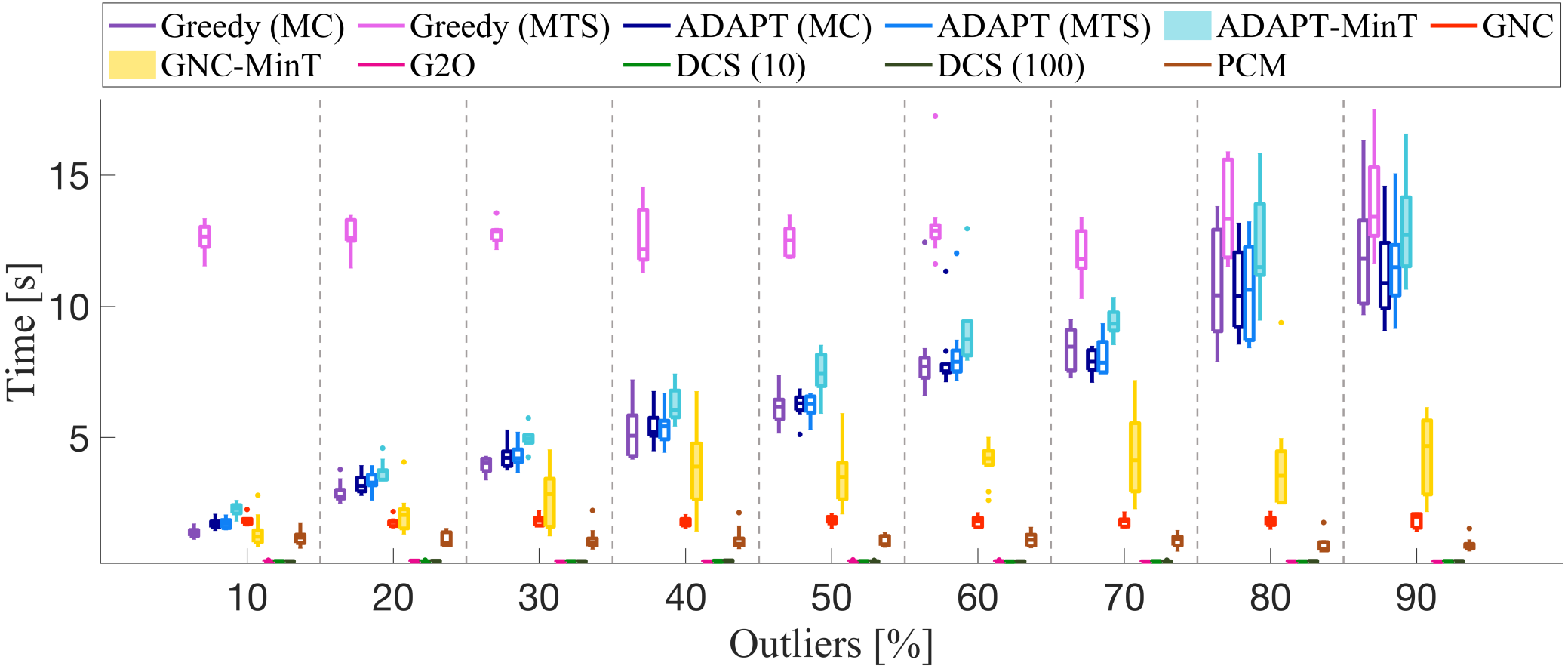} \\
			\end{minipage}
		\end{tabular}
	\end{minipage}
	\mpPostSpace
	\caption{ {\bf 2D SLAM (\scenario{CSAIL}).} Average Trajectory Error (ATE) and running time of the proposed algorithms compared to state-of-the-art techniques on the \scenario{CSAIL} dataset for increasing outliers.}
	\label{fig:slam_csail}
	\vspace{-5mm} 
	\end{center}
\end{figure*}
%!TEX root = ../../main.tex

\begin{figure*}[ht!]
	\begin{center}
	\begin{minipage}{\textwidth}
	\begin{tabular}{ccc}%
		%%%%%%%%%%%%%%%%%%%%%%%%%%%%%%%%%%%%%%%%%%%%%%%%%%%%%%%%%%%%%%%%%%%%%%%%%%%%%%%%%%%%%%%%%%%%%%%%%%%%%%%%%
		\mpPreSpace
			\begin{minipage}{\mpColTwo}%
        \centering%
        \includegraphics[width=\columnwidth]{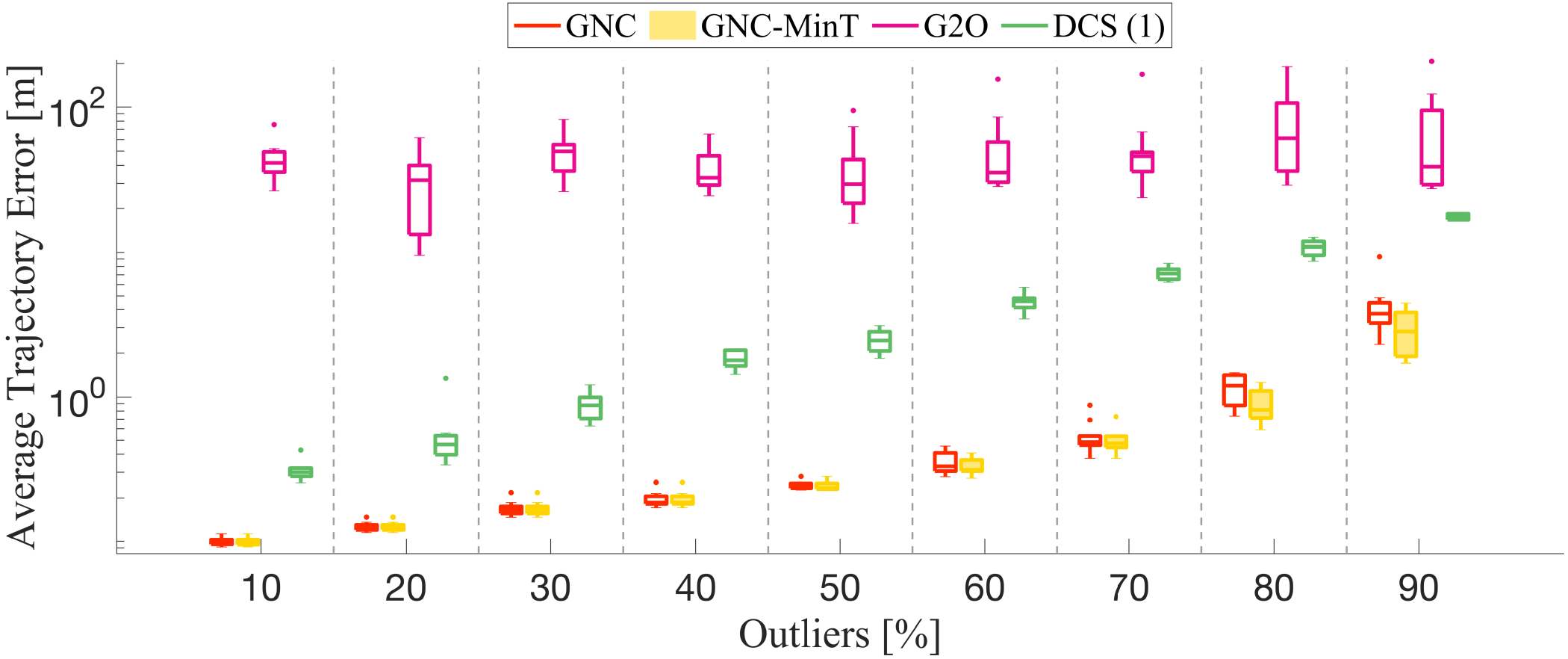} \\
			\end{minipage}
		& \mpMidSpaceTwo
			\begin{minipage}{\mpColTwo}%
        \centering%
        \includegraphics[width=\columnwidth]{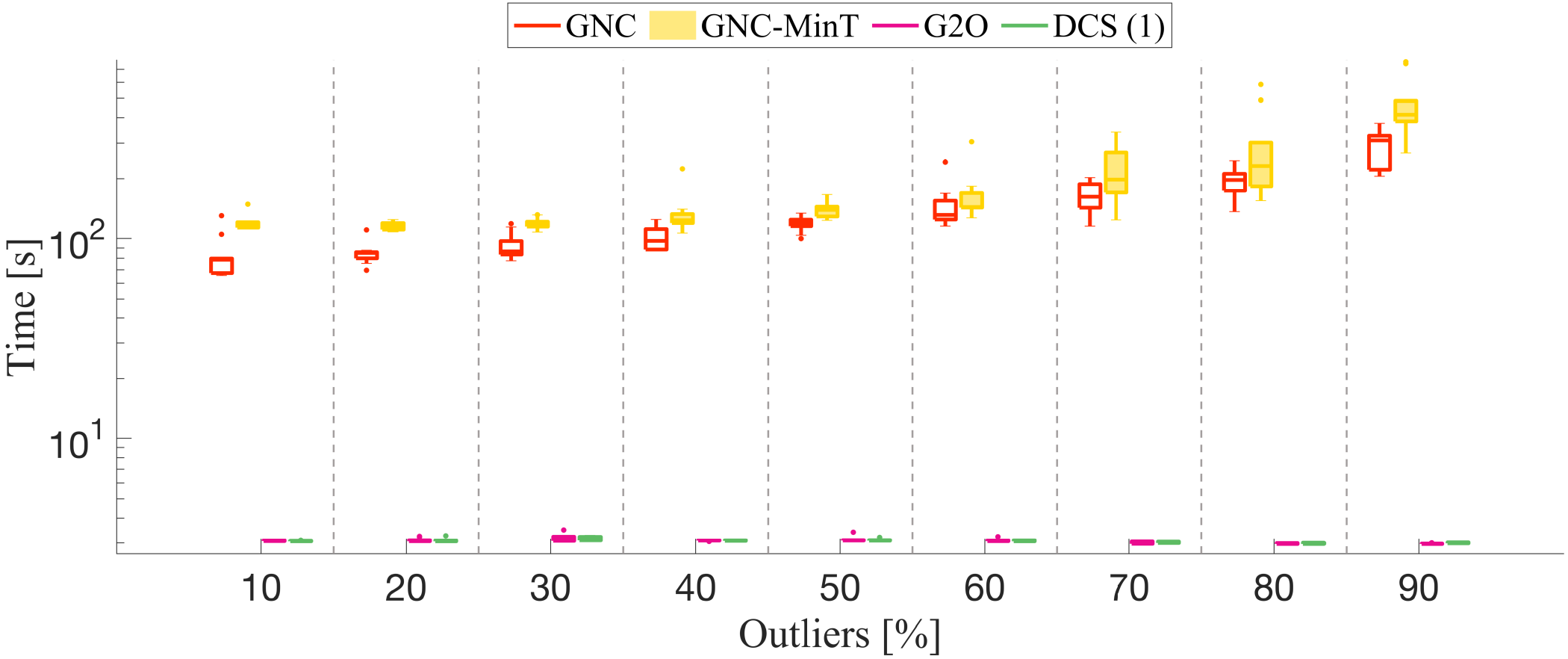} \\
			\end{minipage}
		\end{tabular}
	\end{minipage}
	\mpPostSpace
	\caption{ {\bf 3D SLAM (\scenario{Sphere}).} Average Trajectory Error (ATE) and running time of the proposed algorithms compared to state-of-the-art techniques on a synthetic \scenario{Sphere} dataset for increasing outliers.}
	\label{fig:slam_sphere}
	\vspace{-5mm} 
	\end{center}
\end{figure*}
%!TEX root = ../../main.tex

\begin{figure*}[ht!]
	\begin{center}
	\begin{minipage}{\textwidth}
	\begin{tabular}{ccc}%
		%%%%%%%%%%%%%%%%%%%%%%%%%%%%%%%%%%%%%%%%%%%%%%%%%%%%%%%%%%%%%%%%%%%%%%%%%%%%%%%%%%%%%%%%%%%%%%%%%%%%%%%%%
		\mpPreSpace
			\begin{minipage}{\mpColTwo}%
        \centering%
        \includegraphics[width=\columnwidth]{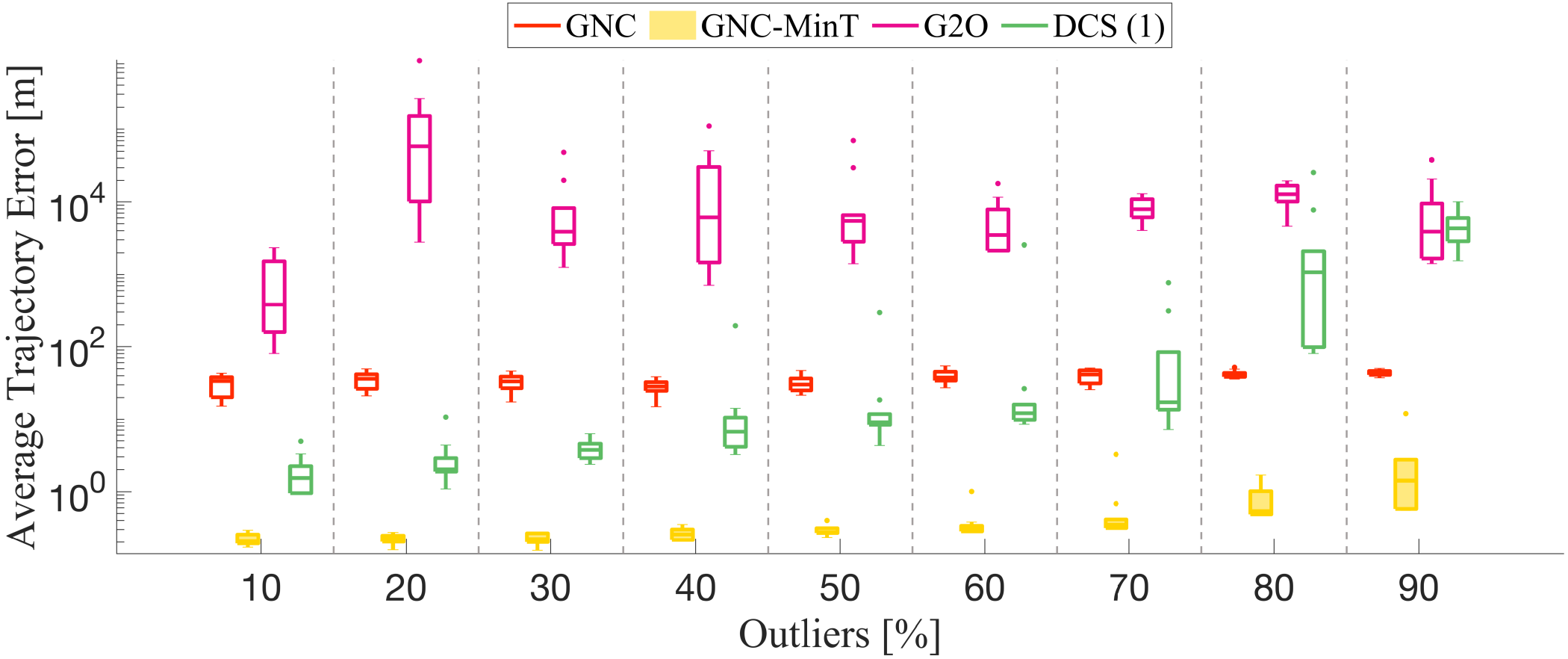} \\
			\end{minipage}
		& \mpMidSpaceTwo
			\begin{minipage}{\mpColTwo}%
        \centering%
        \includegraphics[width=\columnwidth]{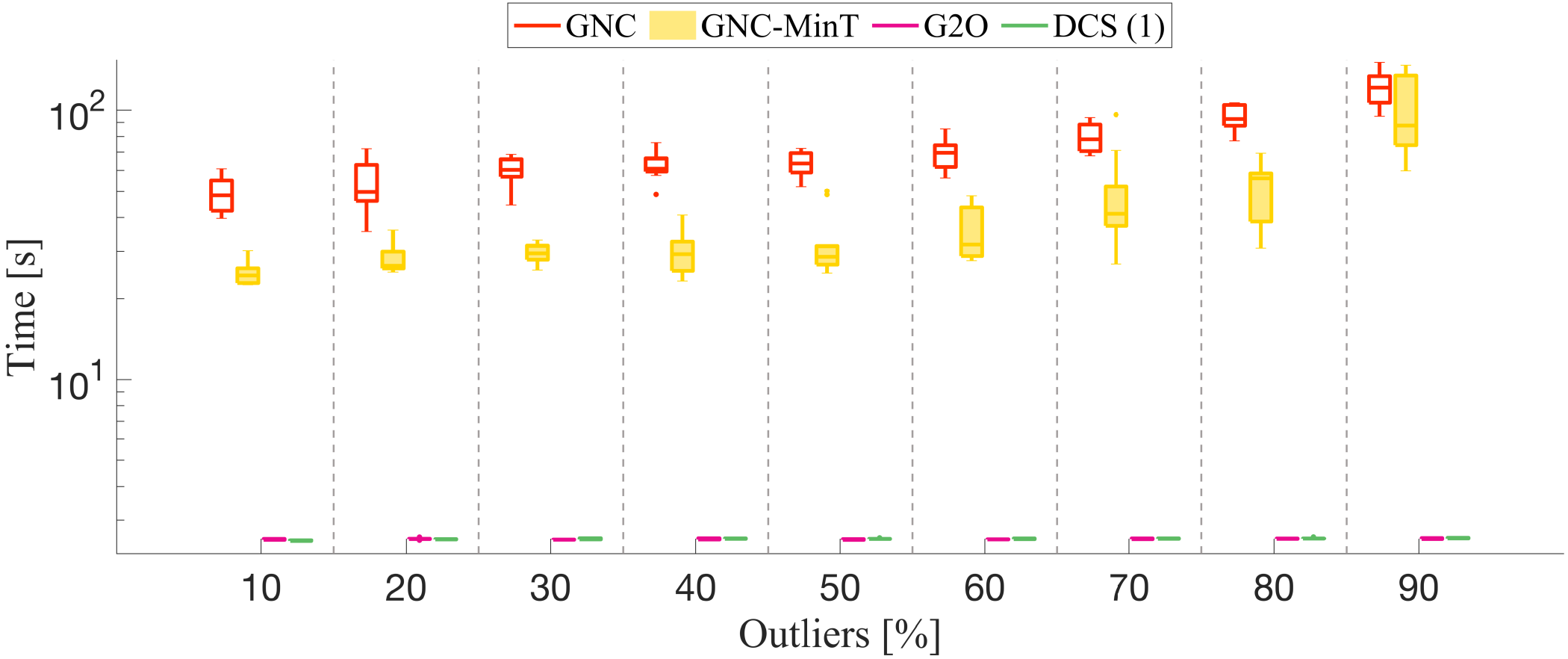} \\
			\end{minipage}
		\end{tabular}
	\end{minipage}
	\mpPostSpace
	\caption{ {\bf 3D SLAM (\scenario{Garage}).} Average Trajectory Error (ATE) and running time of the proposed algorithms compared to state-of-the-art techniques on the \scenario{Garage} dataset for increasing outliers.}
	\label{fig:slam_garage}
	\vspace{-5mm} 
	\end{center}
\end{figure*}
% ===================================

\subsection{Pose Graph Optimization (\PGO)}
\label{sec:exp-slam}

 Pose Graph Optimization (\PGO) is a common backend for Simultaneous Localization and Mapping (SLAM)~\cite{Cadena16tro-SLAMsurvey}.
\PGO estimates a set of poses $(\vt_i,\MR_i)$, $i\in \meas$ from pairwise relative pose measurements $(\bar{\vt}_{ij}, \bar{\MR}_{ij})$ (potentially corrupted with outliers).
The residual error is the distance between the expected relative pose and the relative measurements:
\begin{equation*} % g2o cost function
\sqrt{
  \| \mathrm{Log}(\bar{\MR}_{ij}\tran \MR_i\tran \MR_j) \|_{\MOmega_{ij}^R}^2 + 
  \| \bar{\MR}_{ij}\tran (\bar{\vt}_{ij} - \MR_i\tran(\vt_i - \vt_j)) \|_{\MOmega_{ij}^t}^2 
}
\end{equation*} 
where $\MOmega_{ij}^R$ and $\MOmega_{ij}^t$ are respectively the known rotation and translation measurement information matrix.
For a vector $\va$, the symbol $\|\va\|^2_\MOmega$ denotes the standard Mahalanobis norm: $\|\va\|^2_\MOmega = \va\tran \MOmega \va$. 
The $\mathrm{Log}(\cdot)$ denotes the logarithm map for the rotation group, which, roughly speaking, converts a rotation matrix to a vector (in 3D) or  to a scalar (in 2D).\footnotemark

In the outlier free case, \sesync~\cite{Rosen18ijrr-sesync} provides a global solver for \PGO, and we have used 
it in our 2D SLAM experiments in~\cite{Tzoumas19iros-outliers}. \footnotetext{For simplicity, here we use the geodesic distance $\|\mathrm{Log}(\bar{\MR}_{ij}\tran \MR_i\tran \MR_j) \|$, while alternative rotation distances are often used in \PGO, see~\cite{Carlone15icra-initPGO3D,Rosen18ijrr-sesync}.}
However, \sesync becomes too slow in the 3D SLAM tests considered in this paper: rather than a limitation of \sesync, 
this follows from the fact that in early iterations, both \adapt and \gnc (as well as their \autotunned variants) 
solve problems with many outliers; in these cases, \sesync's relaxation is not tight,\footnote{Indeed, it has been observed that the presence of large noise can easily induce 
failures in relaxations of 3D SLAM~\cite{Carlone15iros-duality3D}, while their 2D counterparts are observed to remain tight in the presence of relatively large noise~\cite{Carlone16tro-duality2D}.} and \sesync tends to perform
 multiple steps in the Riemannian staircase~\cite{Rosen18ijrr-sesync}, becoming impractical.

To circumvent these issues, instead of \sesync, we use \gtwoo~\cite{Kuemmerle11icra}, which is a local solver for \PGO, and use the odometry as initial guess.
We remark this option is only viable when the odometric guess is available and considered reliable.
\arxivVersion{Fig.~\ref{fig:g20_vs_sesync} in the appendix}{Fig.~11 in the appendix of~\cite{\arxivThisPaper}} compares the use of \sesync and~\gtwoo within our algorithms and shows the two achieve 
comparable performance when the odometric guess is reliable.

\myParagraph{Experimental Setup}
We test the performance of our algorithms on synthetic and real datasets for 2D and 3D \PGO.
We use a synthetic grid~\cite{Tzoumas19iros-outliers}, and \scenario{CSAIL}~\cite{Carlone14ijrr-lagoPGO2D} in 2D, and
a synthetic sphere, and \scenario{Garage}~\cite{Carlone15icra-initPGO3D} in 3D.
We compute statistics over 10 Monte Carlo runs, with increasing amounts of outliers.
At each Monte Carlo run, we spoil existing loop closures with random outliers. 
We consider odometric measurements as inliers and use the odometry as initial guess for \gtwoo. 
Since $d=3$ in 2D \SLAM, $\inthr = \sqrt{\chiSqinv(0.99, 3)} = 3.3682$, and since $d=6$ in 3D \SLAM, $\inthr = \sqrt{\chiSqinv(0.99, 6)} = 4.1$.\footnote{In \SLAM we do not need multiply by the covariance because the objective function performs a whitening transformation via the information matrix.}
\!Regarding \gncfree and \adaptfree, we normalize the measurements' covariance matrices provided by each dataset to 
simulate the case in which the covariances are unknown, 
and we set $\noiseupbound = 1$m and $\noiselowbound = 0.01$m.\arxivVersion{\footnote{In detail: we normalize each measurement's information matrix $\Omega_{ij}$ by a factor $\alpha_{ij}$ that represents the mean 
information (inverse variance) of the translation measurements (we ignore the effect of the rotation, since it has 1-2 orders of magnitude smaller errors).}}{}
% Particularly, $\alpha_{ij} = \trace{\Sigma_{ij}^{tr}}/d_{\text{dim}}$, where $d_{\text{dim}}$ is the dimension of the translation ($d_{\text{dim}}=2$ in 2D, and $d_{\text{dim}}=3$ in 3D), and $\Sigma_{ij}^{tr} = \left(\Omega_{ij}^{tr}\right)^{-1}$, where $\Omega_{ij}^{tr}$ is the information matrix of the relative translation measurement between the poses $i$ and $j$.}
% To compensate for the whitening 
% Therefore, we normalize each measurement's information matrix $\Omega_{ij}$ with a factor $\alpha_{ij}$ that represents the mean variance on the translation (we ignore the effect of the rotation, since it has 2 to 3 orders smaller variance).  Particularly, $\alpha_{ij} = \trace{\Sigma_{ij}^{tr}}/d_{\text{dim}}$, where $d_{\text{dim}}$ is the dimension of the translation ($d_{\text{dim}}=2$ in 2D, and $d_{\text{dim}}=3$ in 3D), and $\Sigma_{ij}^{tr} = \left(\Omega_{ij}^{tr}\right)^{-1}$, where $\Omega_{ij}^{tr}$ is the information matrix of the relative translation measurement between the poses $i$ and $j$.
% The information matrices being normalized, we set $\noiseupbound = 1$ meter and $\noiselowbound = 0.01$ meters.
%In \SLAM, we do not need to multiply the inlier threshold by $\sigma$ since the normalization takes place in the cost function.

We benchmark our algorithms against (i) \gtwoo~\cite{Kuemmerle11icra}, 
(ii) \emph{dynamic covariance scaling} (\dcs)~\cite{Agarwal13icra}, and
(iii) \emph{pairwise consistent measurement set maximization} (\pcm)~\cite{Mangelson18icra}.
The performance of \dcs is fairly sensitive to the choice of the kernel size $\Phi$, which is a parameter in the algorithm: we tested different kernel sizes $\Phi = \{1,10,100\}$ for \dcs, and we used the same $\inthr$ for \gnc, \adapt, and \pcm. 
For clarity of visualization, we only report the best two parameters (leading to smallest errors) for \dcs in the figures. 

\myParagraph{2D \PGO Results}
Fig.~\ref{fig:slam_grid} shows the Average Trajectory Error (ATE) and the running time for the synthetic \scenario{grid}.
\gtwoo is a non-robust solver, and performs poorly even when few outliers are present.
\adaptmc and \adaptmts outperform the \greedy algorithm.
\gnc outperforms the state of the art, and is robust to 90\% of outliers.
\gncfree is also robust up to 90\% of outliers, outperforming \adaptfree, which breaks at 70\% of outliers.
\dcs\!\scenario{(10)}  has similar performance to \gnc, being robust until 90\% of outliers;  
\dcs\!\scenario{(100)} degrades with increasing number of outliers, stressing the importance of parameter tuning in \dcs.
\pcm starts degrading at relatively low outlier rates.
Fig.~\ref{fig:slam_grid} shows that the runtimes of \gnc,  \gncfree, \gtwoo, \dcs, and \pcm are roughly constant, 
while \adapt's and \adaptfree's runtime grows linearly in the number of outliers.

Fig.~\ref{fig:slam_csail} shows the ATE and the running time for the \scenario{CSAIL} dataset.
All \adapt, \adaptfree, \gnc, and \gncfree outperform the state of the art, and are robust against 90\% of outliers.
\dcs starts breaking at 50\% of outliers.
\pcm and \gtwoo perform poorly across the whole spectrum.
{Both \adaptfree and \gncfree perform similarly to \adapt and \gnc, although being \autotunned algorithms.}
Similarly to \scenario{grid}, the runtimes of \gnc, \gncfree, \gtwoo, \dcs, and \pcm are roughly constant; \adapt's and \adaptfree's grow linearly.

\myParagraph{3D \PGO Results}
Fig.~\ref{fig:slam_sphere} and Fig.~\ref{fig:slam_garage} show the ATE and the running time in the case of \scenario{Sphere} and \scenario{Garage}, respectively (both in log-scale). 
We omit \greedy, \adapt, \adaptfree, and \pcm because their running times become impractical for these datasets (more than 10 minutes per run).
%their higher running times.
% ; notably, however, \adapt's and \adaptfree's ATE is on average the same as \gncfree's.
% % , and the fact that they have been already observed to be less competitive.
In both \scenario{Sphere} and \scenario{Garage}, we observe that \gnc and \gncfree outperform \dcs and \gtwoo,  regardless of \dcs's parameter choice.  Importantly, \gncfree outperforms \gnc in \scenario{Garage}.
 %, even though it is \autotunned.
 The covariances are unreliable in the \garage dataset, hence causing \gnc to set an incorrect $\inthr$.
 On the other hand, \gncfree is able to \emph{infer} the correct $\inthr$ and ensure accurate estimation.
% The covariances in the \garage dataset contains wrong covariance matrices, making the way we set $\inthr$ in the paper inaccurate, and causing \gnc to fail. Notwithstanding, \gncfree, being able to infer the scale of the covariance, outperforms \gnc and all other algorithms
% \footnote{The \garage dataset contains wrong covariance matrices, making the way we set $\inthr$ in the paper inaccurate, and causing \gnc to fail. Notwithstanding, \gncfree, being able to infer the scale of the covariance, outperforms \gnc and all other algorithms.}}
% \dcs has a constant running time.
\gnc's and \gncfree's running times slightly increase with increasing number of outliers, while \dcs's and \gtwoo's are constant.

Qualitative results are given in Fig.~\ref{fig:introSummary}.
%!TEX root = ../main.tex

\section{Extended Literature Review}
\label{sec:relatedWork}

We extend the literature review in Section~\ref{sec:intro}, to discuss outlier-robust estimation in robotics and computer vision (Section~\ref{sec:relatedWork_robotComp}), and in statistics and control (Section~\ref{sec:relatedWork_statistics}).

%%%%%%%%%%%%%%%%%%%%%%%%%%%%%%%%%%%%%%%%%%%%%%%%%%%%%%%%%%%%%
\subsection{Outlier-robust Estimation in Robotics and Computer Vision}\label{sec:relatedWork_robotComp}

Outlier-robust estimation has been an active research area in robotics and computer vision~\cite{Meer91ijcv-robustVision,Stewart99siam-robustVision,Bosse17fnt}. 
Two of the predominant paradigms to gain robustness against outliers are \emph{consensus maximization}~\cite{Chin17slcv-maximumConsensusAdvances} and \emph{M-estimation}~\cite{Bosse17fnt}. In both paradigms, the literature is mainly divided into (i) \emph{fast heuristics}, algorithms that are efficient but provide little performance guarantees, and (ii) \emph{global solvers}, algorithms that offer optimality guarantees but scale poorly with the problem size.

{\bf Fast Heuristics.} For consensus maximization, \ransac~\cite{Fischler81,Barath20cvpr-magsac} has been a widely adopted heuristic due to its efficiency and effectiveness in the low-outlier regime. Recently, Tzoumas \etal~\cite{Tzoumas19iros-outliers} proposed \adapt for \emph{minimally trimmed squares} (\mts) estimation, a formulation that bears similarity with consensus maximization (\cf~Section~\ref{sec:formulation}). 
For M-estimation, local nonlinear optimization is typically employed, which relies on the availability of a good initial guess~\cite{Schonberger16cvpr-SfMRevisited,Chatterjee13iccv}. Instead, the recently proposed \GNC algorithm by Yang \etal~\cite{Yang20ral-GNC} provides a method for solving M-estimation without requiring an initial guess (also see~\cite{Zhou16eccv-fastGlobalRegistration}). 
Recently, Barron~\cite{Barron19cvpr-adaptRobustLoss} proposes a single parametrized function that generalizes a family of robust cost functions in M-estimation. Chebrolu~\etal~\cite{Chebrolu2020arxiv-adaptiveRobustKernels} design an expectation-maximization algorithm to simultaneously estimate the unknown quantity $\vxx$ and choose the best robust cost $\rho$ in eq.~\eqref{eq:robustEstimation}. These algorithms, however, still rely on an estimate of the inlier noise threshold $\inthr$.

{\bf Global Solvers}. Global solvers essentially perform exhaustive search to ensure global optimality. For instance, \emph{branch-and-bound} (\BnB) has been exploited to globally solve consensus maximization in several low-dimensional perception tasks~\cite{Bazin14eccv-robustRelRot,Hartley09ijcv-globalRotationRegistration,Zheng11cvpr-robustFitting,Li09cvpr-robustFitting,Speciale17cvpr-consensusMaximization,Chin16cvpr-outlierRejection,Izatt17isrr-MIPregistration,Bustos18pami-GORE,Yang2014ECCV-optimalEssentialEstimationBnBConsensusMax,Yang16pami-goicp}. Despite its global optimality guarantees, \BnB has  exponential running time in the worst case. 
It is also possible to globally solve consensus maximization and M-estimation by enumerating all possible minimizers~\cite{Enqvist12eccv-robustFitting,Olsson08cvpr-polyRegOutlier}. However, these algorithms are close to exhaustive search and do not scale to high-dimensional problems. 

%{\bf Certifiable Algorithms}. 
\emph{Certifiably robust} algorithms are a class of global solvers that have been recently shown to strike a good balance between computational complexity and 
global optimality~\cite{Yang20neurips-certifiablePerception,Yang20tro-teaser}. Certifiable algorithms relax non-convex robust estimation problems into convex \emph{semidefinite programs} (SDP), whose solutions can be obtained in polynomial time and provide readily checkable \emph{a posteriori} global optimality certificates~\cite{Yang19iccv-QUASAR,Carlone18ral-robustPGO2D,Lajoie19ral-DCGM,Yang19rss-teaser}. Although solving large-scale SDPs is computationally expensive, recent work has shown that optimality certification (\ie~verifying the global optimality of candidate solutions returned by fast heuristics) can scale to large problems by leveraging efficient first-order methods~\cite{Yang20neurips-certifiablePerception}. 

Finally, we note that adding a preprocessing layer to prune outliers can significantly boost the performance of robust estimation using consensus maximization, M-estimation, and certifiable algorithms~\cite{Bustos18pami-GORE,Yang20tro-teaser,Parra19arXiv-practicalMaxClique}.

Representative outlier-robust methods for registration, shape alignment, and \SLAM, are discussed below. 

%%%%%%%%%%%%%%%%%%%%%%%%%%%%%%%%%%%%%%%%%%%%%%%%%%%%%%%%%%%%%%%%%%%%%%%%%%%%%%%%%%%%%%%%%%%%%%%%%%%%
\myParagraph{Robust Registration} 
% The goal is to find the rigid transformation (translation and rotation) that best aligns two point clouds or a point cloud and a 3D mesh. 
Rigid registration looks for the transformation that best aligns two point clouds or a point cloud and a 3D mesh.
We review correspondence-based registration methods, while we refer the reader to~\cite{Yang20tro-teaser} for a broader
review on 3D registration,  including \emph{Simultaneous Pose and Correspondence} methods (\eg ICP~\cite{Besl92pami}).
Correspondence-based registration methods first 
extract and match features in the two point clouds, using 
hand-crafted~\cite{Rusu09icra-fast3Dkeypoints} or deep-learned~\cite{gojcic19cvpr-3Dsmoothnet,choy19iccv-FCGF} features.
Then, they solve an estimation problem to compute the rigid transformation that best aligns the set of corresponding features. 
In the presence of incorrect correspondences (\ie~outliers), one typically 
resorts to \ransac~\cite{Fischler81,Chen99pami-ransac}, along with a 3-point minimal solver~\cite{Arun87pami,Horn87josa}.
 In the high-outlier regime (\eg~above $80\%$), \ransac tends to be slow and brittle~\cite{Speciale17cvpr-consensusMaximization,Bustos18pami-GORE}.  Thereby, recent approaches adopt either M-estimation or consensus maximization. Zhou\setal~\cite{Zhou16eccv-fastGlobalRegistration} propose \emph{fast global registration}, which minimizes the Geman-McClure robust cost function using \GNC. Tzoumas~\etal use \adapt~\cite{Tzoumas19iros-outliers}, and Yang~\etal use \GNC~\cite{Yang20ral-GNC} to solve point cloud registration with robustness against up to $80\%$ outliers.
Bazin\setal~\cite{Bazin12accv-globalRotSearch} employ \BnB to perform globally optimal rotation search (\ie~3D registration without translation).
 Parra\setal~\cite{Bustos18pami-GORE} add a preprocessing step, that removes gross outliers before \ransac or \BnB.
Yang and Carlone propose invariant measurements to decouple the rotation and translation estimation~\cite{Yang19rss-teaser}, and develop certifiably robust rotation search using semidefinite relaxation~\cite{Yang19iccv-QUASAR}. The joint use of fast heuristics (\eg~\GNC) and optimality certification for both point cloud registration and mesh registration has been demonstrated in~\cite{Yang20tro-teaser,Yang20neurips-certifiablePerception}. The registration approach~\cite{Yang20tro-teaser} has been shown to be robust to $99\%$ outliers.

%%%%%%%%%%%%%%%%%%%%%%%%%%%%%%%%%%%%%%%%%%%%%%%%%%%%%%%%%%%%%%%%%%%%%%%%%%%%%%%%%%%%%%%%%%%%%%%%%%%%
\myParagraph{Robust Shape Alignment} 
Shape alignment consists in estimating the absolute camera pose given putative correspondences between 2D image landmarks and 3D model keypoints (the problem is called 3D \emph{shape reconstruction} when the 3D model is unknown~\cite{Zhou17pami-shapeEstimationConvex,Zhou18PAMI-monocap,Yang20cvpr-shapeStar}). When a full camera perspective model is assumed, the problem is usually referred to as the \emph{perspective-$n$-point} (PnP) problem~\cite{Kneip2014ECCV-UPnP}. \ransac is again the go-to approach to gain robustness against outliers, typically in conjunction with a 3-point minimal solver~\cite{Gao03PAMI-P3P}. Ferraz~\etal propose an efficient robust PnP algorithm based on iteratively rejecting outliers via detecting large algebraic errors in a linear system~\cite{Ferraz14cvpr-fastRobustPnP}. When the 3D model is far from the camera center, a weak perspective camera model can be adopted~\cite{Zhou17pami-shapeEstimationConvex}, which leads to efficient robust estimation using \GNC~\cite{Yang20ral-GNC}. Yang and Carlone~\cite{Yang20neurips-certifiablePerception} develop optimality certification algorithms for shape alignment with outliers, and demonstrate successful application to satellite pose estimation.

%%%%%%%%%%%%%%%%%%%%%%%%%%%%%%%%%%%%%%%%%%%%%%%%%%%%%%%%%%%%%%%%%%%%%%%%%%%%%%%%%%%%%%%%%%%%%%%%%%%%
\myParagraph{Robust SLAM}
Outlier-robust SLAM has traditionally relied on M-estimators, see,~\eg\cite{Bosse17fnt}. 
% the resulting optimization is typically solved using local solvers. 
Olson and Agarwal~\cite{Olson12rss} use a max-mixture distribution to approximate multi-modal
measurement noise. 
S\"{u}nderhauf and Protzel~\cite{Sunderhauf12iros,Sunderhauf12icra} augment the problem with latent binary variables responsible for 
deactivating outliers.  
Tong and Barfoot~\cite{Tong11icra-robustSlam,Yong13astro-robustSLAM} propose
algorithms to classify outliers via Chi-square statistical tests
that account for the effect of noise in the estimate.
Latif~\etal \cite{Latif12rss-rrr} propose \emph{realizing, reversing, and recovering}, which performs loop-closure outlier rejection, by clustering measurements together and checking for consistency using a
Chi-squared-based test.
%the chi-square inverse test as an outlier-free bound. 
Mangelson~\etal \cite{Mangelson18icra} propose a \emph{pair-wise consistency maximization} (\pcm) approach %, is proposed by .
for multi-robot SLAM.
Agarwal \etal~\cite{Agarwal13icra} propose \emph{dynamic covariance scaling} (\dcs), which adjusts the measurement covariances to reduce the influence of outliers.
Lee~\etal~\cite{Lee13iros} use expectation maximization. 
The papers above rely either on the availability of an initial guess for optimization, or on parameter tuning. 
Tzoumas~\etal propose \adapt~\cite{Tzoumas19iros-outliers}, and Yang~\etal propose \GNC~\cite{Yang20ral-GNC} to solve outlier-robust SLAM without initialization.
Recent work also includes convex relaxations for outlier-robust SLAM~\cite{Wang13ima,Carlone18ral-robustPGO2D,Arrigoni18cviu,Lajoie19ral-DCGM}. Lajoie\setal~\cite{Lajoie19ral-DCGM} provide
sub-optimality guarantees, \mbox{which however degrade with the quality of the relaxation.}  
%Additionally,~\cite{Lajoie19ral-DCGM}  requires parameter tuning.

%%%%%%%%%%%%%%%%%%%%%%%%%%%%%%%%%%%%%%%%%%%%%%%%%%%%%%%%%%%%%
\subsection{Outlier-robust Estimation in Statistics and Control}\label{sec:relatedWork_statistics}

Outlier-robust estimation has been also a subject of investigation in statistics and control~\cite{Huber64ams,Kalman60}, 
where  it finds applications to distribution learning~\cite{Diakonikolas16focs-robustEstimation}, linear decoding~\cite{Candes05tit}, and secure state estimation~\cite{Pasqualetti13tac-CPSsecurity}, among others.  

\myParagraph{Statistics} 
In its simplest form, outlier-robust estimation aims at learning the mean and covariance of an unknown distribution, given (i) a portion of 
independent and identically distributed~samples, and (ii) a portion of arbitrarily corrupted samples (outliers), where the percentage of outliers is assumed known.
Researchers provide polynomial time near-optimal algorithms~\cite{Diakonikolas16focs-robustEstimation, Liu19arxiv-TrimmedHardThresholding}.
In scenarios where one instead aims to estimate an unknown parameter given corrupted measurements, Rousseeuw~\cite{Rousseeuw11dmkd} propose \textit{linear trimmed squares} (\scenario{LTS}), which aims to minimize the cumulative inlier residual error given a known number of outliers. Similar greedy-like algorithms, that also assume a known number of outliers,  are the forward greedy by Nemhauser\setal~\cite{Nemhauser78mp-submodularity}, and forward-backward greedy by Zhang~\cite{Zhang11tit-ForwardBackGreedy}. Both algorithms have quadratic running time, which is prohibitive in high-dimensional robotics and computer vision applications, such as \SLAM.
In contrast to~\cite{Rousseeuw11dmkd,Nemhauser78mp-submodularity,Zhang11tit-ForwardBackGreedy}, the greedy algorithm proposed in~\cite{Liu18tsp-GreedyRobustRegression} considers the number of outliers to be unknown.  However, it still requires parameter tuning, this time for an inlier threshold parameter.

\myParagraph{Control} Outlier-robust estimation in control takes the form of secure state estimation in the presence of outliers, including adversarial measurement corruptions. Related works~\cite{Pasqualetti13tac-CPSsecurity,Mishra17tac-secureStateEstimation,Aghapour18cdc-outlierAccomodation} propose exponential-time algorithms, achieving exact state estimation when the inliers are noiseless.
%!TEX root = ../main.tex

\section{Conclusion} 
\label{sec:conclusion}

We investigated fundamental computational limits and general-purpose algorithms for outlier-robust estimation.
We proved that, in the worst-case, outlier-robust estimation is inapproximable even in quasi-polynomial time.
We reviewed and extended two robust algorithms, \adapt and \gnc, and established convergence results and connections
between the corresponding formulations. 
We proposed the first \autotunned algorithms, 
\adaptfree and \gncfree. These algorithms offer a new paradigm for resilient life-long estimation, 
being robust not only against outliers but also against unknown inlier noise statistics.
We theoretically grounded these algorithms by identifying probabilistic interpretations of maximum consensus and truncated least squares estimation.

The proposed algorithms are deterministic, scalable to problems with thousands of variables, and require no initial guess.
Moreover, they dominate the state of the art across several robot perception applications.
We believe the proposed approaches can be a valid replacement for \ransac, and 
constitute an important step towards parameter-free (auto-tuning) algorithms. 
In contrast to \adaptfree and \gncfree, \ransac is non-deterministic,   
requires a minimal solver, relies on careful parameter tuning, and its runtime increases exponentially with the 
percentage of outliers.
 
 The algorithms discussed in this paper (as well as the baselines we compared against) do not 
 guarantee convergence to optimal solutions (this is expected, due to the inapproximability result 
 in Theorem~\ref{th:hardness}). 
 The interested reader can find examples of failure modes in \arxivVersion{\ref{sec:limitations}}{\cite[Appendix~1]{\arxivThisPaper}}.
 Future work includes coupling the algorithms proposed in this paper
  with fast \emph{certifiers}~\cite{Yang20neurips-certifiablePerception} that can 
 detect and reject incorrect estimates.

\vspace{0mm}
\section*{Acknowledgments}
The authors would like to thank Jos\'e Mar\'ia Mart\'inez Montiel
 for the discussions about limitations and failure modes of greedy algorithms for outlier rejection.

\vspace{0mm}
%%%%%%%%%%%%%%%%%%%%%%%%%%%%%%%%%%%%
\setcounter{section}{0}
\renewcommand{\thesection}{Appendix \arabic{section}} 
\renewcommand{\thesubsection}{\thesection-\Alph{subsection}}
%\appendices

\arxivVersion{
	%!TEX root = ../main.tex

\section{Limitations}\label{sec:limitations}

We discuss failure modes of the proposed algorithms.

	%!TEX root = ../main.tex

\subsection{Limitations of \adapt and \adaptfree}

\omitted{We discuss failure modes of \adapt and \adaptfree.
While \adapt self-correction mechanism often improve the estimation sometimes it can harm the estimation, this can happen if \adapt converged to the wrong estimate (thus the outliers residuals look like inliers) or if there are too many outliers and some of them fit the current estimation.}

\omitted{\myParagraph{Effect of \thrdiscount} \adapt's \thrdiscount affects both the algorithm's runtime and the set selection:
if \thrdiscount is set lower, $\inlierNoiseAdapt\at{\iteration}$ will decrease faster (\cf \adapt's line~\ref{line:adapt-thresholdUpdate}), making \adapt to trim more measurements per iteration; this will typically mean that \adapt will terminate faster; however, more measurements may be removed than necessary, possibly resulting to poorer performance.   
%\omitted{If \thrdiscount is set to \num{1} then the algorithm will reject one single outlier per iteration that might increase the running time.}
Nevertheless, it is possible to find instances where a lower \thrdiscount results to better performance. }

\myParagraph{Inaccurate $\minnoisebound$ and $\theta$} If $\minnoisebound$ and $\theta$ are set too low, lower than their true values, \adapt will typically over-reject measurements.  Conversely, if they are set too high, \adapt is more likely to return sets containing outliers.  Both scenarios can result to less accurate estimates.

\omitted{In problems with too many outliers, \adapt is not able to recover the true solution.
The maximum number of tolerable outliers depends on the application.}

\begin{figure}[t!] 
	\begin{center}
		\begin{minipage}{\textwidth}
			\mpMidSpaceSix
			\begin{tabular}{p{\mpColSix}p{\mpColSix}p{\mpColSix}}%
				\begin{minipage}{\mpColSix}%
					\centering%
					\includegraphics[width=\columnwidth]{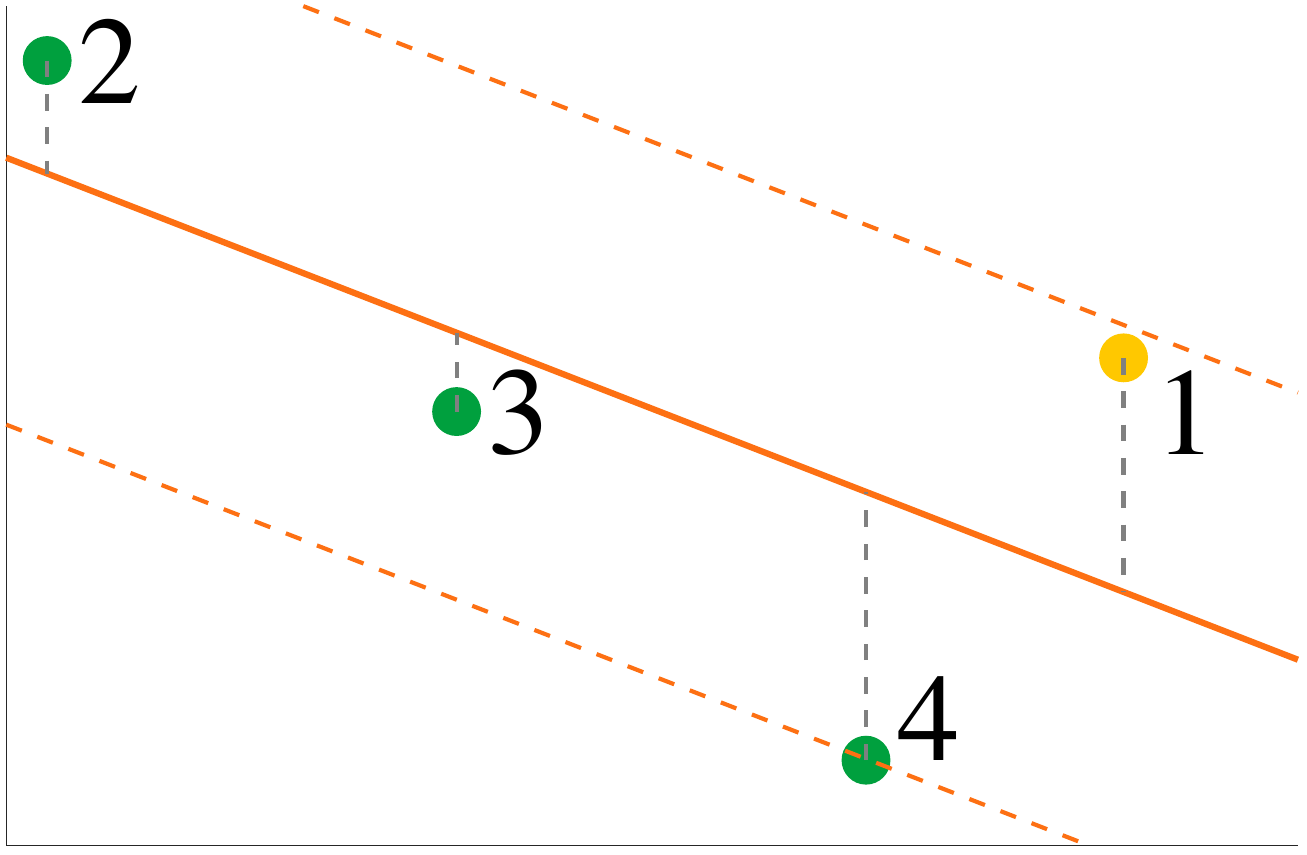}
				\end{minipage}
				& \mpMidSpaceSix
				\begin{minipage}{\mpColSix}%
					\centering%
					\includegraphics[width=\columnwidth]{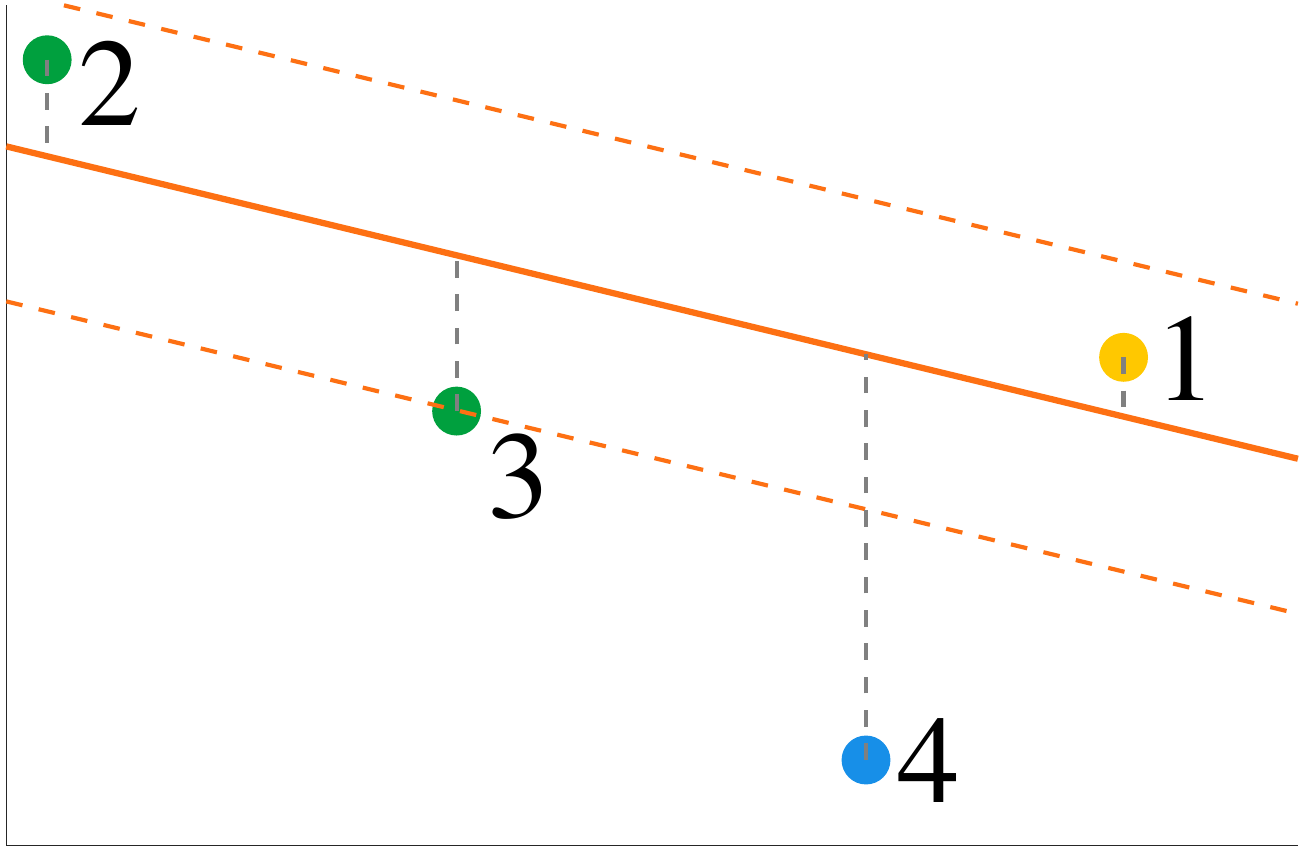}
				\end{minipage}
				& \mpMidSpaceSix\mpMidSpaceSix
				\begin{minipage}{\mpColSix}%
					\centering%
					\includegraphics[width=\columnwidth]{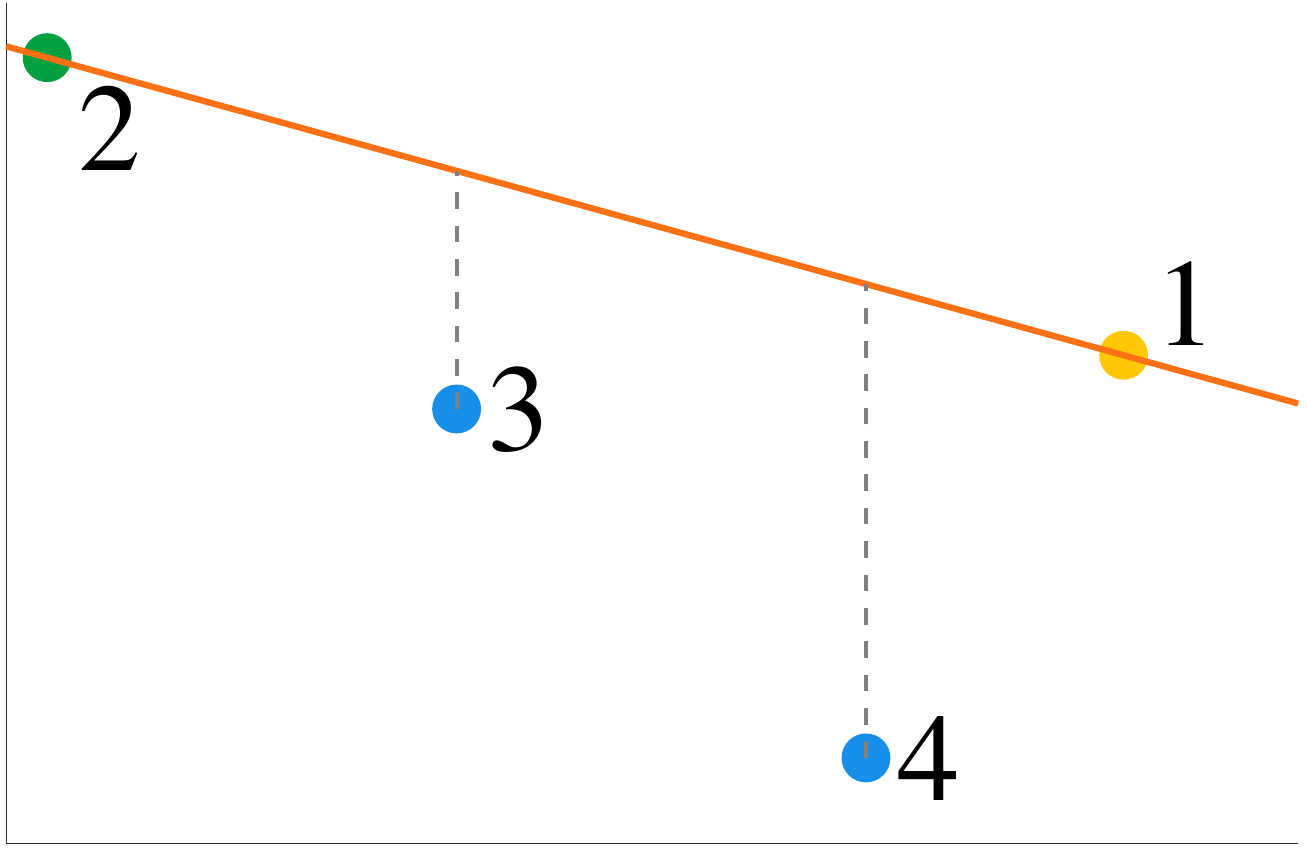}
				\end{minipage}
			\end{tabular}
		\end{minipage}
		\vspace{-3mm} 
		\caption{A single outlier (point 1) leads \adapt to the wrong solution.
			\label{fig:adapt_failure}} 
		\vspace{-2mm} 
	\end{center}
\end{figure}

\myParagraph{Adversarial Outliers} \adapt (and similarly \gnc) can fail due to adversarial outliers.  In Fig.~\ref{fig:adapt_failure}, we present such a scenario for a problem of linear regression, where there are three inliers (points 2-4) and one outlier (point 1).
%Particularly, we assume there are two residuals with similar magnitude, say $y_1 = \alpha y_2$ ($\alpha > 1$), where $y_1$ is an inlier and $y_2$ is the outlier while $y_3 \approx y_4 \approx 0$.
%If we set $\thrdiscount=1$, 
For appropriate $\thrdiscount$, \adapt first rejects the inlier point {4}, moving the new estimate (based on points 1-3) towards the outlier 1.
Then, \adapt rejects point 3, and, then, terminates.
\omitted{Conversely, if $\thrdiscount< 1/\alpha$, in one iteration \adapt will reject both measurements (\num{1} and \num{2}), in the next iteration the remaining inliers (now in the set of estimated inliers) will move the estimate closer to \num{1} that will be re-included in the set of inliers through the self-correction mechanism.}

\myParagraph{High Measurement Noise} 
%\adaptfree is affected by the same limitations of \adapt.  An additional failure mode can be caused by very noisy measurements.
High measurement noise can cause the cluster separation $\delta$, used in \adaptfree, to oscillate more than the chosen \convergthr, thus making the algorithm to reject more measurements than the true number of outliers.

\begin{figure*}[t!]
	\begin{center}
	\begin{minipage}{\textwidth}
	\begin{tabular}{ccc}%
		%%%%%%%%%%%%%%%%%%%%%%%%%%%%%%%%%%%%%%%%%%%%%%%%%%%%%%%%%%%%%%%%%%%%%%%%%%%%%%%%%%%%%%%%%%%%%%%%%%%%%%%%%
		\mpPreSpace
			\begin{minipage}{\mpColTwo}%
        \centering%
        \includegraphics[width=\columnwidth]{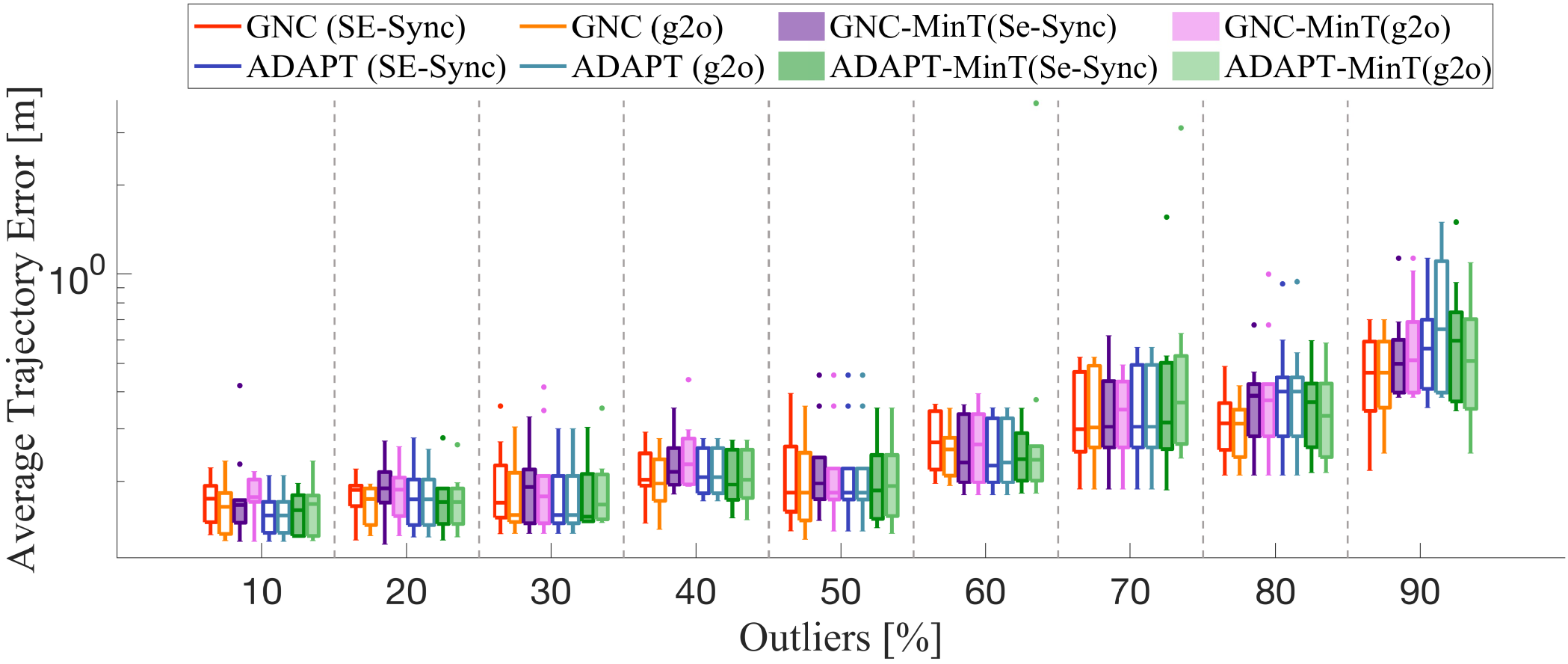} \\
			\end{minipage}
		& \mpMidSpaceTwo
			\begin{minipage}{\mpColTwo}%
        \centering%
        \includegraphics[width=\columnwidth]{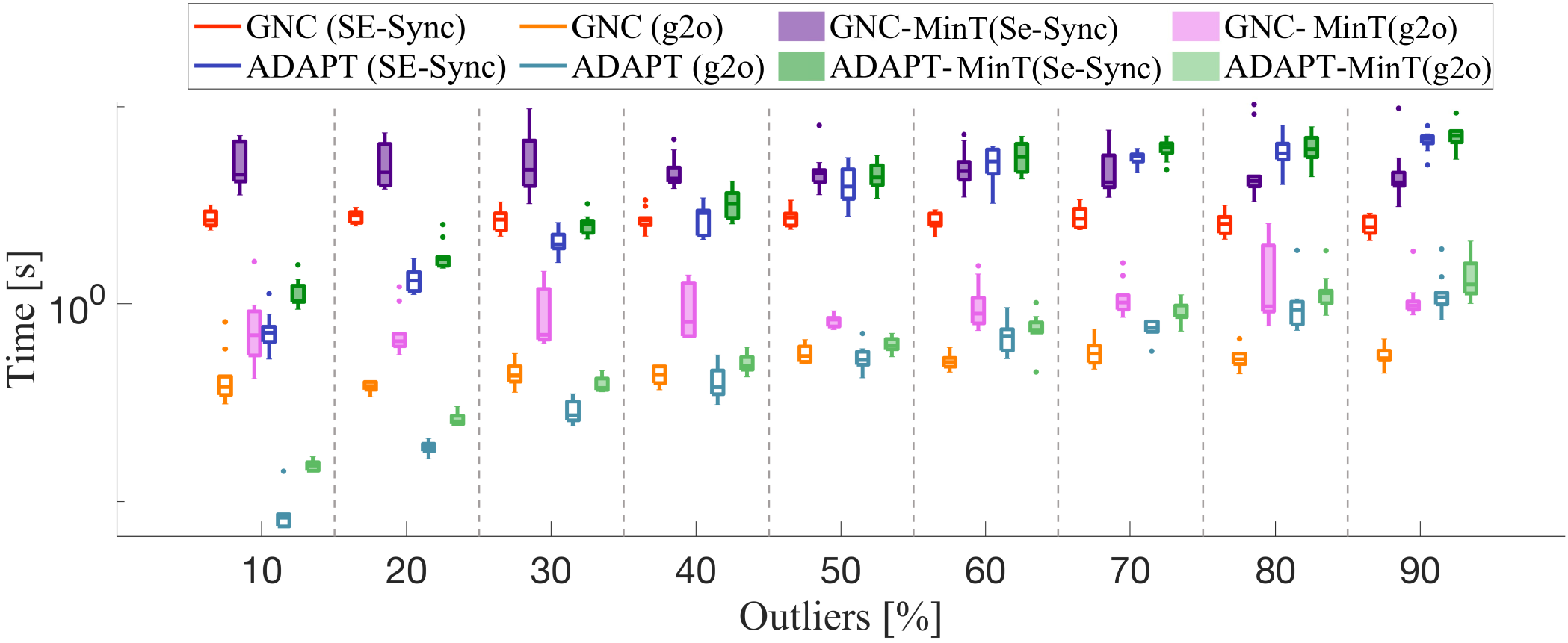} \\
			\end{minipage}
		\end{tabular}
	\end{minipage}
	\mpPostSpace
  \caption{
    Average Trajectory Error and Running time of the proposed algorithms on 2D SLAM (\scenario{Grid}) with two different non-minimal solvers: \gtwoo and \sesync.
    The average performance is comparable while \gtwoo offers a better running time.}
	\label{fig:g20_vs_sesync}
	\vspace{-5mm} 
	\end{center}
\end{figure*}
	
	%!TEX root = ../main.tex

\subsection{Limitations of \gnc and \gncfree}

\omitted{We analyze failure modes of \gnc and \gncfree.}

\omitted{Another possible failure mode for \gnc is over/under-rejection.}
\myParagraph{Inaccurate $\inthr$} 
If $\inthr$ is chosen lower than the real inlier threshold, then \gnc can reject more measurements than the true number of outliers.  Instead, if $\inthr$ is too high, then \gnc tends to reject less measurements, keeping outliers as inliers.  Both scenarios can result to less accurate estimates.

\omitted{In the case with too many outliers, \gnc is not able to recover the true solution.
The maximum number of tolerable outliers depends on the application.}

\myParagraph{Non-Gaussian Measurement Noise} If the residual's distribution is not close to a Gaussian, the $\chi^2$ fitness score may not accurately indicate the presence of outliers.  Thus, \gncfree may return less accurate estimates.

\myParagraph{Arbitrarily Low $\noiseupbound$, and Arbitrarily Large $\noiselowbound$}
If $\noiselowbound-\noiseupbound$ is unnecessarily large, then \gncfree, trying to find the true but unknown inlier threshold, will explore more $\inthr$ values, and, as a result, it will run for longer time.
Also,
{\gncfree stops as soon as the fitness score becomes worse  (\cf~\gncfree's lines~\ref{line:gncAuto-termination-nochange}-\ref{line:gncAuto-termination-convergence})}.  This point, however, may correspond to a local minima (thinking of the fitness score as a function of the inlier threshold guess).  Therefore, if the $\noiseupbound$ is unnecessarily high, there is a higher probability \gncfree stops prematurely.

\omitted{Finally, \gncfree can under- or over-estimate the true inlier threshold: suppose at some iteration $\iteration$, $\inthr\at{\iteration} > \inthr^\circ$, $\inthr^\circ$ being the true inlier threshold, and that the set $\inset\at{\iteration}$ still contains few outliers but all the residuals are smaller than $\inthr^\circ$, \ie $\displaystyle\max_{i\; \in \; \inset\at{\iteration}}\;\{\res(\vy_{i},\vxx\at{\iteration}) \;\text{s.t.}\;\res(\vy_{i},\vxx\at{\iteration}) \!<\!\inthr\at{\iteration}\} < \inthr^\circ$, then \gncfree will underestimate the inlier threshold $\inthr$.}

}{}

%!TEX root = ../main.tex

\section{Proof of results in Section~\ref{sec:formulation}}
\label{app:proofs-formulation}

\subsection{Proof of Proposition~\ref{th:mle_for_mc}}\label{app:proofs-formul-mle_for_mc}

We prove that  any optimal solution to \maxcon (eq.~\eqref{eq:maxConsensus}) is also an optimal solution to eq.~\eqref{eq:mle_for_mc}, and vice versa. To argue this, we use the method of contradiction. 

First, assume $(\vxx_\maxcon,\outset_\maxcon)$ is an optimal solution to \maxcon but not to eq.~\eqref{eq:mle_for_mc}, \ie there exists an optimal solution $(\vxx_{eq.\eqref{eq:mle_for_mc}},\outset_{eq.\eqref{eq:mle_for_mc}})$ to eq.~\eqref{eq:mle_for_mc} such that $|\outset_{eq.\eqref{eq:mle_for_mc}}|\;<|\outset_\maxcon|$ and $\prod_{i\; \in \; \meas\setminus\outset_{eq.\eqref{eq:mle_for_mc}}} u(\res(\vy_i,\vxx_{eq.\eqref{eq:mle_for_mc}}),\inthr)\; > 0$. But the latter inequality implies $\res(\vy_i,\vxx_{eq.\eqref{eq:mle_for_mc}})\leq \inthr$ for all $i \in \meas\setminus\outset_{eq.\eqref{eq:mle_for_mc}}$ (since the uniform distribution has non-zero probability only in $[0,\inthr]$), and, as a result, $(\vxx_{eq.\eqref{eq:mle_for_mc}},\outset_{eq.\eqref{eq:mle_for_mc}})$ is feasible in \maxcon and, yet, $|\outset_{eq.\eqref{eq:mle_for_mc}}|\;<|\outset_\maxcon|$, which contradicts optimality. 

Now assume $(\vxx_{eq.\eqref{eq:mle_for_mc}},\outset_{eq.\eqref{eq:mle_for_mc}})$ is a solution to eq.~\eqref{eq:mle_for_mc} but not to \maxcon, \ie there exist a solution $(\vxx_\maxcon,\outset_\maxcon)$ to \maxcon such that $|\outset_\maxcon|\;<|\outset_{eq.\eqref{eq:mle_for_mc}}|$ and $\res(\vy_i,\vxx_\maxcon)\leq \inthr$ for all $i \in \meas\setminus\outset_\maxcon$. But the latter implies that $\prod_{i\; \in \; \meas\setminus\outset_\maxcon} u(\res(\vy_i,\vxx_\maxcon),\inthr)\;= \inthr^{-|\meas\setminus \outset_\maxcon|} > 0$, and, as a result, $(\vxx_\maxcon,\outset_\maxcon)$ is feasible for \eqref{eq:mle_for_mc} and, yet, $|\outset_\maxcon|\;<|\outset_{eq.\eqref{eq:mle_for_mc}}|$, which again contradicts optimality. 

%%%%%%%%%%%%%%%%%%%%%%%%%%%%%%%%%%%%%%%%%%%%%%%%%%%%%%%%%%%%%%%%%%%%%%%%%%%%%%%%%
\subsection{Proof of Proposition~\ref{th:mle_for_mts}}\label{app:proofs-formul-mle_for_mts}

The proof follows from taking the $\log$ of  ineq.~\eqref{eq:mle_for_mts}.
% is parallel to Proposition~\ref{th:mle_for_mc}'s and is omitted.

%%%%%%%%%%%%%%%%%%%%%%%%%%%%%%%%%%%%%%%%%%%%%%%%%%%%%%%%%%%%%%%%%%%%%%%%%%%%%%%%%
\subsection{Proof of Proposition~\ref{th:mle_for_tls}}\label{app:proofs-formul-mle_for_tls}

Assuming a known number of outliers $|\outset|\;=|\outsettrue|$, the \tls formulation in~\eqref{eq:main_2'} becomes
\beq \label{eq:tls-knownnumberoutliers}
\min_{\substack{ \vxx \;\in \;\calX \\ \outset \; \subseteq \; \measSet,\; |\outset|\;=|\outsettrue|}} \quad   \sumAllPointsIn{\MminusO}\resSqyx{i} \; + \; \inthr^2|\outsettrue|,
\eeq
% where we used \tls's form in eq.~\eqref{eq:main_2'}.  
where $ \inthr^2|\outsettrue|$ becomes a constant and is irrelevant for the optimization. % can now be ignored, since it is a constant. 
It can be now seen that taking the $-\log(\cdot)$ of the objective in eq.~\eqref{eq:mle_for_tls} leads to 
the same optimization as in~\eqref{eq:tls-knownnumberoutliers}.

%%%%%%%%%%%%%%%%%%%%%%%%%%%%%%%%%%%%%%%%%%%%%%%%%%%%%%%%%%%%%%%%%%%%%%%%%%%%%%%%%
\subsection{Proof of Proposition~\ref{th:mle_for_tls-normalAndUni}}\label{app:proofs-formul-mle_for_tls-normalAndUni}

Since $\vxxtrue$ is feasible in eq.~\eqref{eq:mle_for_tls-normalAndUni} and $\prod_{i\;\in\;\meas} \; \hat{g}(\res(\vy_{i}, \vxxtrue)) >0$ (since $\res(\vy_{i}, \vxxtrue) \leq \alpha$ for any $i \in \measSet$), for any optimal solution $\vxx$ to eq.~\eqref{eq:mle_for_tls-normalAndUni}, it also holds true that $\prod_{i\;\in\;\meas} \; \hat{g}(\res(\vy_{i}, \vxx)) >0$, and, as a result, $\res(\vy_{i}, \vxx) \leq \alpha$ for any $i \in \measSet$.  Therefore, after  
simplifying constants, eq.~\eqref{eq:mle_for_tls-normalAndUni} is equivalent to
\begin{equation}\label{eq:mle_for_tls-normalAndUni-1}
\max_{\vxx\; \in\; \calX}\quad\prod_{i\;\in\;\meas} \; \max\left\{e^{-{\res^2/2}} , e^{-{\inthr^2/2}}\right\},
\end{equation}
which is equivalent to
\begin{equation}\label{eq:mle_for_tls-normalAndUni-2}
\max_{\vxx\; \in\; \calX}\quad\sum_{i\;\in\;\meas} \; \max\left\{ \; -r^2/2 \; , -\inthr^2 / 2\; \right\},
\end{equation}
since maximizing the objective function in eq.~\eqref{eq:mle_for_tls-normalAndUni-1} is equivalent to maximizing the $\log$ of it.  Finally,~\eqref{eq:mle_for_tls-normalAndUni-2} is equivalent~to
$\max_{\vxx\; \in\; \calX}\;\sum_{i\;\in\;\meas} \min\left\{r^2, \inthr^2\right\}$,
which is equivalent to \tls.

%%%%%%%%%%%%%%%%%%%%%%%%%%%%%%%%%%%%%%%%%%%%%%%%%%%%%%%%%%%%%%%%%%%%%%%%%%%%%%%%%
\subsection{Proof of Theorem~\ref{th:relationship_linf}}\label{app:proofs-formul-relationship_linf}

{Denote by $(\vxx_\prtwo,\outset_\prtwo)$ any optimal solution to \prtwo.  
We first prove $(\vxx_\prtwo,\outset_\prtwo)$ is feasible to \prone (\ie $\|\; \vres(\vy_{\meas\setminus \outset_\prtwo},\vxx_\prtwo)\;\|_\infty\; \leq \epsilon$), and, then, prove $(\vxx_\prtwo,\outset_\prtwo)$  is actually an optimal solution to \prone.

To prove $\|\; \vres(\vy_{\meas\setminus \outset_\prtwo},\vxx_\prtwo)\;\|_\infty\; \leq \epsilon$, first observe 
\bea\label{eq:aux_proof_equiv}
\begin{split}
	&|\meas\setminus\outset_\prtwo|\;\cdot\; \|\; \vres(\vy_{\meas\setminus \outset_\prtwo},\vxx_\prtwo)\;\|_\infty^2\; +\\
&\hspace{4cm}\inthr^2|\outset_\prtwo|\; \leq \inthr^2 |\meas|,
\end{split}
\eea 
since  $\inthr^2$ is the value of \prtwo's objective function for $\outset=\meas$ (given any $\vxx \in \calX$), while $(\vxx_\prtwo,\outset_\prtwo)$ is an optimal solution to \prtwo.
Now, assume  $\|\vres(\vy_{\meas\setminus \outset_\prtwo},\vxx_\prtwo)\|_\infty\; > \epsilon$.  Then, the value of \prtwo's objective function at $(\vxx_\prtwo,\outset_\prtwo)$ is strictly more than $\inthr^2|\meas|$, which contradicts eq.~\eqref{eq:aux_proof_equiv}. Hence, $\|\; \vres(\vy_{\meas\setminus \outset_\prtwo},\vxx_\prtwo)\;\|_\infty\; \leq \epsilon$, and, as a result, $(\vxx_\prtwo,\outset_\prtwo)$ is feasible to \prone.

We now prove $\outset_\prtwo$ is also optimal for \prone. 
Assume by contradiction $\outset_\prtwo$ is not optimal for \prone. 
Then, $|\outset_\prone|\; < |\outset_\prtwo|$ (or, equivalently, 
$|\outset_\prone|+1 \leq |\outset_\prtwo|$), since $\outset_\prone$ is optimal.
Since also $\| \vresyx{\measSet\setminus \outset_\prone} \;\|_\infty \;< \inthr$, the following hold:
\bea
\| \vresyx{\measSet\setminus \outset_\prone} \;\|_\infty^2 + \lag^2 |\outset_\prone| < \label{eq:app_linf1} \\
\inthr^2 + \lag^2 |\outset_\prone| = \\ 
\lag^2 (|\outset_\prone|+1) \leq \lag^2 |\outset_\prtwo| \leq \\
|\meas\setminus\outset_\prtwo|\;\cdot\; \| \vresyx{\measSet\setminus \outset_\prtwo} \;\|_\infty^2 + \lag^2 |\outset_\prtwo|. \label{eq:app_linf2}
\eea
Comparing~\eqref{eq:app_linf1} and~\eqref{eq:app_linf2}, we notice that $\outset_\prone$ achieves a better cost in \prtwo,
contradicting the optimality of $\outset_\prtwo$. 
}
%%%%%%%%%%%%%%%%%%%%%%%%%%%%%%%%%%%%%%%%%%%%%%%%%%%%%%%%%%%%%%%%%%%%%%%%%%%%%%%%%
\subsection{Proof of Theorem~\ref{th:relationship}}\label{app:proofs-formul-relationship}

To prove the theorem,  consider the following problem:
\beq 
\optmshort{\min} \quad \|\; \vresyx{\measSet\setminus \outset} \;\|_2^2 \;\;\;\text{{s.t.}}\;\;\;  |\outset| \;= |\outset_\mts|. \label{eq:main_1'}
\eeq
Note that $\outset_\mts$ is feasible for \mts, hence the optimal objective of~\eqref{eq:main_1'} is smaller than $\outfreethr^2$.   

Now consider the   Lagrangian of~\eqref{eq:main_1'}: 
\begin{align}
l(\lagtls) &\triangleq 
\optmshort{\min} \quad \|\; \vresyx{\measSet\setminus \outset} \;\|_2^2 \;+\lagtls^2 \left(|\outset|-|\outset_\mts|\right)\nonumber\\
&= f_\tls(\lagtls)-\lagtls^2 |\outset_\mts|.
\end{align}
By weak duality~\cite{Boyd04book}:
\beq\label{app:eq:weak_duality}
f_\tls(\lagtls)-\lagtls^2 |\outset_\mts|\;\leq  \|\; \vres(\vy_{\measSet\setminus \outset_{\eqref{eq:main_1'}}}, \vxx_{\eqref{eq:main_1'}}) \;\|_{2}^2,
\eeq
where $(\vxx_{\eqref{eq:main_1'}}, \outset_{\eqref{eq:main_1'}})$ is an optimal solution to eq.~\eqref{eq:main_1'}.  Since $ \|\; \vres(\vy_{\measSet\setminus \outset_{\eqref{eq:main_1'}}}, \vxx_{\eqref{eq:main_1'}}) \;\|_2 \;\leq \tau$, then
\beq\label{eq:equivalency}
{f_\tls(\lagtls)}-\lagtls^2 |\outset_\mts| \;\leq {\outfreethr}^2,
\eeq
From the inequality~\eqref{eq:equivalency} it follows:

\begin{itemize}
	\item if $\outfreethr^2 = r^2_\tls(\lagtls)$, then eq.~\eqref{eq:equivalency} implies $|\outset_\tls|\; \leq |\outset_\mts|$; since $(\vxx_\tls, \outset_\tls)$ is also feasible for \eqone, $|\outset_\tls|\; = |\outset_\mts|$, and $(\vxx_\tls$, $\outset_\tls)$ is also a solution to $\mts$.
	\item if $\outfreethr^2 > r^2_\tls(\lagtls)$, then $|\outset_\tls|\;\geq |\outset_\mts|$,  since $(\vxx_\tls, \outset_\tls)$ is feasible in \eqone. 
	\item if $\outfreethr^2 < r^2_\tls(\lagtls)$, then $|\outset_\tls|\; < |\outset_\mts|$,   since $(\vxx_\tls, \outset_\tls)$ is infeasible in \eqone. 
\end{itemize}
 
\arxivVersion{%!TEX root = ../main.tex

\section{Alternative Justification for \tls}
\label{app:mle_for_tls-weibull}

\begin{proposition}[Weibull Distribution Leads to \tls]\label{th:mle_for_tls-weibull}
Assume $\res(\vy_{i}, \vxxtrue)\leq \inthr$ for any $i \in \measSet \setminus \outsettrue$.  If $\res(\vy_{i}, \vxxtrue)$ is a Weibull random variable for each $i \in \measSet$, with \emph{cumulative probability distribution} $\mathrm{Weib}(\res)\triangleq 1-\exp{(-\res^2/2)}$, then \tls is equivalent to 
the maximum likelihood estimator
\begin{equation}\label{eq:mle_for_tls-weibull}
{\optmshort{\max}}\;\;\prod_{i\;\in\;\MminusO} \; \left[1-\mathrm{Weib}(\resyx{i})\right] \; \prod_{i\; \in \; \outset}\;  \left[1-\mathrm{Weib}(\inthr)\right].
\end{equation}
\end{proposition}
Broadly speaking, the Weibull distribution is commonly used in statistics to model the probability of an outcome's \emph{failure} when the failure depends on sub-constituent failures: \eg a chain breaks if any of its rings breaks~\cite{Weibull1951applmech-weibull}.  
Similarly, an outlier-robust estimate ``breaks'' if measurements are misclassified as inliers instead of outliers and vice versa, and if the inliers' residuals are unnecessarily large:
\begin{itemize}
	\item if a measurement $i$ is classified as an outlier ($i \in \outset$), then $1-\mathrm{Weib}(\inthr)$ models the probability of a \emph{successful} estimation given that $i$'s residual is \emph{at least} $\inthr$;
	\item if a measurement $i$ is classified as an inlier ($i \in \MminusO$), then $1-\mathrm{Weib}(\resyx{i})$ models the probability of a \emph{successful} estimation given that $i$'s residual is at least $\resyx{i}$ \emph{but not more than} $\inthr$: indeed, if $\resyx{i}>\inthr$, then eq.~\eqref{eq:mle_for_tls-weibull} classifies $i$ as an outlier, so to maximize the joint probability likelihood, since $1-\mathrm{Weib}(\resyx{i})<1-\mathrm{Weib}(\inthr)$. Therefore, for all $i\in \MminusO$, $\resyx{i} \leq \inthr$.
\end{itemize} 
In summary, eq.~\eqref{eq:mle_for_tls-weibull} aims to find $(\vxx, \outset)$ that maximize the probability of the estimator's success, and, particularly, it does so by forcing the inliers' $\resyx{i}$ to be as small as possible, since indeed $1-\mathrm{Weib}(\resyx{i})\rightarrow 1$ when $\resyx{i}\rightarrow 0$.

%%%%%%%%%%%%%%%%%%%%%%%%%%%%%%%%%%%%%%%%%%%%%%%%%%%%%%%%%%%%%%%%%%%%%%%%%%%%%%%%%
\subsection{Proof of Proposition~\ref{th:mle_for_tls-weibull}}\label{app:proofs-formul-mle_for_tls-weibull}

The proof is derived by taking the $-\log(\cdot)$ of the objective function in eq.~\eqref{eq:mle_for_tls-weibull}, resulting in the \tls cost in eq.~\eqref{eq:main_2'}.}{}
%!TEX root = ../main.tex

\section{Proof of Theorem~\ref{th:hardness}} 
\label{app:hardness}

We prove the theorem based on the inapproximability of the {\em variable selection} problem, reviewed in \ref{app:inapprox-prelim}.  In particular, we first prove the inapproximability of \prone, by proving the inapproximability of  \mts 
and \maxcon (\ref{app:inapprox-mts} and \ref{app:inapprox-maxcon}, respectively).  Then, we prove the inapproximability of \prtwo, by proving the inapproximability of \tls (\ref{app:inapprox-tls}). For all cases we consider a linear measurement model, which results in residuals of the form:
$$\resyx{i}=|y_i-\va_i\tran \vxx|,$$
for all $i \in \meas$, where $y_i$ is scalar, and $\va_i$ is a column vector.

\subsection{Preliminary Definitions and Results}\label{app:inapprox-prelim}

We present the \emph{variable selection} problem, recall a known result on its inapproximability in even quasi-polynomial time, and 
review results that we will subsequently use for the proof of Theorem~\ref{th:hardness}.
We use the standard notation $\|\vxx\|_0$ to denote the number of non-zero elements in $\vxx$.

\begin{problem}[Variable selection]\label{pr:selection}
Assume a matrix $\MU \in \mathbb{R}^{\urows \times \measn}$, a vector $\vz \in \mathbb{R}^\urows$, and a non-negative scalar $\xi$. Find a vector $\vd \in \mathbb{R}^\measn$ that solves the optimization problem
\begin{equation}\label{eq:varsel}
	\min_{\vd \;\in \;\mathbb{R}^\measn}\quad \|\;\vd\;\|_{0},\;\;\;\text{\emph{s.t.}}\;\;\; \|\;\MU\vd-\vz\;\|_{2}\;\leq b.
\end{equation}
\end{problem}

The following lemma describes inapproximable instances of \emph{variable selection} even in quasi-polynomial time.

\begin{lemma}[Inapproximability of Variable Selection in Quasi-polynomial Time {\cite[Proposition 6]{Foster15colt-variableSelectionHard}}]\label{th:inapprox_selection} 
For any $\delta \in (0,1)$, unless $\np \notin\bptime(\measn^{\poly\log \measn})$, there exist
	\begin{itemize} \setlength\itemsep{0.09em}
		\item a function $q_1(\measn)=2^{\Omega(\log^{1-\delta} \measn)}$, 
		\item {a polynomial $p_1(\measn)=O(\measn)$,}
		\item {a polynomial $\xi(\measn)$,}
		\item {a polynomial $\urows(\measn)$,} 
		\item {and a zero-one} matrix $\MU \in \mathbb{R}^{\urows(\measn) \times \measn}$,
	\end{itemize} 		
such that, for large enough $\measn$, no quasi-polynomial algorithm {finds a $\vd \in {\mathbb{R}^\measn}$ distinguishing the mutually-exclusive cases:}\footnote{If $\measn$ is large enough,  then $q_1(\measn) > 1$ (since $q_1(\measn) = 2^{ \Omega ( \log^{1-\delta} \measn)}$, where $\delta\in (0,1)$), and, as a result, S$_1$ and S$_2$ are mutually exclusive.}
	\begin{enumerate} \setlength\itemsep{0.09em}
		\item[\emph{S}$_1$.] {There exists a vector $\vd \in {\mathbb{R}^\measn}$ such that $\MU\vd = \mathbf{1}_{\urows(\measn)}$ and $||\;\vd\;||_{0} \;\leq p_1(\measn)$.}
		\item[\emph{S}$_2$.] {For any $\vd \in \mathbb{R}^\measn$, if $||\;\MU\vd - \mathbf{1}_{\urows(\measn)}\;\|_{2}^2\; \leq \xi(\measn)$, then $||\;\vd\;\|_{0}\; \geq p_1(\measn) q_1(\measn)$.}
	\end{enumerate}
The observation holds true even if the algorithm knows that $\MU\vd=\bm{1}_{\urows(\measn)}$ is feasible for some $\vy \in \mathbb{R}^\measn$, where $\vy$ itself is unknown to the algorithm but $\|\;\vy\;\|_0$ is known.
\end{lemma}  

In the next section, we use the inapproximability of variable selection to prove that \mts is inapproximable.
Towards this goal, we prove two intermediate results. 

We start with the following optimization problem and prove that it is also inapproximable:
\begin{equation}\label{pr:aux_1}
\min_{\vd\; \in\; \mathbb{R}^\measn}\quad \|\;\vd\;\|_{0},\;\;\;\text{s.t.}\;\;\; \MU\vd=\mathbf{1}_{\urows(\measn)}.
\end{equation} 

\textbf{Proof that eq.~\eqref{pr:aux_1} is inapproximable:}
It suffices to set $b=0$ in eq.~\eqref{eq:varsel}, and then apply Lemma~\ref{th:inapprox_selection}. \hfill $\square$

\medskip

Given eq.~\eqref{pr:aux_1}'s inapproximability, we now prove the inapproximability of the optimization problem
\begin{equation}\label{pr:aux_2}
\min_{\displaystyle\substack{\vd \;\in\; \mathbb{R}^\measn\\ \!\vxx\;\in\; \mathbb{R}^n}}\quad \|\;\vd\;\|_{0},\;\text{s.t.}\;\; \vy=\MA\vxx+\vd,
\end{equation} 
for an appropriate class of matrices $\MA$.

\textbf{Proof that eq.~\eqref{pr:aux_2} is inapproximable:} Given the inapproximable instances of eq.~\eqref{pr:aux_1}  (see Lemma~\ref{th:inapprox_selection}), consider the instances for eq.~\eqref{pr:aux_2} where (i) $\vy$ is any solution to $\MU\vy=\bm{1}_{\urows(\measn)}$ (because of Lemma~\ref{th:inapprox_selection}, such a $\vy$ exists), and (ii) $\MA$ is a matrix in $\mathbb{R}^{\measn\times n}$, where $n=\measn-\text{rank}(\MU)$,
such that the columns of $\MA$ span the null space of $\MU$ ($\MU\MA=\zero$).  Any such instance is constructed in polynomial time in $\measn$, since solving a system of equations and finding eigenvectors that span a matrix's null space happen in polynomial time.  

We now prove the following statements are indistinguishable, where we consider $\xi'(\measn)\triangleq \urows^{-2.5}(\measn)\xi(\measn)$:
\begin{enumerate}\setlength\itemsep{0.09em}
	\item[S$_1'$.] \emph{There exist $\vd \in {\mathbb{R}^\measn}$ and $\vxx \in {\mathbb{R}^n}$ such that $\vy=\MA\vxx+\vd$ and $||\;\vd\;||_{0} \;\leq p_1(\measn)$.}
	\item[S$_2'$.] \emph{For any $\vd \in {\mathbb{R}^\measn}$ and $\vxx \in {\mathbb{R}^n}$, if $||\vy-\MA\vxx-\vd||^2_{2} \;\leq \xi'(\measn)$}, then $||\;\vd\;\|_{0} \;\geq p_1(\measn) q_1(\measn)$.
\end{enumerate}
To this end, we prove that (i) if S$_1$ is true (which is for any feasible $\vd$ in eq.~\eqref{pr:aux_1}), then S$_1'$ also is, and (ii) if S$_2$ is true, then also S$_2'$ is.  
Therefore, no quasi-polynomial time algorithm can distinguish S$_1'$ and S$_2'$, since the opposite would contradict that S$_1$ and S$_2$ are indistinguishable.  In particular:

\paragraph{a) Proof that when S$_1$ is true then S$_1'$ also is}  Since $\MU\vy=\MU\MA\vxx+\MU\vd$ implies that $\bm{1}_{\urows(\measn)}=\MU\vd$, if S$_1$ is true, then S$_1'$ also is; moreover, $\vxx$ is the unique solution to $\MA\vxx=\vy-\vd$ ($\vxx$ is unique since $\MA$ is full column rank).
	
\paragraph{b) Proof that when S$_2$ is true then S$_2'$ also is}  Assume $\vd \in {\mathbb{R}^\measn}$ and $\vxx \in {\mathbb{R}^n}$ such that $||\;\vy-\MA\vxx-\vd\;||^2_{2} \; \leq \xi'(\measn)$ and $||\;\vd\;\|_{0} \;< p_1(\measn) q_1(\measn)$. 
If $||\vy-\MA\vxx-\vd||^2_{2} \;\leq \xi'(\measn)$, then $||\;\vy-\MA\vxx-\vd\;||^2_{1} \;\leq [\urows(\measn)]^{0.5}\;\xi'(\measn)$, due to norms' equivalence. Hence, $||\;\MU\;||^2_{1}\; ||\vy-\MA\vxx-\vd||^2 _{1} \;\leq ||\;\MU\;||^2_{1} \;[\urows(\measn)]^{0.5}\;\xi'(\measn)$, which implies $||\;\MU(\vy-\MA\vxx-\vd)\;||^2_{1} \; \leq ||\;\MU\;||^2_{1}\; [\urows(\measn)]^{0.5}\;\xi'(\measn)$, \ie $||\;\bm{1}_{\urows(\measn)}-\MU\vd\;||^2_{1}  \;\leq ||\;\MU\;||^2_{1}\; [\urows(\measn)]^{0.5}\;\xi'(\measn)$, and as a result $||\;\bm{1}_{\urows(\measn)}-\MU\vd\;||^2_{1}  \;\leq \urows(\measn)^{2.5}\;\xi'(\measn)$, where the last holds true because $\MU$ is a zero-one matrix. Consequently, $||\bm{1}_{\urows(\measn)}-\MU\vd||^2_{2}  \;\leq [\urows(\measn)]^{2.5}\;\xi'(\measn)$, due to norms' equivalence. Finally, due to $\xi'(\measn)$'s definition, $[\urows(\measn)]^{2.5}\; \xi'(\measn)=\xi(\measn)$; thus,  $||\;\bm{1}_{\urows(\measn)}-\MU\vd\;||^2_{2} \; \leq \xi(\measn)$.  Overall, there exist $\vd$ such that $||\;\bm{1}_{\urows(\measn)}-\MU\vd\;||^2_{2} \;\leq \xi(\measn)$ and $||\;\vd\;\|_{0} \;< p_1(\measn) q_1(\measn)$, which contradicts S$_2$.
\hfill $\square$

%%%%%%%%%%%%%%%%%%%%%%%%%%%%%%%%%%%%%%%%%%%%%%%%%%%%%%%%%%%%%%%%%%%%%%%%%%%%%%%%%%%%%%%%%%%%%%%%%%%%%%%%%%%%
\subsection{Proof that \mts is Inapproximable}\label{app:inapprox-mts}

We use the notation:
\begin{itemize}
	\item $\vy_{\grSet\setminus \selSet}\triangleq \{y_i\}_{i \; \in \; \MminusO}$, \ie $\vy_{\grSet\setminus \selSet}$ is the stack of all measurements $i\in \grSet\setminus\selSet$;
	\item  $\vd_{\grSet\setminus \selSet}\triangleq \{d_i\}_{i \; \in \; \MminusO}$,  \ie $\vd_{\grSet\setminus \selSet}$ is the stack of all noises $i\in \grSet\setminus\selSet$;
	\item $\MA_{\grSet\setminus\selSet}\triangleq \{\va_i\tran\}_{i \; \in \; \MminusO}$, \ie $\MA_{\grSet\setminus\selSet}$ is the matrix with rows the row-vectors $\va_i\tran$, $i\in \grSet\setminus\selSet$.
\end{itemize}
The \mts problem in eq.~\eqref{eq:mts} now takes the form
\begin{equation}\label{pr:aux_3}
\min_{\displaystyle\substack{\selSet\;\subseteq\;\grSet\\ \;\vxx\;\in\; \mathbb{R}^n}}\quad  |\selSet|,\;\;\;\text{s.t.}\;\;\; \|\;\vy_{{\grSet}\setminus {\selSet}}-\MA_{{\grSet}\setminus {\selSet}}\vxx\;\|^2_{2}\;\leq \outfreethr^2.
\end{equation}

To prove eq.~\eqref{pr:aux_3}'s inapproximability, we first consider an {inapproximable instance of eq.~\eqref{pr:aux_2}, and  in eq.~\eqref{pr:aux_3} let $\meas = \{1,2,\ldots,\measn\}$ and $\outfreethr^2=\xi'(\measn)$.}
Then, we prove the following statements are indistinguishable:
\begin{enumerate}\setlength\itemsep{0.09em}
	\item[S$_1''$.] \emph{There exist  ${\selSet} \subseteq\meas$ and $\vxx \in {\mathbb{R}^n}$ such that $\vy_{{\grSet}\setminus {\selSet}}=\MA_{{\grSet}\setminus {\selSet}}\vxx$ and $|{\selSet}| \;\leq p_1(\measn)$.}
	\item[S$_2''$.] \emph{For any ${\selSet} \subseteq\meas$ and $\vxx \in {\mathbb{R}^n}$, if $||\;\vy_{{\grSet}\setminus {\selSet}}-\MA_{{\grSet}\setminus {\selSet}}\vxx\;||^2_{2} \;\leq \xi'(\measn)$, then $|{\selSet}| \;\geq p_1(\measn) q_1(\measn)$.}
\end{enumerate}
To this end, we prove that (i) if S$_1'$ is true, then S$_1''$ also is, and (ii) if S$_2'$ is true, then also S$_2''$ is.  In more detail:
\setcounter{paragraph}{0}
\paragraph{a) Proof that if S$_1'$ is true then S$_1''$ also is} 
Assume S$_1'$ is true and let ${\selSet}=\{i \;\text{ s.t. } \; d_i\neq 0,\;\; i\in \meas\}$. Then, $\vy_{{\grSet}\setminus {\selSet}}=\MA_{{\grSet}\setminus {\selSet}}\vxx$, since $\vd_{{\grSet}\setminus {\selSet}}=0$ and {$|{\selSet}|\;=||\;\vd\;\|_{0}\;\leq p_1(\measn)$.}

\paragraph{b) Proof that if S$_2'$ is true then S$_2''$ also is} Assume ${\selSet} \subseteq \meas$ and $\vxx \in {\mathbb{R}^n}$ such that $||\;\vy_{{\grSet}\setminus {\selSet}}-\MA_{{\grSet}\setminus {\selSet}}\vxx\;||^2_{2} \;\leq \xi'(\measn)$ and $|{\selSet}|\; < p_1(\measn) q_1(\measn)$. Let $\vd_{{\grSet}\setminus {\selSet}}=0$, and $\vd_{\selSet}=\vy_{\selSet}-\MA_{\selSet}\vxx$.  Then, $||\;\vd\;\|_{0}=|\selSet|\;< p_1(\measn) q_1(\measn)$ and $||\;\vy-\MA\vxx-\vd\;||^2_{2}=||\;\vy_{{\grSet}\setminus {\selSet}}-\MA_{{\grSet}\setminus {\selSet}}\vxx\;||^2_{2}\;\leq \xi'(\measn)$, which contradicts~S$_2'$.
%\hfill $\square$

%%%%%%%%%%%%%%%%%%%%%%%%%%%%%%%%%%%%%%%%%%%%%%%%%%%%%%%%%%%%%%%%%%%%%%%%%%%%%%%%%%%%%%%%%%%%%%%%%%%%%%%%%%%%
\subsection{Proof that \maxcon is Inapproximable}\label{app:inapprox-maxcon}

\arxivVersion{The proof proceeds along the same line of \mts's proof. We use the same notation used in \ref{app:inapprox-mts}.  

We first consider an inapproximable instance of eq.~\eqref{pr:aux_2}, and in eq.~\eqref{eq:maxConsensus} set $\inthr^2=\xi'(\measn)$.
We then prove that the following statements are indistinguishable:
\begin{enumerate}\setlength\itemsep{0.09em}
	\item[S$_1'''$.] \emph{There exist  ${\selSet} \subseteq \meas$ and $\vxx \in {\mathbb{R}^n}$ such that $\vy_{{\grSet}\setminus {\selSet}}=A_{{\grSet}\setminus {\selSet}}\vxx$ and $|{\selSet}| \;\leq p_1(\measn)$.}
	\item[S$_2'''$.] \emph{For any ${\selSet} \subseteq \meas$ and $\vxx \in {\mathbb{R}^n}$, if $||\vy_{{\grSet}\setminus {\selSet}}-\MA_{{\grSet}\setminus {\selSet}}\vxx\|^2_{\infty} \;\leq \xi'(\measn)$, then $|{\selSet}| \;\geq p_1(\measn) q_1(\measn)$.}
\end{enumerate}
To this end, we prove that (i) if S$_1''$ is true, then S$_1'''$ also is, and (ii) if S$_2''$ is true, then also S$_2'''$ is. Specifically:
\setcounter{paragraph}{0}
\paragraph{a) Proof that if S$_1'$ is true then S$_1'''$ also is} Assume S$_1'$ is true and let ${\selSet}=\{i \;\text{ s.t. } \;d_i\neq 0,\;\; i\in \meas\}$. Then, $\vy_{{\grSet}\setminus {\selSet}}=\MA_{{\grSet}\setminus {\selSet}}\vxx$, since $\vd_{{\grSet}\setminus {\selSet}}=0$ and $|{\selSet}|\;=||\;\vd\;\|_{0}\;\leq p_1(\measn)$.

\paragraph{b) Proof that if S$_2'$ is true then S$_2'''$ also is} Consider ${\selSet} \subseteq \meas$ and $\vxx \in {\mathbb{R}^n}$ such that $||\;\vy_{{\grSet}\setminus {\selSet}}-\MA_{{\grSet}\setminus {\selSet}}\vxx\;||^2_{1} \;\leq \xi'(\measn)$ and $|{\selSet}|\; < p_1(\measn) q_1(\measn)$. Let $d_{{\grSet}\setminus {\selSet}}=0$, and $\vd_{\selSet}=\vy_{\selSet}-\MA_{\selSet}\vxx$.  Then, $||\;\vd\;\|_{0}\;=|\selSet|\;< p_1(\measn) q_1(\measn)$ and $||\;\vy-\MA\vxx-\vd\;||^2_{1}\;\leq ||\;\vy-\MA\vxx-\vd\;||^2_{2}\;=||\;\vy_{{\grSet}\setminus {\selSet}}-\MA_{{\grSet}\setminus {\selSet}}\vxx\;||^2_{2}\;\leq \xi'(\measn)$, where  the first inequality holds due to the norms' equivalence, while the latter inequality contradicts~S$_2'$.}{Due to space limitations, the proof is presented in~\cite{\arxivThisPaper}.}

\subsection{Proof that \tls problem is Inapproximable}\label{app:inapprox-tls}

\arxivVersion{We prove the inapproximability of eq.~\eqref{eq:tls} by using the inapproximability of eq.~\eqref{pr:aux_3}. To this end, we use the notation in \ref{app:inapprox-mts}, along with the notation $$f(\vxx,\vw)\triangleq \sumAllPointsIn{\meas}\; \min_{w_i \;\in\; \{0,1\}} \;\left[w_i\;(y_i-\va_i\tran \vxx)^2  +  (1-w_i)\;\inthr^2\right].$$

Consider an inapproximable instance of \eqref{pr:aux_3}, and in~\eqref{eq:tls} set $\inthr^2=\nicefrac{1}{p_1(\measn)}$.  
We prove the following are indistinguishable:
\begin{itemize}
	\item[$\bar{\text{S}}_1$.] \emph{There exist  $\vw \in \{0,1\}^\measn$ and $\vxx \in {\mathbb{R}^n}$ such that $f(\vxx,\vw)\leq 1$ and $||\;\vw\;||_{0}\; \leq p_1(\measn)$.}
	\item[$\bar{\text{S}}_2$.] \emph{For any $\vw \in \{0,1\}^\measn$ and $\vxx \in {\mathbb{R}^n}$, if $f(\vxx,\vw) \leq \xi'(\measn)$, then $||\;\vw\;||_{0} \;\geq p_1(\measn) q_1(\measn)$.}
\end{itemize}
To this end, we prove that (i) if S$_1''$ is true, then $\bar{\text{S}}_1$ also is, and (ii) if S$_2''$ is true, then also $\bar{\text{S}}_2$ is. Specifically:
	\setcounter{paragraph}{0}
	\paragraph{a) Proof that if S$_1''$ is true then $\bar{\text{S}}_1$ also is} Assume S$_1''$ is true and let $w_i=1$ for all $i\in \selSet$, and $0$ otherwise.  Then, $||\vw||_{0}\;=|\selSet|\;\leq p_1(\measn)$, and $f(\vxx,\vw)=|\selSet| \inthr^2\leq p_1(\measn)\inthr^2=1$.
	
	\paragraph{b) Proof that if S$_2''$ is true then $\bar{\text{S}}_2$ also is} Assume $\vw \in \{0,1\}^\measn$ and $\vxx \in {\mathbb{R}^n}$ such that $f(\vxx,\vw) \leq \xi'(\measn)$ and $||\;\vw\;||_{0} \;< p_1(\measn) q_1(\measn)$. Let ${\selSet}=\{i \;\text{ s.t. }\; w_i = 1\}$, and as a result, $|{\selSet}|\; < p_1(\measn) q_1(\measn)$.  Since $f(\vxx,\vw) \leq \xi'(\measn)$ and $f(\vxx,\vw) =\|\vy_{{\grSet}\setminus \selSet}-\MA_{{\grSet}\setminus \selSet}\vxx\|^2_{2}$, it holds true that $\|\vy_{{\grSet}\setminus \selSet}-\MA_{{\grSet}\setminus \selSet}\vxx\|^2_{2}\; \leq \xi'(\measn)$, which contradicts~S$_2''$.}{Due to space limitations, the proof is presented in~\cite{\arxivThisPaper}.}

%!TEX root = ../main.tex

\section{Proof of Theorem~\ref{th:weightsTLSlimit}} 
\label{app:proofs-weightsTLSlimit}

\arxivVersion{The proof follows by taking $\iteration\rightarrow +\infty$ (or, equivalently $\mu\at{\iteration} \rightarrow +\infty$) in eq.~\eqref{eq:weightUpdate-TLS}.  In more detail, it suffices to observe that $\limtinf \;\mu\at{\iteration-1}\;/\;(\mu\at{\iteration-1} + 1) = 1$, $\limtinf \;(\mu\at{\iteration-1}+1)/\mu\at{\iteration-1}= 1$, and 
$\limtinf\; (\barc\sqrt{\mu\at{\iteration-1}(\mu\at{\iteration-1}+1)}/\res_i\at{t}- \mu\at{\iteration-1})=1/2$.  
In particular, the latter is true since $\limtinf\; (\barc\sqrt{\mu\at{\iteration-1}(\mu\at{\iteration-1}+1)}/\res_i\at{t}- \mu\at{\iteration-1}) = \limtinf [\barc\sqrt{\mu\at{\iteration-1}+1}/(\sqrt{\mu\at{\iteration-1}}\res_i\at{t})- 1]/(1/\mu\at{\iteration-1})$, where now L'H\^{o}spital's rule implies the latter is equal to 
\begin{align*}
&\limtinf
\frac{\frac{d\mbox{ }}{d\mu\at{\iteration-1}}\left(\frac{\barc\sqrt{\mu\at{\iteration-1}+1}}{\res_i\at{t}\sqrt{\mu\at{\iteration-1}}}- 1\right)}
{\frac{d\mbox{ }}{d\mu\at{\iteration-1}}\left(\frac{1}{\mu\at{\iteration-1}}\right)}=\\
&\limtinf \frac{\barc}{\res_i\at{t}}
\frac{\frac{\sqrt{\mu\at{\iteration-1}}}{2\sqrt{\mu\at{\iteration-1}+1}}-\frac{\sqrt{\mu\at{\iteration-1}+1}}{2\sqrt{\mu\at{\iteration-1}}}}
{\mu\at{\iteration-1}\frac{-1}{(\mu\at{\iteration-1})^2}}=\\
&\limtinf \frac{\barc}{\res_i\at{t}}
\frac{\frac{-1}{2\sqrt{\mu\at{\iteration-1}}\sqrt{\mu\at{\iteration-1}+1}}}
{\mu\at{\iteration-1}\frac{-1}{(\mu\at{\iteration-1})^2}}=\frac{1}{2},
\end{align*}
where to derive the last equation we also took into account that $\limtinf {\barc}/{\res_i\at{t}}=1$ (since the domain of $\barc\sqrt{\mu\at{\iteration-1}(\mu\at{\iteration-1}+1)}/\res_i\at{t}- \mu\at{\iteration-1}$, with respect to $\res_i\at{t}$, becomes the set $\{\inthr\}$ for $\iteration \rightarrow +\infty$).}{Due to space limitations, the proof is presented in~\cite{\arxivThisPaper}.}
%!TEX root = ../main.tex

\section{$\separation$ algorithm}\label{app:alg-separation}

\adaptfree's subroutine $\separation$ is presented in Algorithm~\ref{alg:separation}.  Therein, for any real-vector $\vz \in \Real{l}$ such that $z_i \geq 0$, and for all $i = 1,2,\ldots, l$,
$\kdist(\vz) \triangleq \sum_{i=1}^{l}\;|z_i - \mean(\vz)|^2$,
and $\mean(\vz) = \frac{1}{l}\sum_{i=1}^{l}  z_i$; \ie $\kdist$ captures the cumulative deviation of all $z_i$ from their mean ---their ``centroid''--- and, as such, can be interpreted as a diameter.

\vspace*{-2mm}
%%%%%%%%%%%%
%!TEX root = ../main.tex

\begin{algorithm}[h]\label{alg:separation}
	\caption{\mbox{ClustersSeparation (\adaptfree's subroutine).}}
	\SetAlgoLined
	\KwIn{A real-valued vector $\vres \in \Real{l}$.}
	\KwOut{Centroids' distance that separates two clusters of entries in $\vres$.}
	\BlankLine
	$\bm{z} = \sort(r_1, r_2, \ldots, r_l)$; \hspace{2mm}\tcp{increasing order}   
	$i = \argmin_{j\in\{1,2,\ldots,l-1\}} \;\kdist(\bm{z}_{1:j}) + \kdist(\bm{z}_{j+1:\last})$\;
	$c_{\text{left}}=\mean(\bm{z}_{1:i})$;\quad$c_{\text{right}}=\mean(\bm{z}_{i+1:\last})$\;
  % $\delta=c_{\text{right}} - c_{\text{left}}$\;
  % $\delta=\mean(\bm{z}_{1:i}) - \mean(\bm{z}_{i+1:\last})$\;
  % \Return{$\delta$.}
  \Return{$c_{\text{right}} - c_{\text{left}}$.}
\end{algorithm}
\vspace*{-4mm}
%%%%%%%%%%%%

%!TEX root = ../main.tex

\section{$\fitChi$ algorithm}\label{app:alg-fitcdf}

$\fitChi$ is presented in Algorithm~\ref{alg:fitcdf}.
$\fitChi$ scores the fit of the empirical distribution of the residuals to the {$\mathrm{Gamma}(d/2,2{\sigma}^2)$} distribution, which is equivalent to the desired $\chi^2$ with degree of freedom $d$ and variance $\sigma^2$.
Since $\inthr$ is unknown, the true variance of the residuals' error is also unknown. 
For this reason, in $\fitChi$'s line~\ref{line:fitCDF-variance} an unbiased estimator for the variance is employed~\cite{Bolch68amstat-unbiasedvarEstimator}. 
% Then, line~\ref{line:fitCDF-2} stores the empirical cumulative distribution function (CDF) of the squared residuals.
Then, line~\ref{line:fitCDF-CVM} uses the Cram\'{e}r–von Mises test to score the fit.
% of the empirical cumulative distribution function of the residuals to the cumulative distribution function of the $\chi^2$ distribution.

\vspace{-2mm}
%%%%%%%%%%
%!TEX root = ../main.tex

\begin{algorithm}[h]\label{alg:fitcdf}
	\caption{\mbox{$\fitChi$ (\gncfree's subroutine).}}
	\SetAlgoLined
	\KwIn{\mbox{Real-valued vector $\vres \in \Real{n}$;} \mbox{$\chi^2$ distribution's degrees of freedom $d>0$.}}
	\KwOut{\mbox{Similarity statistic of $\chi^2$ distribution with empi-} rical distribution of $\vres$'s squared elements.}
	\BlankLine
	%$n=\operatorname{length}(\vres$)\;
	${\sigma}^2 = \frac{1}{(n-1)d}\sum_{i=1}^n r_i^2$\label{line:fitCDF-variance}\;
	% $F_n = \mathrm{EmpiricalCDF}(r_1^2, r_2^2,\ldots, r_n^2)$ \label{line:fitCDF-2}\;
	% $s = \mathrm{KolmogorofSmirnof}(F_n, {\mathrm{Gamma}(\frac{{d}}{2},2{\sigma}^2)})$\label{line:fitCDF-3}\tcp*[I]{	{$\mathrm{Gamma}(\frac{d}{2},2{\sigma}^2)$} = $\chi^2$ with degree of freedom $d$ and variance $\sigma^2$}
	$s = \mathrm{CramerVonMises}(\vres, {\mathrm{Gamma}(\frac{{d}}{2},2{\sigma}^2)})$\label{line:fitCDF-CVM}\tcp*[I]{	{$\mathrm{Gamma}(\frac{d}{2},2{\sigma}^2)$} = $\chi^2$ with degree of freedom $d$ and variance $\sigma^2$}
	\Return{$s$.} %\tcp{Smaller $s$ means better fit}
\end{algorithm}
%%%%%%%%%%
\vspace{-6mm}

\arxivVersion{%!TEX root = ../main.tex

\section{Additional Experimental Results: True and False Positive Rates~\label{app:true-false-positive}}

In Figs.~\ref{fig:stats_mesh}-\ref{fig:slam_stats_garage}, we report the True Positive Rate (number of correctly identified outliers over the number of ground truth outliers) and False Positive Rate (number of incorrectly identified outliers over the number of ground truth inliers) corresponding to the numerical results in Section~\ref{sec:experiments}.

Figs.~\ref{fig:stats_mesh}-\ref{fig:slam_stats_garage} agree with the observed accuracy performance of the proposed algorithms: all proposed algorithms reject most of the outliers, achieving high True Positive rate; and they reject a few inliers, achieving, typically, a 10\%-20\% False Positive rate for increasing outliers.  
In particular, \gnc and \gncfree exhibit superior performance among the proposed algorithms, rejecting the most number of true outliers (achieving high True Positive rate) and the least number of true inliers (achieving low False Positive rate).  Exceptions are Fig.~\ref{fig:stats_shape}, where \gnc is shown to reject on average 15\%-35\% of inliers (still, the accuracy performance of the algorithm is uncompromised for the displayed spectrum of outlier rates, per Fig.~\ref{fig:shape}); and Fig.~\ref{fig:slam_stats_garage}, where both the True Positive and False Positive performance of \gnc deteriorates for increasing outliers (a trend that agrees with the deteriorating accuracy performance of the algorithm observed in Fig.~\ref{fig:slam_garage}).

%!TEX root = ../../main.tex

\begin{figure*}[t!]
	\begin{center}
	\begin{minipage}{\textwidth}
	\begin{tabular}{cc}%
		%%%%%%%%%%%%%%%%%%%%%%%%%%%%%%%%%%%%%%%%%%%%%%%%%%%%%%%%%%%%%%%%%%%%%%%%%%%%%%%%%%%%%%%%%%%%%%%%%%%%%%%%%
		\mpPreSpace
			\begin{minipage}{\mpColTwo}%
        \centering%
        \includegraphics[width=\columnwidth]{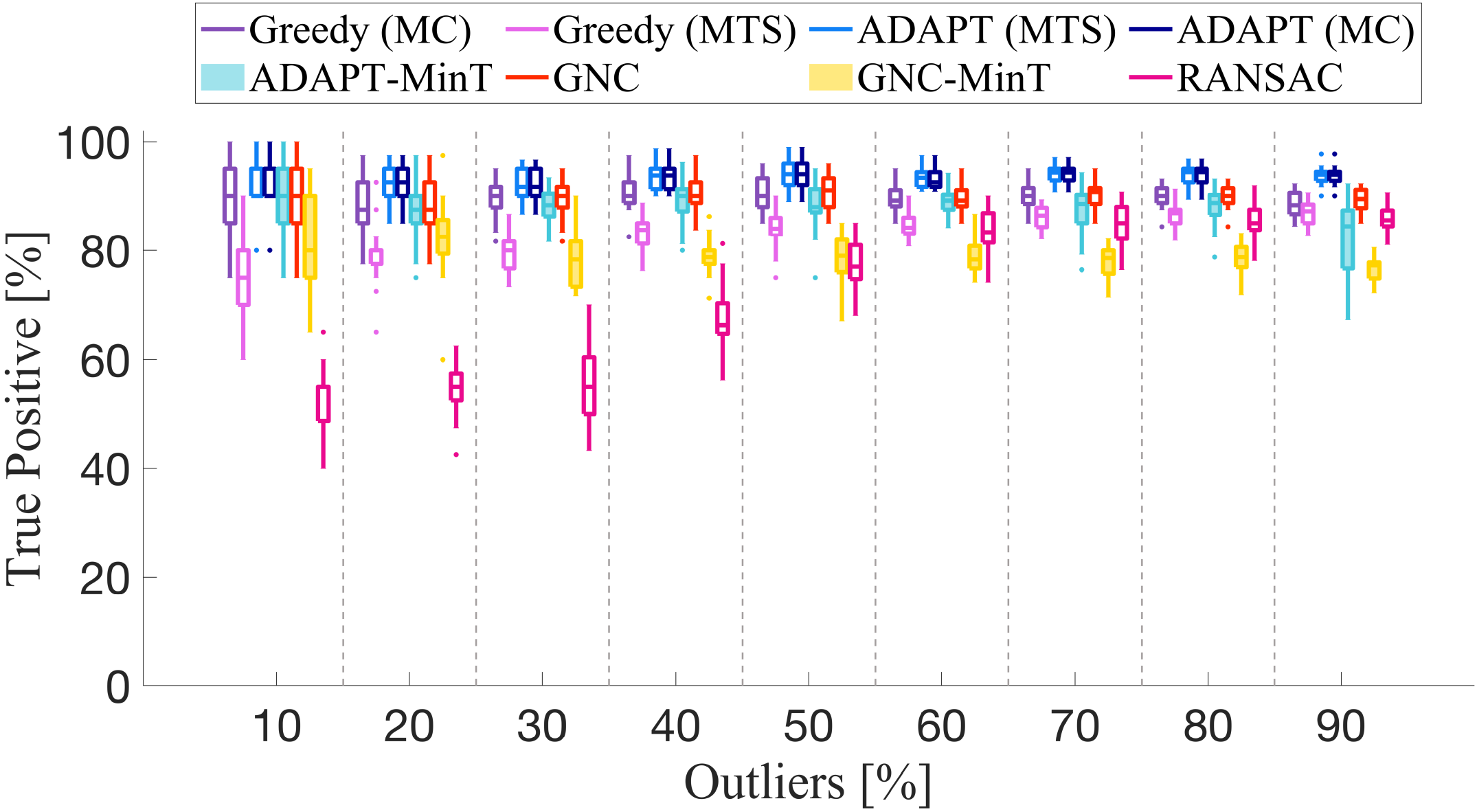} \\
			\end{minipage}
		& \mpMidSpaceTwo
			\begin{minipage}{\mpColTwo}%
        \centering%
        \includegraphics[width=\columnwidth]{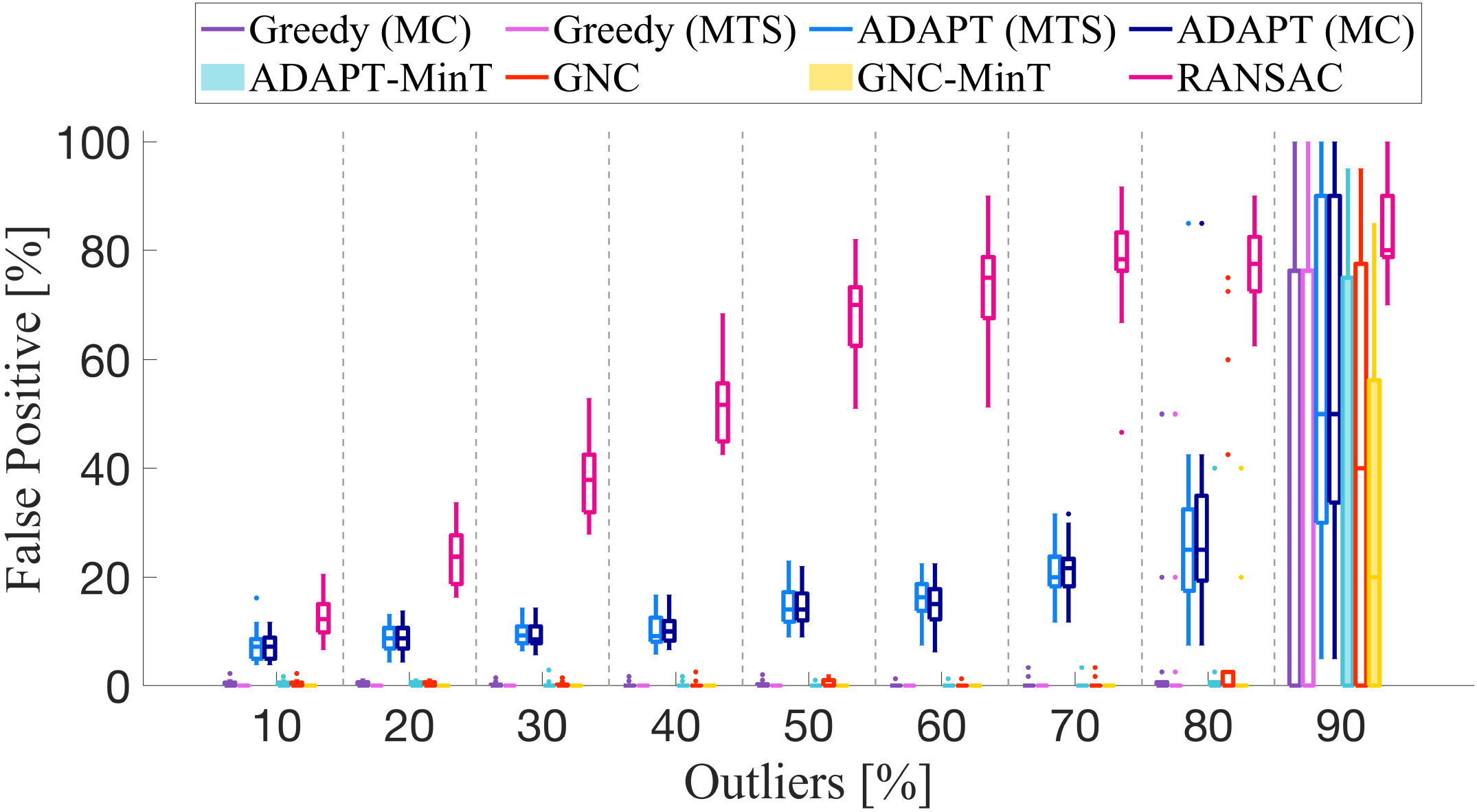} \\
      \end{minipage}
		\end{tabular}
	\end{minipage}
	\mpPostSpace
	\caption{ {\bf Mesh Registration.} True Positive (left) and False Positive (right) of the proposed algorithms, compared to \ransac, on the \scenario{PASCAL+}``aeroplane-2'' dataset~\cite{Xiang2014WACV-PASCAL+}.
	Statistics are computed over 25 Monte Carlo runs and for increasing percentage of outliers.}
	\label{fig:stats_mesh}
	\vspace{-5mm} 
	\end{center}
\end{figure*}

\begin{figure*}[ht!]
	\begin{center}
	\begin{minipage}{\textwidth}
	\begin{tabular}{cc}%
		%%%%%%%%%%%%%%%%%%%%%%%%%%%%%%%%%%%%%%%%%%%%%%%%%%%%%%%%%%%%%%%%%%%%%%%%%%%%%%%%%%%%%%%%%%%%%%%%%%%%%%%%%
		\mpPreSpace
			\begin{minipage}{\mpColTwo}%
        \centering%
        \includegraphics[width=\columnwidth]{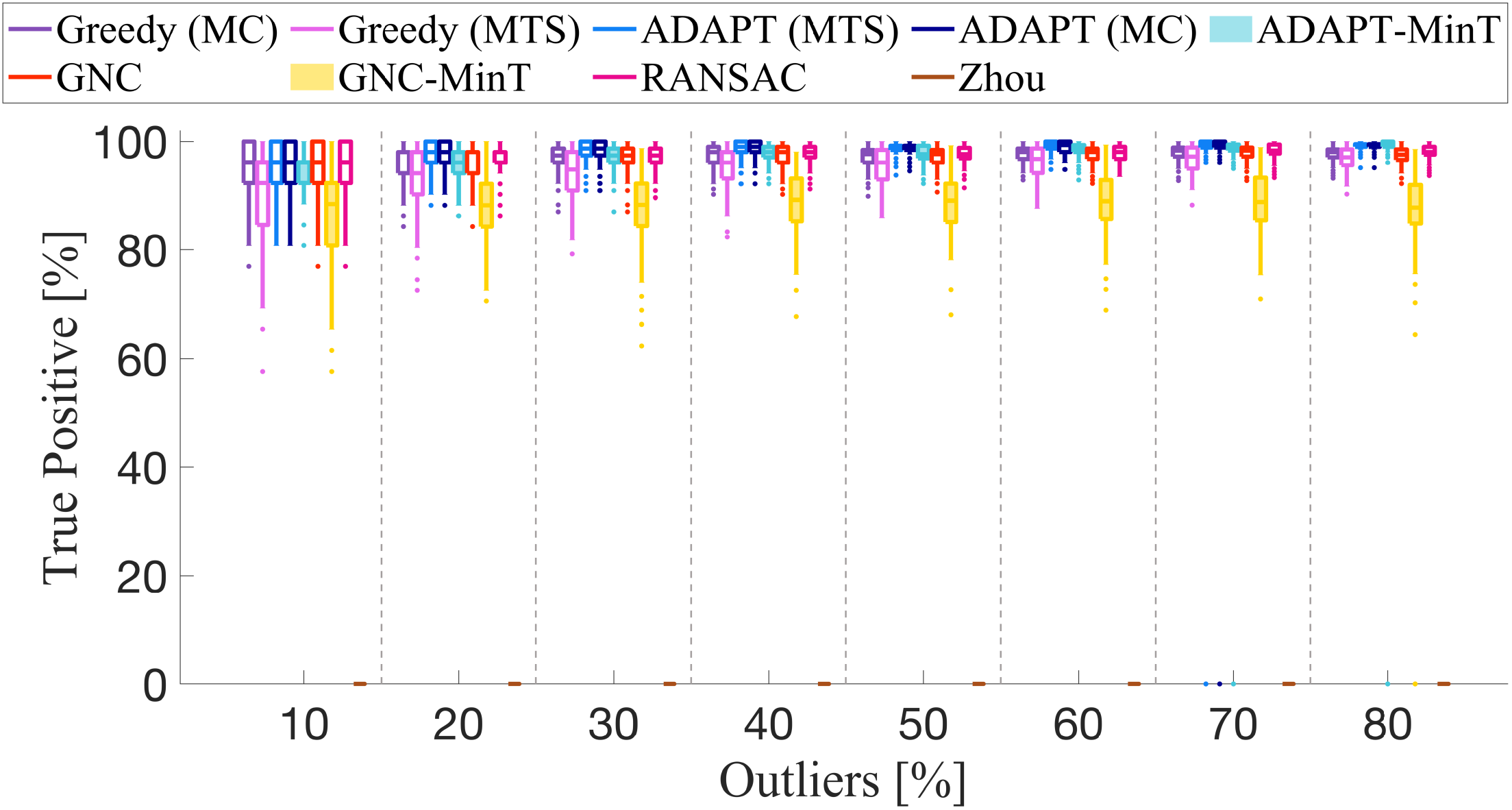} \\
			\end{minipage}
		& \mpMidSpaceTwo
			\begin{minipage}{\mpColTwo}%
        \centering%
        \includegraphics[width=\columnwidth]{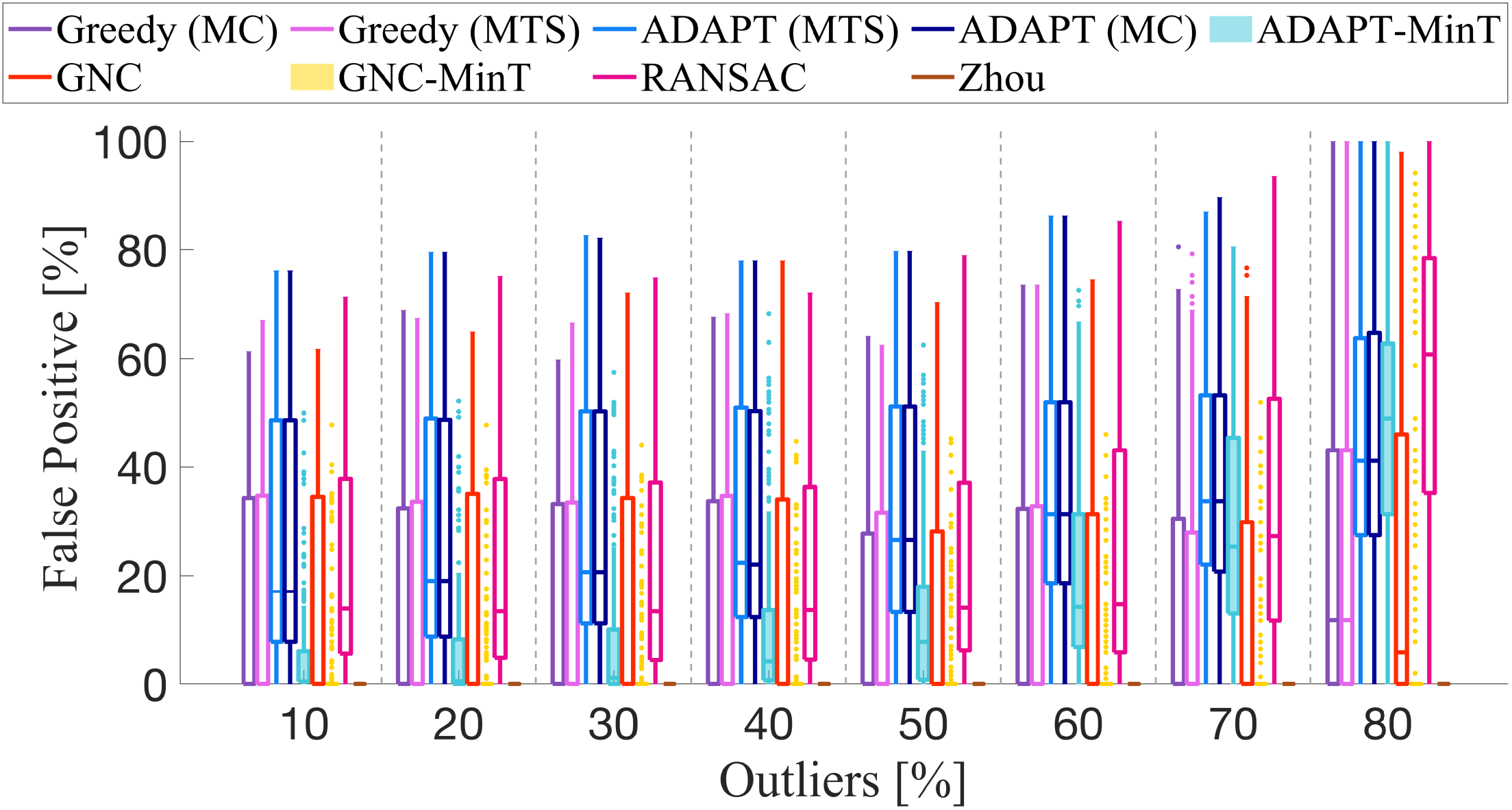} \\
      \end{minipage}
		\end{tabular}
	\end{minipage}
	\mpPostSpace
	\caption{ {\bf Shape Alignment.} 
True Positive (left) and False Positive (right) of the proposed algorithms, compared to state-of-the-art techniques, on the \scenario{FG3DCar} dataset~\cite{Lin14eccv-modelFitting}.
	Statistics are computed over 25 Monte Carlo runs and for increasing percentage of outliers.}
	\label{fig:stats_shape}
	\vspace{-5mm} 
	\end{center}
\end{figure*}

\begin{figure*}[ht!]
	\begin{center}
	\begin{minipage}{\textwidth}
	\begin{tabular}{ccc}%
		%%%%%%%%%%%%%%%%%%%%%%%%%%%%%%%%%%%%%%%%%%%%%%%%%%%%%%%%%%%%%%%%%%%%%%%%%%%%%%%%%%%%%%%%%%%%%%%%%%%%%%%%%
		\mpPreSpace
			\begin{minipage}{\mpColTwo}%
        \centering%
        \includegraphics[width=\columnwidth]{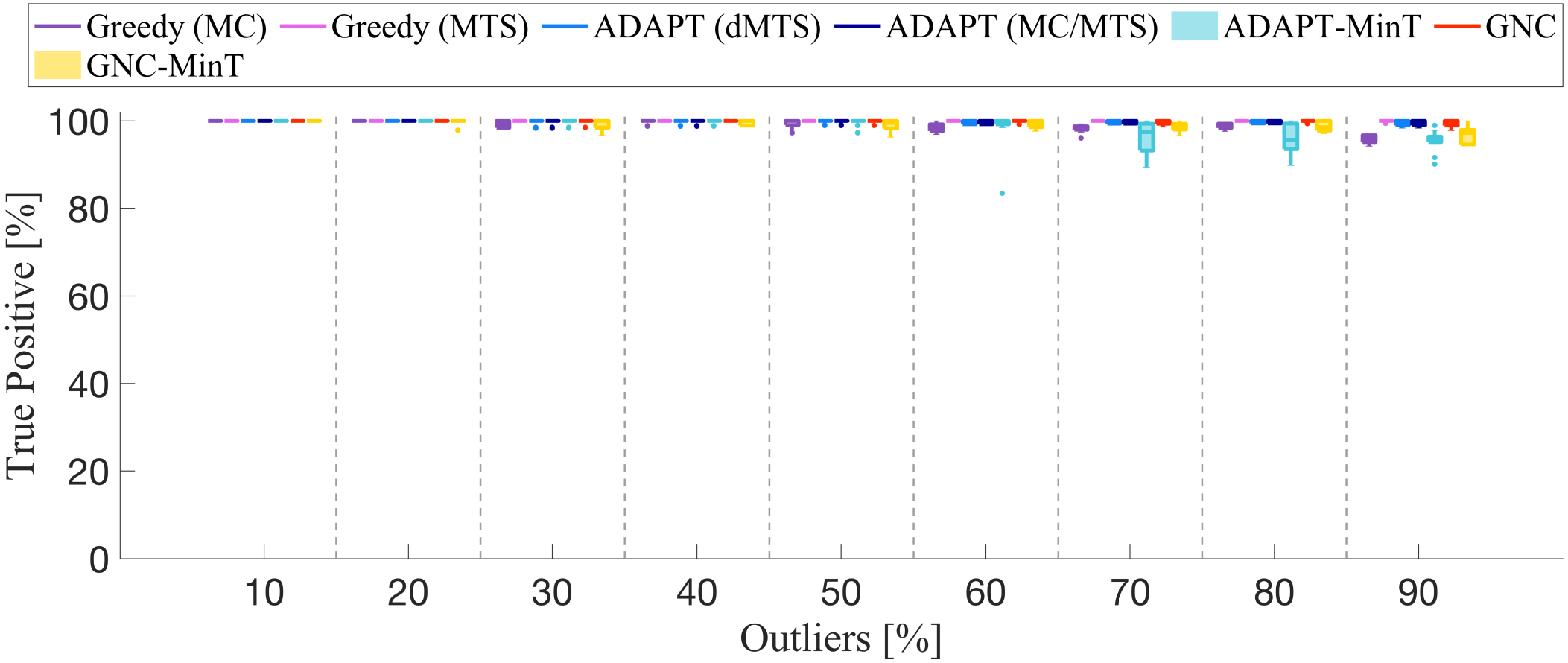} \\
			\end{minipage}
		& \mpMidSpaceTwo
			\begin{minipage}{\mpColTwo}%
        \centering%
        \includegraphics[width=\columnwidth]{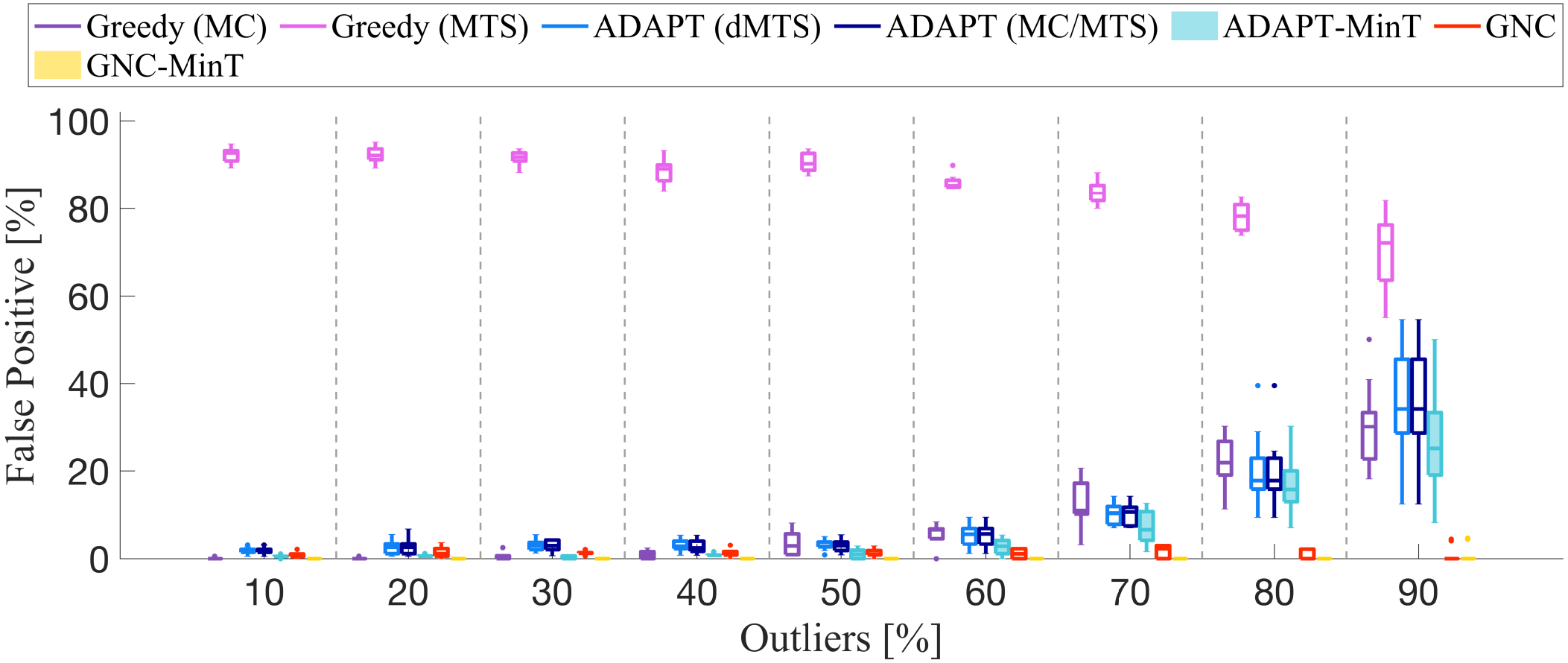} \\
			\end{minipage}
		\end{tabular}
	\end{minipage}
	\mpPostSpace
	\caption{ {\bf 2D SLAM (Grid).} True Positive (left) and False Positive (right) of the proposed algorithms on a synthetic grid dataset for increasing outliers.}
	\label{fig:stats_slam_grid}
	\vspace{-5mm} 
	\end{center}
\end{figure*}

\begin{figure*}[ht!]
	\begin{center}
	\begin{minipage}{\textwidth}
	\begin{tabular}{ccc}%
		%%%%%%%%%%%%%%%%%%%%%%%%%%%%%%%%%%%%%%%%%%%%%%%%%%%%%%%%%%%%%%%%%%%%%%%%%%%%%%%%%%%%%%%%%%%%%%%%%%%%%%%%%
		\mpPreSpace
			\begin{minipage}{\mpColTwo}%
        \centering%
        \includegraphics[width=\columnwidth]{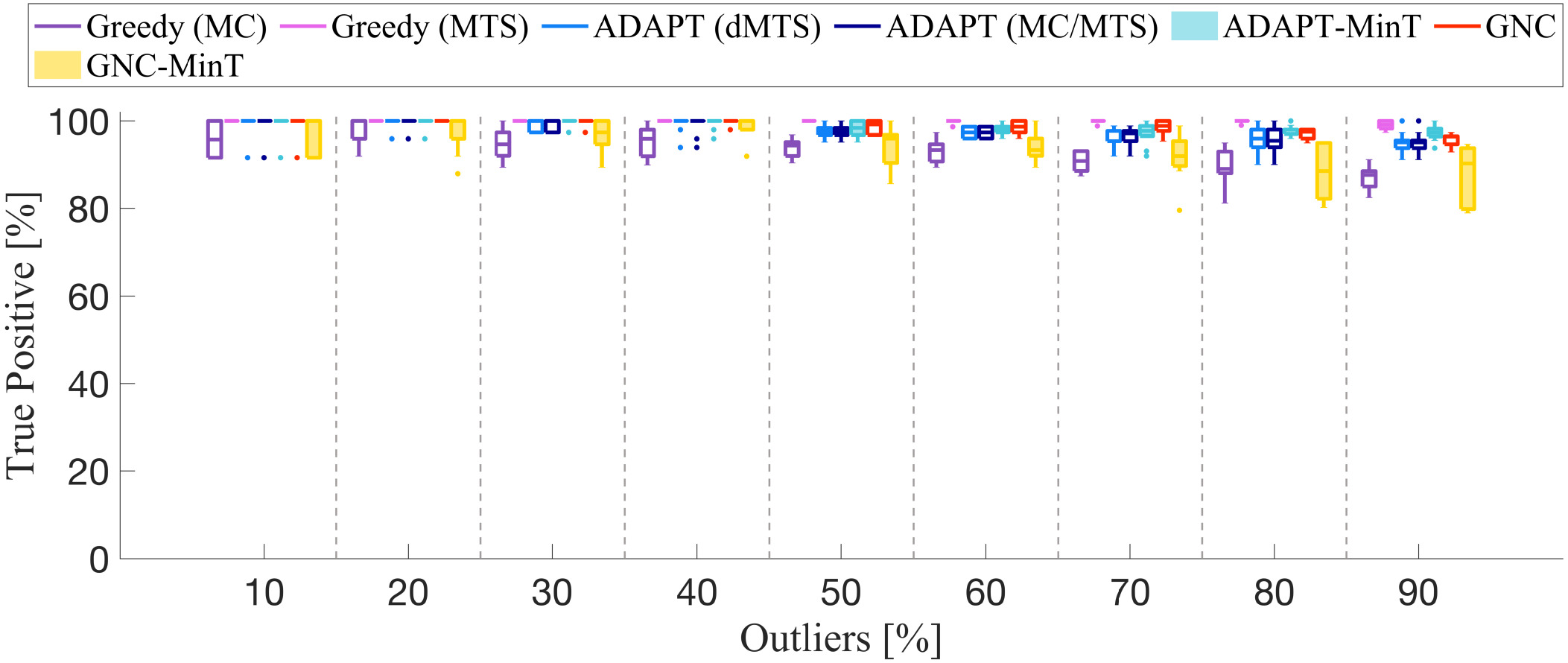} \\
			\end{minipage}
		& \mpMidSpaceTwo
			\begin{minipage}{\mpColTwo}%
        \centering%
        \includegraphics[width=\columnwidth]{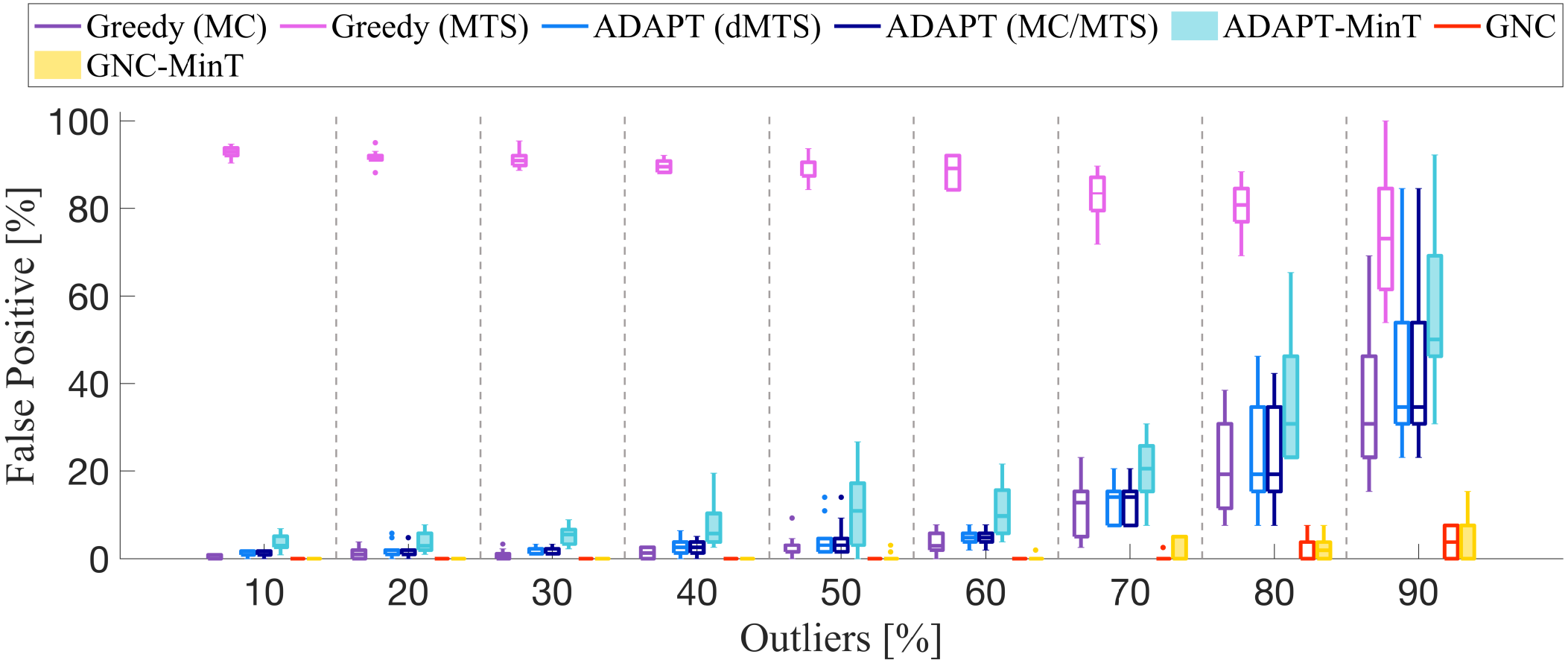} \\
			\end{minipage}
		\end{tabular}
	\end{minipage}
	\mpPostSpace
	\caption{ {\bf 2D SLAM (\scenario{CSAIL}).} True Positive (left) and False Positive (right) of the proposed algorithms on the \scenario{CSAIL} dataset for increasing outliers.}
	\label{fig:slam_stats_csail}
	\vspace{-5mm} 
	\end{center}
\end{figure*}
%!TEX root = ../../main.tex

\begin{figure*}[ht!]
	\begin{center}
	\begin{minipage}{\textwidth}
	\begin{tabular}{ccc}%
		%%%%%%%%%%%%%%%%%%%%%%%%%%%%%%%%%%%%%%%%%%%%%%%%%%%%%%%%%%%%%%%%%%%%%%%%%%%%%%%%%%%%%%%%%%%%%%%%%%%%%%%%%
		\mpPreSpace
			\begin{minipage}{\mpColTwo}%
        \centering%
        \includegraphics[width=\columnwidth]{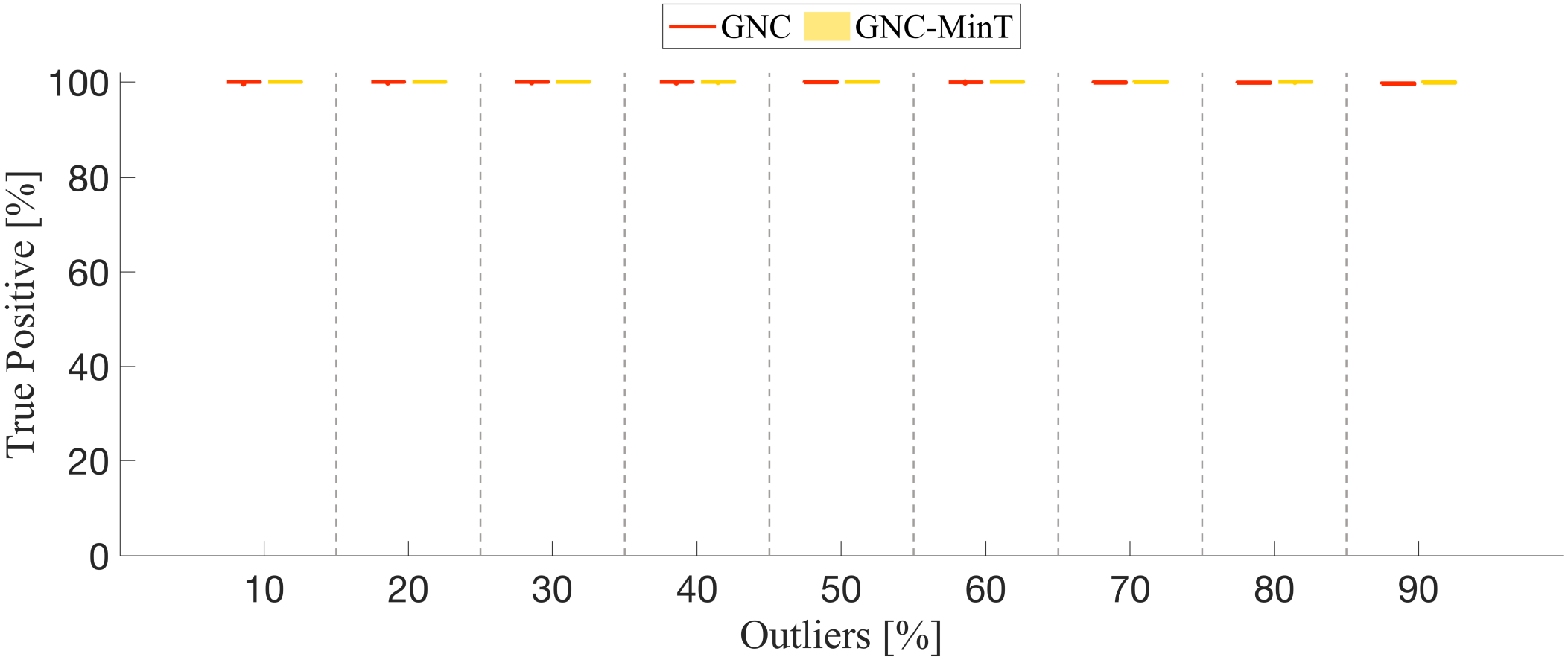} \\
			\end{minipage}
		& \mpMidSpaceTwo
			\begin{minipage}{\mpColTwo}%
        \centering%
        \includegraphics[width=\columnwidth]{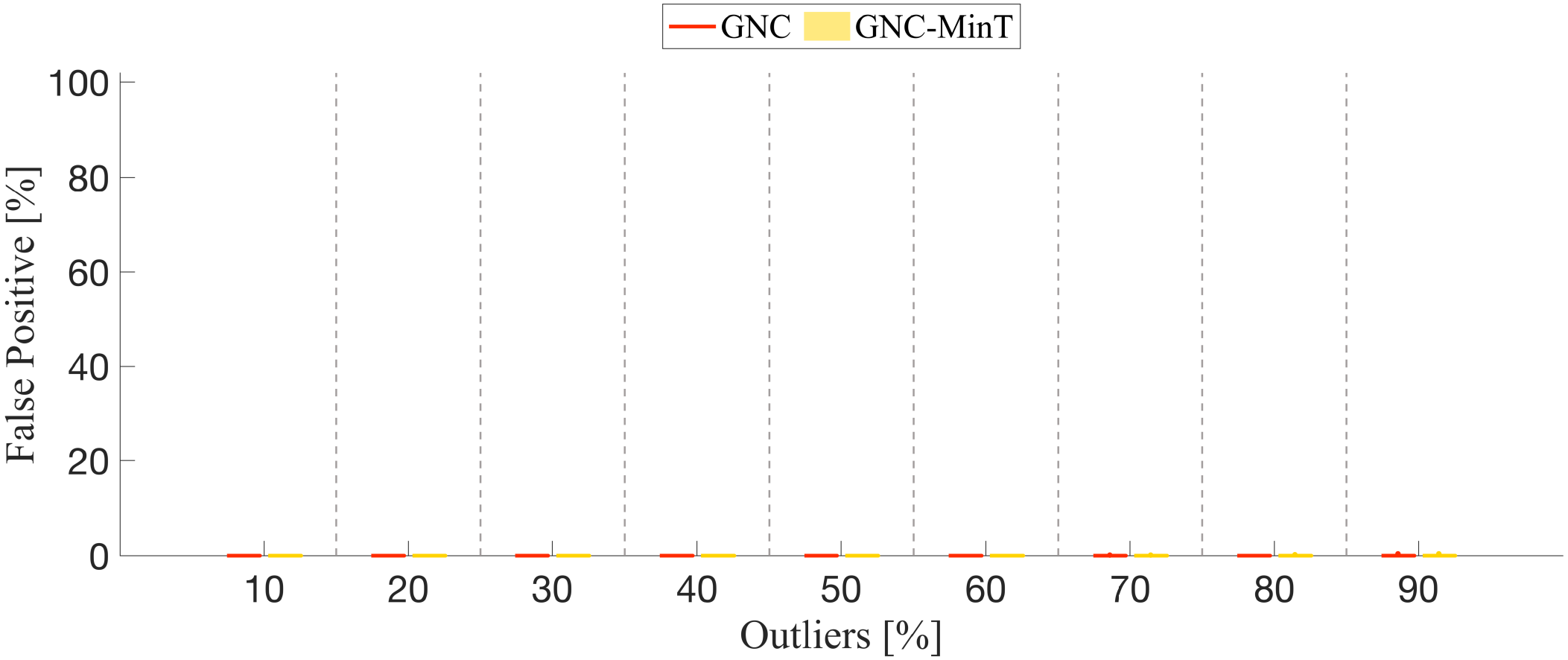} \\
			\end{minipage}
		\end{tabular}
	\end{minipage}
	\mpPostSpace
	\caption{ {\bf 3D SLAM (\scenario{Sphere}).} True Positive (left) and False Positive (right) of the proposed algorithms on a synthetic \scenario{Sphere} dataset for increasing outliers.}
	\label{fig:slam_stats_sphere}
	\vspace{-5mm} 
	\end{center}
\end{figure*}
%!TEX root = ../../main.tex

\begin{figure*}[ht!]
	\begin{center}
	\begin{minipage}{\textwidth}
	\begin{tabular}{ccc}%
		%%%%%%%%%%%%%%%%%%%%%%%%%%%%%%%%%%%%%%%%%%%%%%%%%%%%%%%%%%%%%%%%%%%%%%%%%%%%%%%%%%%%%%%%%%%%%%%%%%%%%%%%%
		\mpPreSpace
			\begin{minipage}{\mpColTwo}%
        \centering%
        \includegraphics[width=\columnwidth]{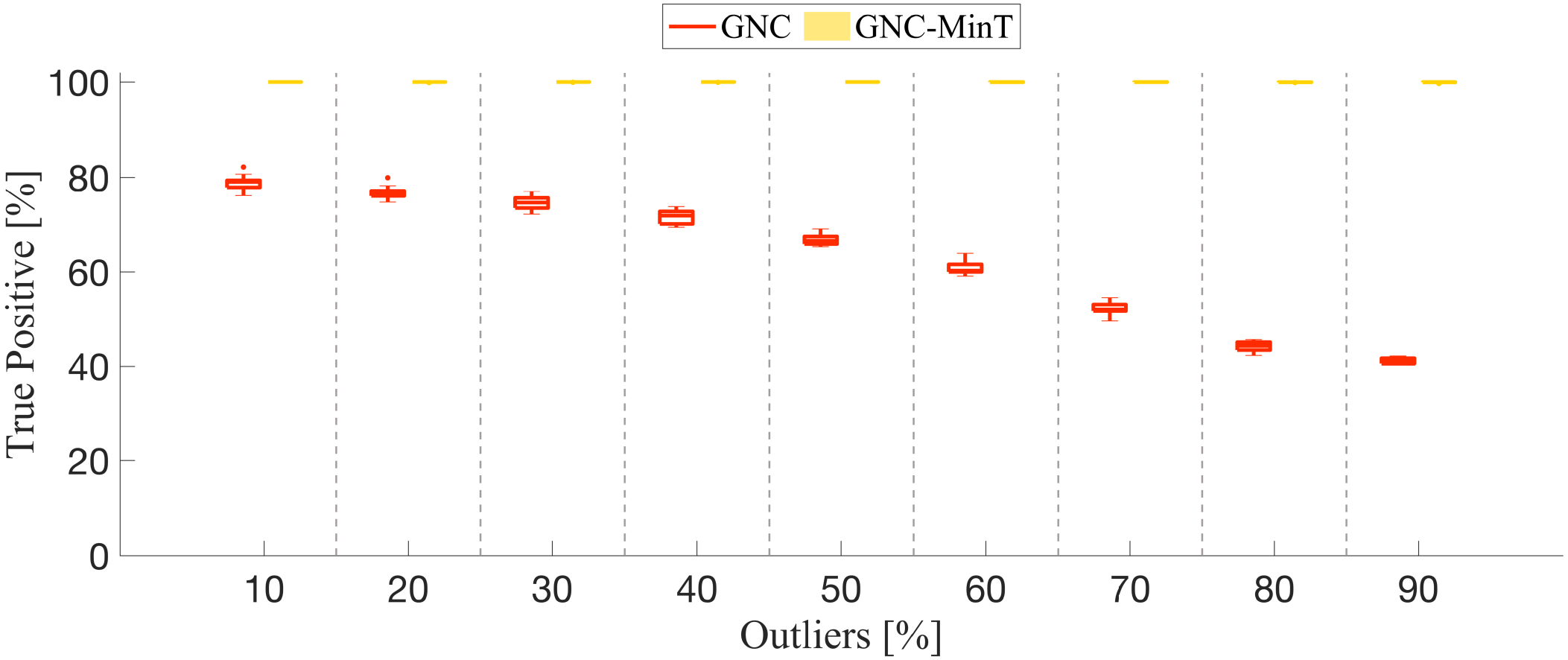} \\
			\end{minipage}
		& \mpMidSpaceTwo
			\begin{minipage}{\mpColTwo}%
        \centering%
        \includegraphics[width=\columnwidth]{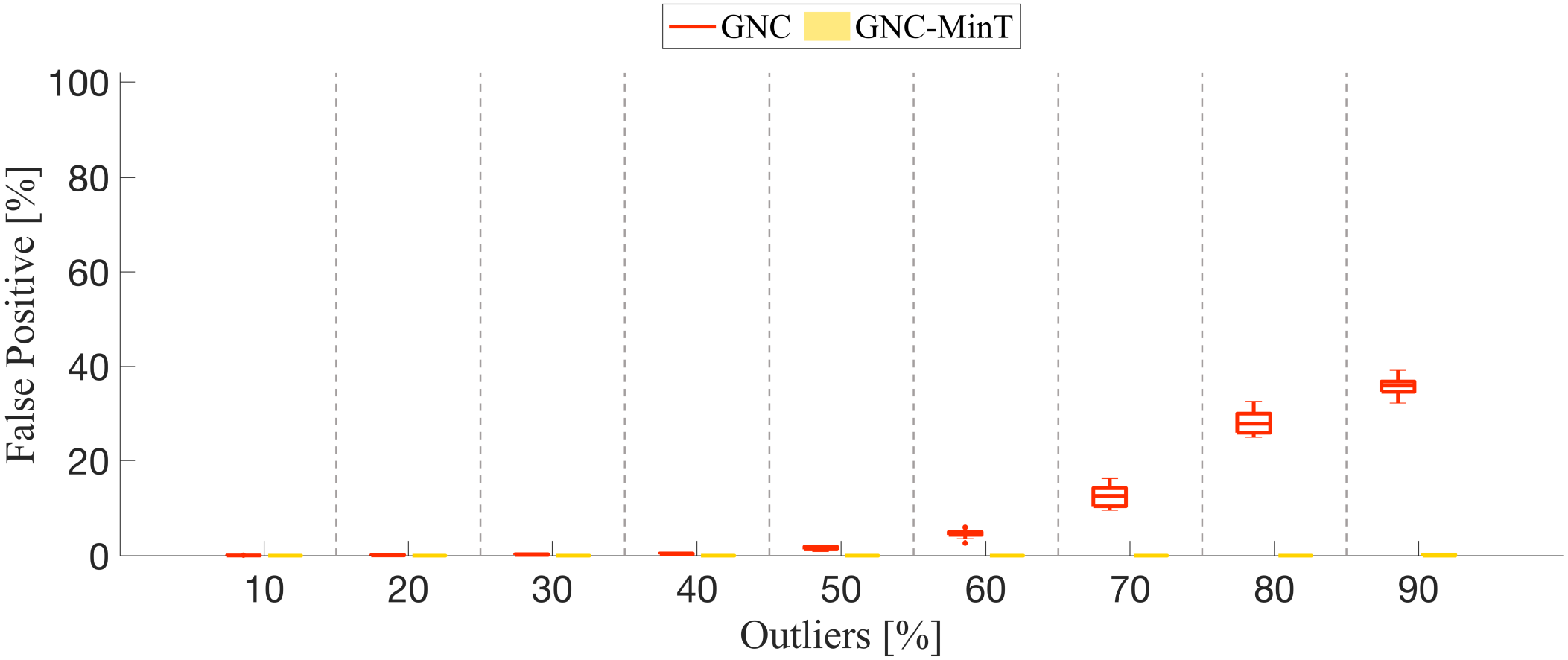} \\
			\end{minipage}
		\end{tabular}
	\end{minipage}
	\mpPostSpace
	\caption{ {\bf 3D SLAM (\scenario{Garage}).} True Positive (left) and False Positive (right) of the proposed algorithms on the \scenario{Garage} dataset for increasing outliers.}
	\label{fig:slam_stats_garage}
	\vspace{-5mm} 
	\end{center}
\end{figure*}}{}

% \vspace{-5mm}
\bibliographystyle{IEEEtran} 
\bibliography{../../../references/refs.bib,../../../references/myRefs.bib}

\vspace{-5mm}
\begin{biography}[{\includegraphics[width=1in,height=1.25in,keepaspectratio]{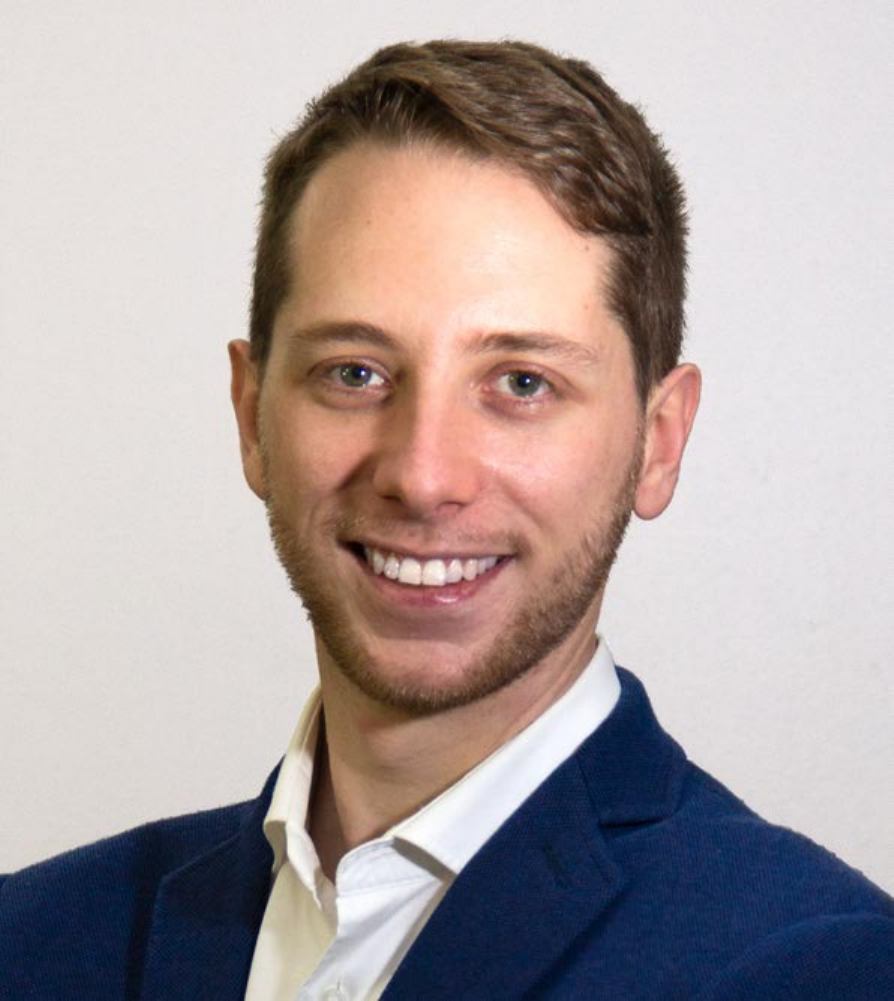}}]{Pasquale Antonante} is a Ph.D.~candidate in the Department of Aeronautics and Astronautics and the Laboratory for Information and Decision Systems (LIDS) at the Massachusetts Institute of Technology, where he is working with Prof.~Luca Carlone at the SPARK Lab.
He has obtained a B.Sc.~degree in Computer Engineering from the University of Pisa, Italy, in 2014; and a S.M.~degree (with honors) in Embedded Computing Systems from the Scuola Superiore Sant'Anna of Pisa, Italy, in 2017.
Prior to MIT, he was a research scientist at the United Technologies Research Center in Cork (Ireland).
His interests include safe and trustworthy perception with applications to single and multi-robot autonomous systems.
Pasquale Antonante is the recipient of the MathWorks Engineering Fellowship, the Best Paper Award in Robot Vision at the 2020 IEEE International Conference on Robotics and Automation (ICRA) and a Honorable Mention from the 2020 IEEE Robotics and Automation Letters (RA-L). 
\end{biography}

\begin{biography}[{\includegraphics[width=1in,height=1.25in,keepaspectratio]{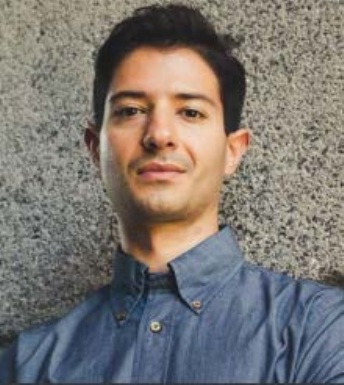}}]{Vasileios Tzoumas} received his Ph.D. in Electrical and Systems Engineering at the University of Pennsylvania (2018). He holds a Master of Arts in Statistics from the Wharton School of Business at the University of Pennsylvania (2016); a Master of Science in Electrical Engineering from the University of Pennsylvania (2016); and a diploma in Electrical and Computer Engineering from the National Technical University of Athens (2012).
	Vasileios is as an Assistant Professor in the Department of Aerospace Engineering, University of Michigan, Ann Arbor. 
	Previously, he was at the Massachusetts Institute of Technology (MIT), in the Department of Aeronautics and Astronautics, and in the Laboratory for Information and Decision Systems (LIDS), were he was a research scientist (2019-2020), and a post-doctoral associate (2018-2019). 
	Vasileios works on control, learning, and perception, as well as combinatorial and distributed optimization, with applications to robotics, cyber-physical systems, and self-reconfigurable aerospace systems.  He cares for trustworthy collaborative autonomy.  His work includes foundational results on robust and adaptive combinatorial optimization, with applications to multi-robot information gathering for resiliency against robot failures and adversarial removals.
	Vasileios is a recipient of the Best Paper Award in Robot Vision at the 2020 IEEE International Conference on Robotics and Automation (ICRA), of an Honorable Mention from the 2020 IEEE Robotics and Automation Letters (RA-L), and was a Best Student Paper Award finalist at the 2017 IEEE Conference in Decision and Control (CDC).
\end{biography}

\vspace{-10mm}
\begin{biography}[{\includegraphics[width=1in,height=1.8in,clip,keepaspectratio]{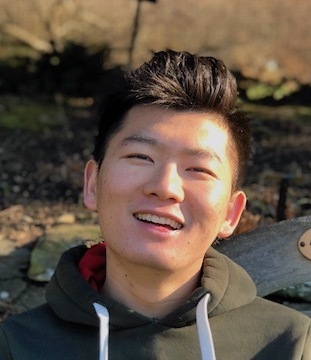}}]{Heng Yang} is a Ph.D.~candidate in the Department
	of Mechanical Engineering and the Laboratory for
	Information \& Decision Systems (LIDS) at the
	Massachusetts Institute of Technology (MIT), where he is
	working with Prof.~Luca Carlone at the SPARK
	Lab. He has obtained a B.Sc.~degree in Mechanical
	Engineering (with honors) from the Tsinghua University, Beijing, China, in 2015; and an S.M.~degree
	in Mechanical Engineering from MIT in 2017.
	His research interests include convex optimization, semidefinite and
	moment/sums-of-squares relaxation, robust estimation and~machine learning, applied to
	robot perception and computer vision. 
	His~work includes developing certifiable outlier-robust machine perception algorithms, large-scale semidefinite programming solvers, and self-supervised geometric perception frameworks.
	Heng Yang is a recipient of the Best Paper Award in Robot Vision at the 2020 IEEE International Conference on Robotics and Automation (ICRA), and a Best Paper Award Honorable Mention from the 2020 IEEE Robotics and Automation Letters (RA-L). He is a Class of 2021 Robotics: Science and Systems (RSS) Pioneer. 
\end{biography}

\vspace{-10mm}
\begin{biography}[{\includegraphics[width=1in,height=1.25in,clip,keepaspectratio]{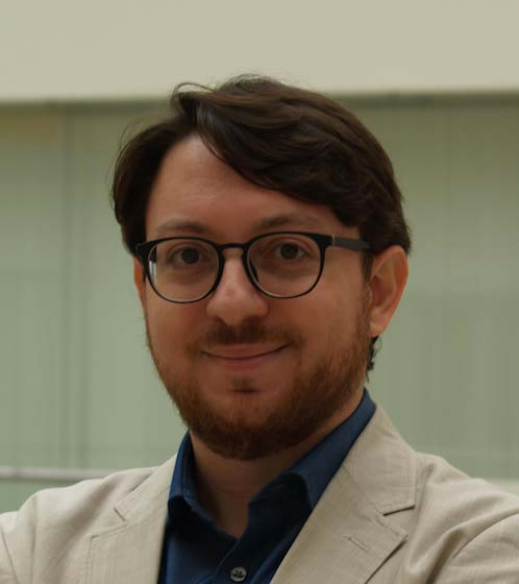}}]{Luca Carlone} is the Leonardo Career Development Assistant Professor in the Department of Aeronautics
	and Astronautics at the Massachusetts Institute of
	Technology, and a Principal Investigator in the Laboratory for Information \& Decision Systems (LIDS). 	He joined LIDS as a postdoctoral associate (2015) and later as a Research
	Scientist (2016), after spending two years as a postdoctoral fellow at the
	Georgia Institute of Technology (2013-2015). 
	He has obtained a B.S.~degree in mechatronics
	from the Polytechnic University of Turin, Italy	(2006); 
	an S.M.~degree in mechatronics from the
	Polytechnic University of Turin, Italy (2008); an
	S.M.~degree in automation engineering from the
	Polytechnic University of Milan, Italy (2008); and
	a Ph.D.~degree in robotics from the Polytechnic University of Turin (2012).
	His research interests include
	nonlinear estimation, numerical and distributed optimization, and probabilistic
	inference, applied to sensing, perception, and decision-making in single and
	multi-robot systems. 
	%His work includes seminal results on certifiably correct
	% algorithms for localization and mapping, as well as approaches for visual-inertial navigation and distributed mapping.
	%
	He is a recipient of the 
	Best Paper Award in Robot Vision at ICRA 2020, 
	a 2020 Honorable Mention from the IEEE Robotics and Automation Letters, 
	a Track Best Paper award at the 2021 IEEE Aerospace Conference, 
	the 2017 Transactions on Robotics King-Sun Fu Memorial Best Paper Award, 
	the Best Paper Award at WAFR 2016, the Best Student Paper Award at the 2018 Symposium on VLSI Circuits, and he was best paper finalist at RSS 2015. He is also a recipient of the NSF CAREER Award (2021), the RSS Early Career Award (2020), the Google Daydream (2019) and the Amazon Research Award (2020), and the MIT AeroAstro Vickie Kerrebrock Faculty Award (2020). 
\end{biography}

\end{document}